\documentclass[lettersize,journal]{IEEEtran}
\usepackage{amsmath,amsfonts}
\usepackage{algorithmic}
\usepackage{algorithm}
\usepackage{array}
\usepackage[caption=false,font=normalsize,labelfont=sf,textfont=sf]{subfig}
\usepackage{textcomp}
\usepackage{stfloats}
\usepackage{url}
\usepackage{hyperref}
\usepackage{verbatim}
\usepackage{graphicx}
\usepackage{cite}
\usepackage{cleveref}
\usepackage{xspace}
\usepackage{bbm}
\usepackage{multicol}
\usepackage{multirow}
\usepackage[table, xcdraw]{xcolor}
\usepackage{colortbl}
\usepackage{hhline}
\usepackage{pifont}
\usepackage{extarrows} 
\def\onedot{.\xspace}
\def\eg{{e.g}\onedot} 
\def\ie{{i.e}\onedot}

\def\etal{\emph{et al}\onedot}

\newcommand{\layerIdx}{\mathit{i}}
\newcommand{\layerTotal}{I}
\newcommand{\classIdx}{\mathit{n}}
\newcommand{\classTotal}{N}

\newcommand{\featureset}{\mathbf{F}}
\newcommand{\embeddingvec}{\mathbf{z}}
\newcommand{\embeddingset}{\mathbf{Z}}

\newcommand{\anchor}{\mathbf{a}}
\newcommand{\anchors}{\mathbf{A}}

\newcommand{\gtfunc}{g}

\newcommand{\segGT}{\mathbf{Y}}
\newcommand{\segPred}{\hat{\segGT}}

\newcommand{\binaryImg}{\mathbf{B}}
\newcommand{\distImg}{\mathbf{D}}

\newcommand{\edgeSet}{\mathbf{E}}

\def\etal{\emph{et al}\onedot}

\definecolor{instructioncolor}{rgb}{0.0, 0.2, 1.0}
\definecolor{instructioncolor2}{rgb}{0.0, .0, 0.0}
\definecolor{instructioncolor3}{rgb}{0.0, .0, 1.0}
\definecolor{instructioncolor4}{rgb}{1.0, 0.3, 0.0}

\definecolor{ForestGreen}{rgb}{0.13, 0.55, 0.13}
\definecolor{BrickRed}{rgb}{0.8, 0.16, 0.1}  % Define BrickRed manually

\begin{document}

\title{Contextrast++: Robust Multi-Scale Contextual Contrastive Learning for Semantic Segmentation}

\author{Changki Sung$^{1\dagger}$, Hyungtae Lim$^{2\dagger}$, Wanhee Kim$^3$, Youngwoo Seo$^4$, and Hyun Myung$^{5\ast}$,~\IEEEmembership{Senior Member,~IEEE}
        % <-this % stops a space
\thanks{
© 2026 IEEE. Personal use of this material is permitted. Permission from IEEE must be obtained for all other uses, in any current or future media, including reprinting/republishing this material for advertising or promotional purposes, creating new collective works, for resale or redistribution to servers or lists, or reuse of any copyrighted component of this work in other works. DOI: 10.1109/TPAMI.2026.3721845

$\ast$Corresponding author: Hyun Myung 

$^\dagger$These authors contributed equally to this work.

$^1$Changki Sung is with the Information \& Electronics Research Institute, KAIST, Daejeon, 34141, Republic of Korea. { cs1032@kaist.ac.kr}

$^2$Hyungtae Lim is with Zoox Inc., Foster City, CA 94404, USA.
{ hlim@zoox.com}. This work was conducted in personal time and independently
of the author's organization.

$^3$Wanhee Kim is with the Robotics Program, KAIST (Korea Advanced Institute of Science and Technology), Daejeon, 34141, Republic of Korea. { gml78905@kaist.ac.kr}

$^4$Youngwoo Seo is with Hanwha Aerospace, Pangyo, Gyeonggi-do, 13488, Republic of Korea. { youngwoo.seo@hanwha.com}

$^5$Hyun Myung is with the School of Electrical Engineering, KAIST, Daejeon, 34141, Republic of Korea. { hmyung@kaist.ac.kr}

A preliminary version of this work has appeared in CVPR 2024 \url{https://doi.org/10.1109/CVPR52733.2024.00358}}}
%   This paper was produced by the IEEE Publication Technology Group. They are in Piscataway, NJ.}% <-this % stops a space

% \markboth{Journal of \LaTeX\ Class Files,~Vol.~\#, No.~\#, March~2025}%
% {Sung \MakeLowercase{\textit{et al.}}: A Sample Article Using IEEEtran.cls for IEEE Journals}


{
    \maketitle
    \begin{abstract}
    Semantic segmentation has rapidly advanced with deep learning; however, challenges remain in effectively capturing local and global contexts as well as addressing the long-tailed distribution problem.
    To tackle these issues, we present \textit{Contextrast++}, a robust contrastive learning method for semantic segmentation that improves multi-scale feature integration and mitigates class imbalance issues.
    Our method consists of two key components: 1)~contextual contrastive learning (CCL) and 2)~boundary-aware negative (BANE) sampling.
    CCL includes three subcomponents: adaptive fusion module, pixel-to-anchor (PA) loss, and anchor-to-anchor (AA) loss.
    The adaptive fusion module dynamically balances local and global feature integration, resulting in a more context-aware representation.
    While the PA loss leverages the fused multi-scale features to improve feature representation learning, the AA loss focuses on addressing the long-tailed distribution problem by utilizing a memory bank that stores a fixed number of class-balanced representative anchors.
    Meanwhile, BANE sampling enhances segmentation precision by selecting hard negatives from misclassified boundary regions, which refines fine-grained details during contrastive learning.
    As verified in extensive experiments using public datasets,  
    we demonstrate that Contextrast++ substantially improves semantic segmentation performance over existing contrastive learning-based state-of-the-art approaches, while introducing no additional computational overhead during inference.
\end{abstract}

    \section{Introduction}

 Semantic segmentation, a fundamental task in computer vision, plays a vital role in applications such as autonomous driving and robotics~\cite{sanderson2022fcn, hurtado2022semantic, tzelepi2021semantic,yang2024lcfnets, ni2023robust,feng2024segmentation,gu2024clft, wu2024s, liang2024multi, fan2022mlfnet, ziwen2023multi}. Recent advancements in deep learning, driven by the availability of extensive datasets~\cite{deng2009imagenet,cordts2016cityscapes,mottaghi2014role,caesar2018coco,brostow2009semantic,zhou2017scene}, have boosted segmentation performance. Researchers have introduced larger deep neural networks~\cite{chen2017deeplab, chen2017rethinking, zhao2017pyramid, chen2018encoder, xiao2018unified, li2018pyramid, ke2018adaptive, yu2018learning, yurtkulu2019semantic, sun2019high, fu2019dual, yu2020context, yuan2020object, choi2020cars, hong2021deep, huynh2021progressive, liu2021swin, zhong2023understanding, ye2024invpt++, xu2024mctformer+, zhou2024object, chen2024frequency, zhou2024prototype, zhang2020causal, pei2022hierarchical, chen2023multi, zhang2021self, chen2024spatial, li2021ctnet, liu2013weakly, zhu2025merging, fu2025segman, shi2025llmformer, yang2025m} and novel loss functions~\cite{yuan2020segfix,wang2022active,tan2022semantic} to further improve segmentation accuracy.

 \begin{figure}[t!]
    \centering
    \includegraphics[scale=0.3]{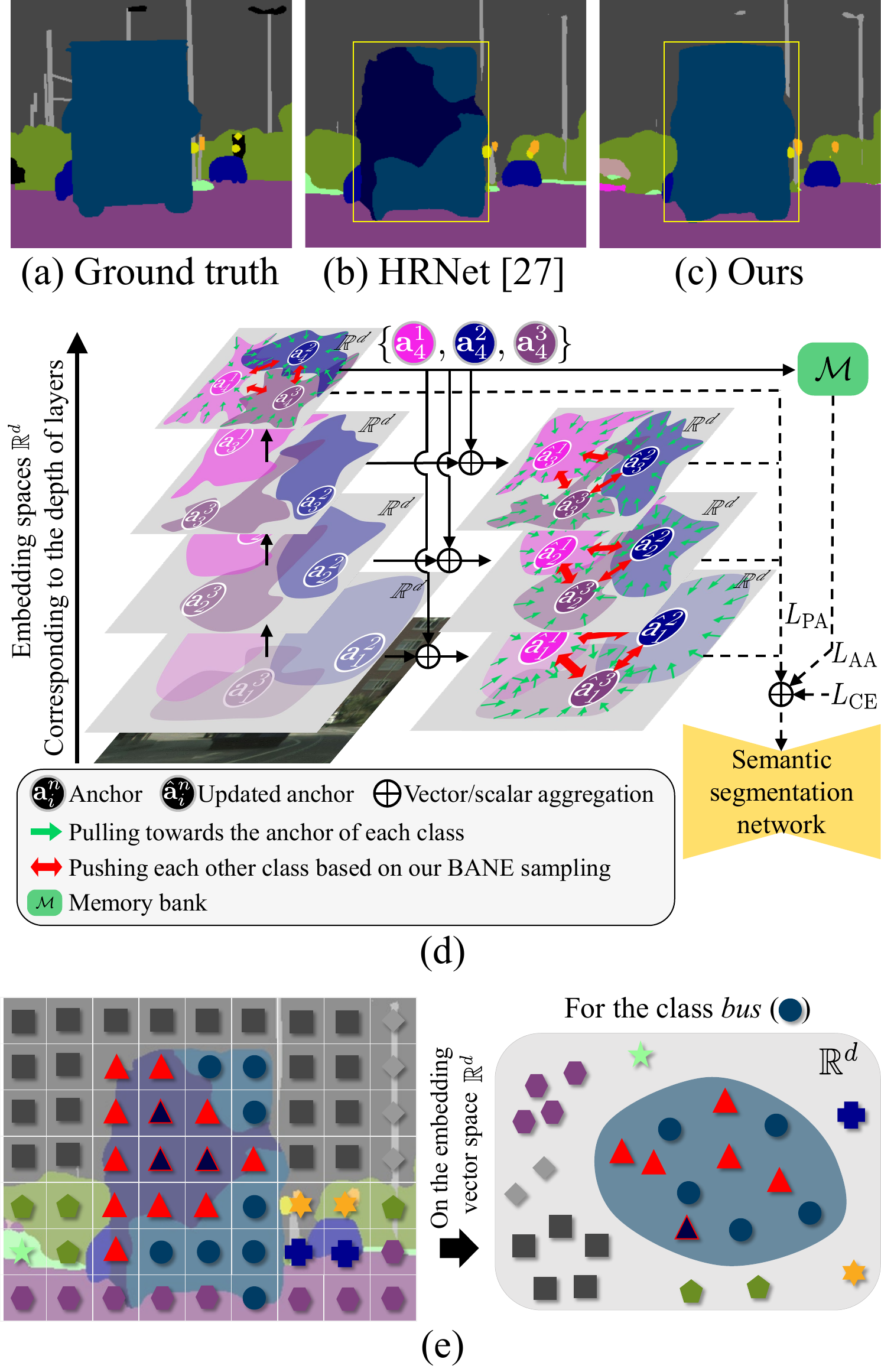}
    \caption{Visualization of (a)~ground truth annotations, (b)~results obtained from HRNet~\cite{sun2019high}, and (c)~output after being trained with our contrastive learning approach.
      (d)~Overview of the proposed contextual contrastive learning framework, referred to as \textit{Contextrast++},
      which refines class-specific anchors by aggregating the deepest-layer embeddings into earlier layers to enhance semantic and fine-grained feature representation.
      (e)~Illustration of boundary-aware negative (BANE) sampling strategy, 
      which selects feature embeddings near object boundaries (solid red triangles) as harder negative samples.
      This helps contrastive learning focus more on boundary regions, improving boundary prediction accuracy.
    }	
    \label{fig:key_ideas}
\end{figure}

Despite recent advancements, some semantic segmentation models still produce inaccurate results, as illustrated in Fig.~\ref{fig:key_ideas}(b). More specifically, semantic segmentation models continue to face challenges in accurately predicting small objects, object boundaries, and underrepresented (or long-tailed) classes. While increasing model complexity~\cite{strudel2021segmenter, li2022deep, woo2023convnext} can help mitigate these challenges, it results in higher memory consumption and slower inference speeds, which are critical drawbacks for autonomous driving and robotics. This underscores the need for efficient performance enhancements that do not require additional computational cost during inference.

To address the challenges of inaccurate predictions in small objects, boundaries, and underrepresented classes, as well as the high computational cost associated with complex models, contrastive learning-based methods~\cite{wang2021exploring, hu2021region, zhou2024cross, pissas2022multi,sung2024contextrast} have emerged as an effective approach. Contrastive learning is applied exclusively during training, so semantic segmentation models retain identical memory consumption and processing time during inference. Consequently, contrastive learning offers a highly efficient solution for scenarios where low computational overhead and fast inference are essential.

However, existing contrastive learning methods for semantic segmentation either overlook multi-scale feature integration~\cite{wang2021exploring, hu2021region, zhou2024cross} or rely on separate multi-scale and cross-scale processes that cause inconsistent feature relationships~\cite{pissas2022multi}. Our previous work, Contextrast~\cite{sung2024contextrast}, addressed this through fused representative anchors across scales, but had two limitations: (1)~static weighted fusion lacks adaptability for diverse contexts, and (2)~class imbalance within mini-batches leads to insufficient long-tailed distribution handling, where underrepresented classes have limited training instances.

Therefore, in this paper, we propose Contextrast++ to address these limitations. First, Contextrast++ introduces a lightweight adaptive fusion module that replaces the static weighted sum with a learnable mechanism, using self-attention, concatenation, and multi-layer perceptron~(MLP) to adaptively integrate multi-scale anchors. This generates expressive class-specific feature vectors that enable the enhanced pixel-to-anchor (PA) loss to learn more consistent global-local representations.

Second, Contextrast++ introduces an anchor-to-anchor~(AA) loss with a memory bank that stores class-balanced representative anchors. Unlike existing methods~\cite{wang2021exploring,zhou2024cross} that store pixel-level features leading to imbalance, our anchor selection strategy extracts one anchor per class from the deepest layer when the class is present, storing only refined-then-fused class-level anchors. This enables class-balanced learning, effectively mitigating the long-tailed distribution problem.

In summary, this paper presents the following contributions: 
\begin{itemize}
    \item Contextrast++ introduces a lightweight adaptive fusion module that replaces static weighting with learnable integration via self-attention and MLP, enabling adaptively fused representative anchors that enhance both PA loss and AA loss.
    \item We introduce an efficient memory bank design that stores class-balanced representative anchors across all classes, enhancing the AA loss to effectively mitigate the long-tailed distribution problem.
    % \item The BANE sampling strategy allows the model to gather discriminative hard negatives for contrastive learning while maintaining high-resolution semantic details. It progressively guides the model to focus on difficult areas, such as boundaries of misclassified areas, throughout the training process.
    \item Contextrast++ outperforms prior contrastive learning approaches across a range of CNN architectures~\cite{sun2019high,yuan2020object,chen2017rethinking} and transformer architectures~\cite{liu2021swin, cheng2022masked}, evaluated on diverse datasets~\cite{cordts2016cityscapes, brostow2009semantic, mottaghi2014role, caesar2018coco, zhou2017scene}.
\end{itemize}

    \section{Related work}

\subsection{Semantic segmentation}
Semantic segmentation assigns a category label to each pixel in an image. CNN-based methods from FCN~\cite{long2015fully} to ASPP~\cite{chen2017deeplab}, HRNet~\cite{sun2019high}, and OCRNet~\cite{yuan2020object} have progressively improved spatial and contextual understanding. More recently, transformer-based models~\cite{dosovitskiy2020image, liu2021swin, xie2021segformer, cheng2022masked} have further advanced segmentation performance by capturing long-range dependencies, though typically with larger model size and higher computational cost. Contrastive learning-based methods offer a complementary direction by optimizing the embedding space to improve feature discriminability, without introducing additional inference overhead.
% Semantic segmentation assigns a category label to each pixel in an image. 
% Deep learning advances from FCN~\cite{long2015fully} to recent methods including 
% ASPP~\cite{chen2017rethinking}, HRNet~\cite{sun2019high}, OCRNet~\cite{yuan2020object}, 
% and transformer-based models~\cite{dosovitskiy2020image, liu2021swin} have 
% progressively improved spatial and contextual understanding. While these 
% architectures have significantly advanced segmentation performance, contrastive 
% learning-based methods offer a complementary direction by optimizing the 
% embedding space without inference overhead.
\begin{table}[t]
    \centering
    \caption{Comparison of core components between our method and existing state-of-the-art approaches. CL: Contrastive learning, MB: Memory bank, AL: Auxiliary loss, SYN: Synthetic negative sample generation.}
    \resizebox{0.48\textwidth}{!}{%
    \begin{tabular}{l|ccc}
    \hline
    \rowcolor[HTML]{DAE8FC}
    \multicolumn{1}{c|}{Method} & Components & Scale                  & \begin{tabular}[c]{@{}c@{}}Boundary\\
    awareness\end{tabular} \\ \hline
    Baseline                                                & -          & -                      & -                                                            \\
    Wang~\etal \tiny\textit{ICCV} \textit{21}~\cite{wang2021exploring}           & CL, MB     & Single                 & \textcolor{BrickRed}{\ding{55}}                                            \\
    Hu~\etal \tiny\textit{ICCV} \textit{21}~\cite{hu2021region}        & CL, MB, AL & Single                 & \textcolor{BrickRed}{\ding{55}}                                                            \\
    Pissas~\etal \tiny\textit{ECCV} \textit{22}~\cite{pissas2022multi}    & CL         & Multi                  & \textcolor{BrickRed}{\ding{55}}                                                            \\
    Zhou~\etal \tiny\textit{TPAMI} \textit{24}~\cite{zhou2024cross}     & CL, MB, SYN     & Single                 & \textcolor{BrickRed}{\ding{55}}                                                           \\
    Sung~\etal \tiny\textit{CVPR} \textit{24}~\cite{sung2024contextrast} & CL         & Multi-fusion (static)  & \textcolor{ForestGreen}{\ding{51}}                                                        \\
    \rowcolor[HTML]{EFEFEF}
    Ours                                                              & CL, MB     & Multi-fusion (adaptive) & \textcolor{ForestGreen}{\ding{51}}                                                         \\ \hline
    \end{tabular}}
    \label{tab:prop_diff}
\end{table}

\subsection{Contrastive learning approaches in semantic segmentation}
In contrastive learning, feature representations are learned by minimizing intra-class distances while maximizing inter-class separability. Recent works applying contrastive learning to semantic segmentation~\cite{wang2021exploring,pissas2022multi,hu2021region,sung2024contextrast, zhou2024cross} have shown promising results. Notably, contrastive learning is applied only during training, enhancing performance without incurring additional computational costs during inference. Table~\ref{tab:prop_diff} compares the core components of recent contrastive learning approaches.

Previous approaches~\cite{wang2021exploring,hu2021region,zhou2024cross} operated on single-scale features from the final layer. Wang~\etal~\cite{wang2021exploring} and Zhou~\etal~\cite{zhou2024cross} utilized memory banks to store pixel-level features, with Zhou~\etal further introducing synthetic negative sample generation to create harder negatives near class boundaries. Hu~\etal~\cite{hu2021region} additionally employed an auxiliary loss based on class-specific weighted region centers. However, these single-scale methods miss valuable multi-scale contextual information.

Pissas~\etal~\cite{pissas2022multi} addressed this by incorporating multi-scale features, integrating contrastive learning across multi-scale and cross-scale features. Nevertheless, treating these as separate processes leads to inconsistencies, where feature shifts at different scales may cause misaligned relationships.

Our previous work, Contextrast~\cite{sung2024contextrast}, tackled these 
inconsistencies by introducing contextual contrastive learning (CCL) with static 
multi-scale fusion of representative anchors and boundary-aware negative (BANE) 
sampling for harder negatives. However, CCL's static weighted fusion lacks 
adaptability, and its class balancing is limited to classes present within each 
mini-batch, leaving the long-tailed distribution problem partially unresolved.

Contextrast++ advances the CCL framework by replacing the static fusion with 
an adaptive fusion module that employs learnable integration for more expressive 
anchor representations. Additionally, it introduces a memory bank that maintains 
class-balanced anchors across all classes, effectively resolving both the 
inflexible fusion and class imbalance limitations.

\subsection{Multi-scale feature fusion in semantic segmentation}
Several recent works~\cite{zhou2024boundary, gu2022multi, shi2023transformer} have explored multi-scale feature fusion to better integrate global semantics and fine-grained local details in semantic segmentation. These methods, including HRViT~\cite{gu2022multi}, TSG~\cite{shi2023transformer}, and BSCNet~\cite{zhou2024boundary}, commonly rely on weighted summation or static alignment-based aggregation over spatial feature maps, which limits their ability to model richer cross-scale interactions.

More recently, Fu~\etal~\cite{fu2025segman} integrated local attention with state space models for multi-scale context extraction, Shi~\etal~\cite{shi2025llmformer} leveraged large language model priors with scaled visual attention for open-vocabulary segmentation, and Yang~\etal~\cite{yang2025m} proposed a multi-scale encoder enhancement framework for weakly supervised semantic segmentation. These recent methods perform fusion on spatially dense feature maps that preserve pixel-level location information. In contrast, our adaptive fusion module operates on class-specific anchor embeddings that aggregate spatial information into compact class-level vectors, without retaining explicit pixel-level spatial structure.

\subsection{Context modeling and clustering for semantic segmentation}
Recent works have explored context modeling and clustering as core mechanisms for semantic segmentation. Liu~\etal~\cite{liu2013weakly} proposed a weakly supervised dual clustering framework, and Zhu~\etal~\cite{zhu2025merging} introduced CCViM, which integrates context clustering within a Vision Mamba architecture. While these works demonstrate the value of clustering-based representations in weakly supervised and domain-specific settings, our approach tackles supervised semantic segmentation through contrastive learning.

Li~\etal~\cite{li2021ctnet} proposed CTNet, a tandem framework of a channel context module (CCM) and a spatial context module (SCM) that interactively explores channel-wise semantic dependencies and pixel-to-category spatial correlations. CCM generates class-specific feature representations that serve as prior knowledge to guide the SCM. These class-specific feature vectors play a similar role to the class-specific anchors in our CCL, as both encode category-level semantic representations.

Despite this structural similarity, the two methods pursue different objectives and operate in distinct spaces. While CTNet's class features reside in the feature-map space to provide prior knowledge for architectural context modeling, our anchors are constructed in the contrastive embedding space to explicitly regularize the feature distribution via pixel-anchor and anchor-anchor losses. Furthermore, whereas the CCM and SCM are integral components of the network architecture during both training and inference, our CCL functions as an auxiliary training objective and incurs no inference-time overhead. Consequently, we view these two approaches as complementary: CTNet focuses on enhancing feature-map-level context, while CCL strengthens the underlying encoder representations through contrastive supervision.
    \section{Contextrast++: Robust Multi-Scale Contextual Contrastive Learning for Semantic Segmentation}
\label{sec:method}

\begin{figure*}[t]
    \centering
    \includegraphics[scale=0.52]{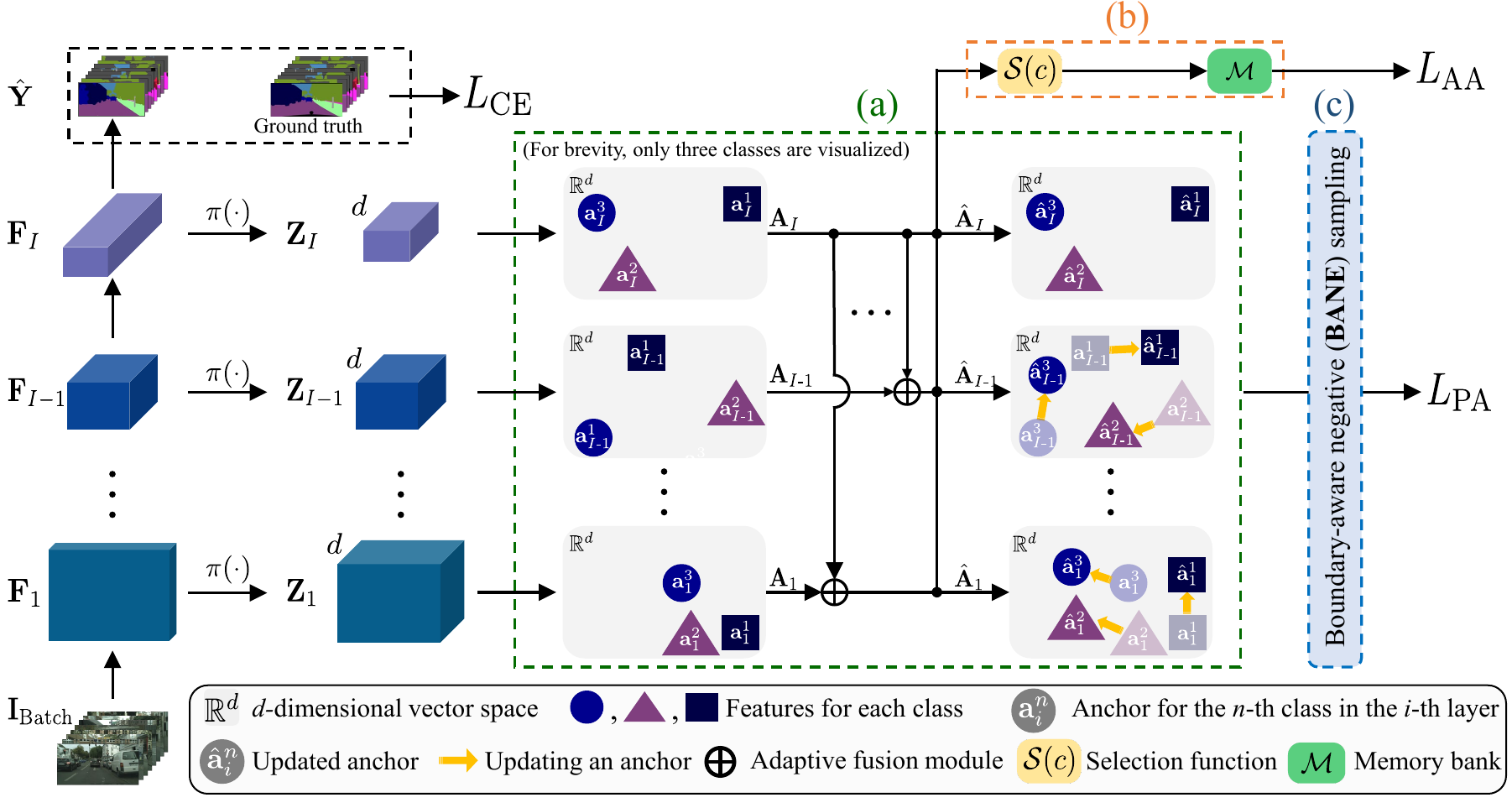}
    \caption{Overview of the proposed pipeline, called Contextrast++. (a) The contextual contrastive learning~(CCL) module, comprising the adaptive fusion module and pixel-to-anchor~(PA) loss $L_\mathrm{PA}$. (b) AA loss with a memory bank, where $\mathcal{S}(\cdot)$ is the selection function to generate class-prioritized anchors. (c) Boundary-aware negative~(BANE) sampling, which selects informative negative samples along prediction error boundaries for use in PA loss. $I_\text{Batch}$ denotes mini-batch images, $\hat{Y}$ the predicted segmentation output, $\featureset_\layerIdx$ the feature representation at the $\layerIdx$-th layer, $\embeddingset_\layerIdx$ the transformed feature set via encoding function $\pi(\cdot)$, $\anchors_\layerIdx$ the representative anchors derived from $\embeddingset_\layerIdx$, $\hat{\anchors}_\layerIdx$ the updated anchors after adaptive fusion, and $L_\text{CE}$ the cross-entropy loss. Semantic classes are distinguished using various shapes and colors~(best viewed in color).}
    % \caption{Overview of the proposed pipeline, called \textit{Contextrast++}.
    % \textbf{(a)}~The contextual contrastive learning~(CCL) module, which consists of the adaptive fusion module and pixel-to-anchor~(PA) loss~$L_\mathrm{PA}$ (See Sec.~\ref{sec:adaptive} and Sec.~\ref{sec:pa}).
    % \textbf{(b)}~AA loss with memory bank. $\mathcal{S}(c)$ is the selection function to generate class-prioritized anchor (See Sec.~\ref{sec:aa}).
    % \textbf{(c)}~Boundary-aware negative~(BANE) sampling, which selects more informative negative samples located along the boundaries of prediction error regions (See Sec.~\ref{sec:boundary-aware}).
    % These sampled embedding vectors are used as negative samples in PA loss. 
    % $\mathbf{I}_\text{Batch}$ is the images of mini-batch. $\segPred$ denotes the model's predicted segmentation output. $\featureset_\layerIdx$ refers to the feature representation obtained at the $\layerIdx$-th layer.
    % After applying the encoding function~$\pi(\cdot)$, the transformed feature set at this layer is represented as $\embeddingset_\layerIdx$. $\anchors_\layerIdx$ denotes the representative anchors derived from $\embeddingset_\layerIdx$. These anchors are further refined by the adaptive fusion module, yielding the updated anchors $\hat{\anchors}_\layerIdx$. The cross-entropy loss, used to supervise the segmentation task, is denoted as $L_{\text{CE}}$. For visualization, semantic classes are distinguished using various shapes and colors~(best viewed in color).}
    \label{fig:pipeline}
\end{figure*}

\subsection{Overall framework}
We introduce a supervised contrastive learning framework, termed Contextrast++, and visually summarize it in Fig.~\ref{fig:pipeline}. Contextrast++ incorporates several methods specifically designed for supervised semantic segmentation.

First, for the reader's convenience, we briefly revisit the preliminaries of the CCL framework in \Cref{sec:preliminaries}.
We then describe how the previously proposed static fusion module and PA loss are enhanced in Contextrast++~(\Cref{sec:adaptive} and \Cref{sec:pa}) and AA loss with a memory bank (\Cref{sec:aa}), respectively.
In particular, we explain how the AA loss with a memory bank effectively addresses the long-tailed distribution issue~\cite{li2022targeted} in semantic segmentation tasks.
Second, the boundary-aware negative (BANE) sampling method, which was originally proposed in Contextrast~\cite{sung2024contextrast},
is summarized in \Cref{sec:boundary-aware} to briefly explain its role in guiding the model's attention toward challenging boundary regions.

\subsection{Contextual contrastive learning (CCL)}
\label{sec:ccl}

\subsubsection{Overview of CCL}
The CCL framework optimizes the class representations in the embedding space, strengthening the encoder's discriminative power without additional inference overhead. It comprises three mutually reinforcing components, each addressing a distinct challenge in contrastive learning for semantic segmentation. First, the adaptive fusion module overcomes the limited expressiveness of the static fusion in Contextrast~\cite{sung2024contextrast} by adaptively integrating multi-scale anchors. Second, the PA loss resolves the representation switching problem of conventional contrastive learning objectives, in which each sample alternately serves as an anchor, causing inconsistent optimization, by fixing the fused anchor as a stable class-level reference. Third, the AA loss mitigates long-tailed class imbalance within mini-batches by leveraging class-prioritized anchors in a memory bank for stable, class-balanced supervision. Together, these components operate synergistically, where context-enriched anchors from the adaptive fusion module amplify the effectiveness of the PA loss, and the AA loss extends supervision across the class distribution.

\subsubsection{Preliminaries: Embedding and representative anchors}\label{sec:preliminaries}
We consider an encoder architecture comprising $\layerTotal$ layers.
As shown in Fig.~\ref{fig:pipeline}, given a mini-batch of images $\mathbf{I}_\text{Batch}$ with spatial resolution $H \times W$, the $\layerIdx$-th layer, where $\layerIdx \in \{1, \cdots, \layerTotal\}$, has feature map $\featureset_\layerIdx$ whose height and width are $H_\layerIdx$ and $W_\layerIdx$, respectively, satisfying $H_\layerIdx < H$ and $W_\layerIdx < W$.
Note that $\featureset_\layerIdx$ at deeper layers has smaller height and width, \ie $H_{\layerIdx-1} > H_{\layerIdx}$ and $W_{\layerIdx-1} > W_{\layerIdx}$, for $\layerIdx \in \{2, \cdots, \layerTotal\}$. 
Next, taking $\featureset_\layerIdx$ as an input, a projection head $\pi(\cdot)$ outputs the $\layerIdx$-th embedding feature vector set~$\embeddingset_\layerIdx$,~\ie $\embeddingset_\layerIdx = \pi(\featureset_\layerIdx)$,
where each pixel location corresponds to a feature vector $\embeddingvec \in \mathbb{R}^d$, where $d$ is the channel dimension of the embedding space.

By denoting the total number of classes in a dataset as $N$, the number of class IDs present in each mini-batch is denoted by $N_\layerIdx$, where $N_\layerIdx \leq N$.
We define the set of representative anchors at the $\layerIdx$-th encoder layer as $\anchors_\layerIdx$,
which consists of $\classTotal_\layerIdx$ class-wise representative anchors, and each anchor $\anchor^\classIdx_\layerIdx \in \mathbb{R}^d$ in $\anchors_\layerIdx$ represents the mean of the embedded feature vectors associated with the ground truth labels in the mini-batch. 

To compute each $\anchor^\classIdx_\layerIdx$, the set of embedding vectors whose spatial locations correspond to pixels labeled with class $\classIdx$ in the ground truth label map $\segGT_\layerIdx$, is required.
To this end, let $\segGT$ be the set of ground truth semantic labels in the current mini-batch and $\segGT_\layerIdx$ be the downsampled ground truth labels aligned with the spatial resolution of $\embeddingset_\layerIdx$, \ie $\segGT_\layerIdx$ and $\embeddingset_\layerIdx$ share the same width and height.
We define $\gtfunc(\embeddingvec)$ as a mapping function that retrieves the ground truth semantic label from $\segGT_\layerIdx$, which spatially aligns with the embedding vector $\embeddingvec \in \embeddingset_\layerIdx$.
Next, let $\mathcal{\classTotal}_\layerIdx$ be the set of unique class labels present in $\segGT_\layerIdx$, where $|\mathcal{\classTotal}_\layerIdx| = N_\layerIdx \leq N$.
Then, $\anchor^\classIdx_\layerIdx$ is defined as follows:

\begin{align}
    \anchor^\classIdx_\layerIdx = \frac{\sum\limits_{\embeddingvec \in \embeddingset_\layerIdx} \embeddingvec \mathbbm{1}[g(\embeddingvec) = \classIdx]}{\sum\limits_{\embeddingvec \in \embeddingset_\layerIdx} \mathbbm{1}[g(\embeddingvec) = \classIdx]}, \text{where}\; \classIdx \in \mathcal{\classTotal}_\layerIdx.
    \label{eq:anchor}
\end{align}
 
Here, $\mathbbm{1}[\cdot]$ denotes the Iverson bracket, which returns one when the given condition is satisfied, and zero otherwise. By using~\eqref{eq:anchor}, $\anchors_\layerIdx$ is expressed as $\anchors_\layerIdx = \{\anchor^\classIdx_\layerIdx \mid \classIdx \in \mathcal{\classTotal}_\layerIdx\}$.
For convenience, we interchangeably express $\anchors_\layerIdx$ in a matrix form, \ie $\anchors_\layerIdx = [\anchor^{n_1}_\layerIdx \; \anchor^{n_2}_\layerIdx \; ... \;  \anchor^{n_{\classTotal_\layerIdx}}_\layerIdx] \in \mathbb{R}^{d \times \classTotal_\layerIdx}$, where $\{n_1, \dots, n_{N_\layerIdx}\} = \mathcal{N}_\layerIdx$.

\begin{figure}[t]
    \centering
    \includegraphics[scale=0.4]{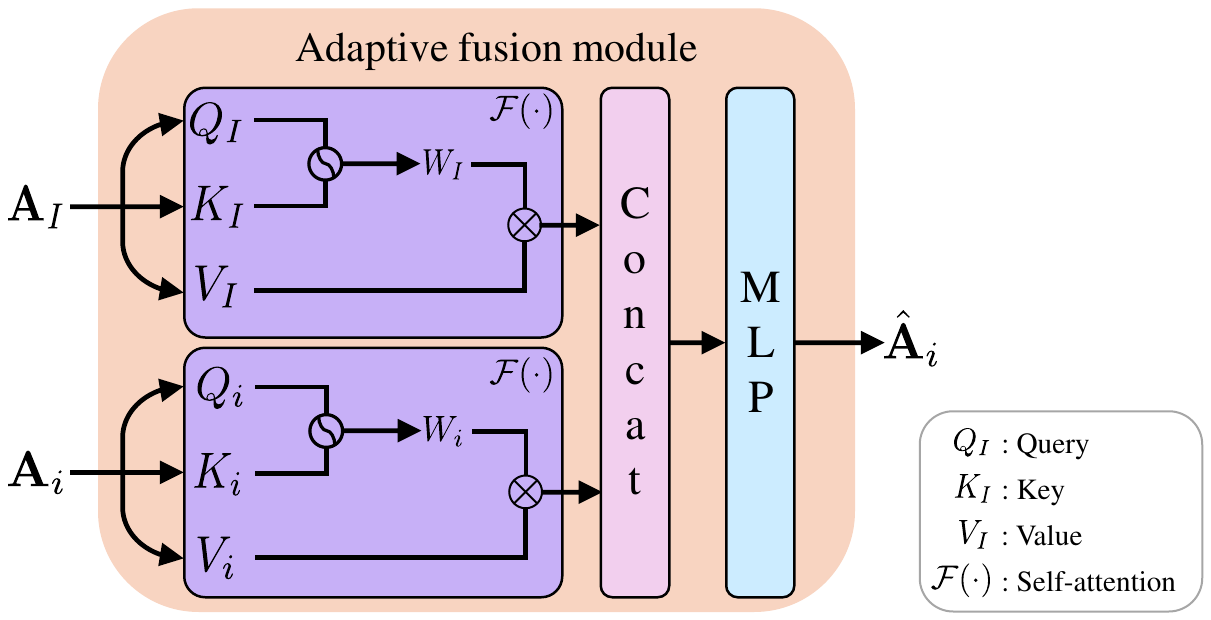}
    \caption{The pipeline of the proposed adaptive fusion module with the deepest-layer representative anchor $\anchors_I$ and earlier-layer representative anchor $\anchors_i$. $Q, K,$ and $V$ represent query, key, and value, respectively. The purple box $\mathcal{F}(\cdot)$ indicates self-attention~\cite{vaswani2017attention}.}
    \label{fig:adaptive_fusion_module}
\end{figure}

\subsubsection{Adaptive fusion module}
\label{sec:adaptive}
In our previous work Contextrast~\cite{sung2024contextrast}, multi-scale representative anchors were fused using a fixed weighted sum with the deepest layer anchor as follows:
\begin{align}
    \hat{\anchor}_\layerIdx^\classIdx =
    \begin{cases}
        w_l \anchor_\layerIdx^\classIdx + w_h \anchor_{\layerTotal}^\classIdx, & \text{if}\, \classIdx \in \mathcal{\classTotal}_\layerIdx \cap \mathcal{\classTotal}_\layerTotal\\
        \anchor_\layerIdx^\classIdx, & \text{Otherwise}
    \end{cases},
\label{eq:weighted_sum}
\end{align}
where $w_l$ and $w_h$ are weighting hyperparameters for anchor update, satisfying $w_l + w_h = 1$. This formulation injects global semantic context into early-layer anchors, but its reliance on fixed hyperparameters limits the capability to adaptively balance semantic context and fine-grained details. More critically, element-wise summation forces a linear superposition of features, preventing the model from capturing richer cross-scale interactions.

To address these limitations, we adopt self-attention and MLP to construct our adaptive fusion module, following a two-stage \textit{refine-then-fuse} strategy. This design is particularly suited to our setting: multi-scale anchors from different encoder layers encode semantics at varying levels of abstraction, where each anchor's channel dimensions carry scale-specific semantic attributes. Unlike context modeling approaches~\cite{li2021ctnet}, clustering~\cite{liu2013weakly, zhu2025merging} and multi-scale fusion modules~\cite{zhou2024boundary, gu2022multi, shi2023transformer, fu2025segman, shi2025llmformer, yang2025m} that operate on spatial feature maps for feature enhancement, our adaptive fusion module operates on compact class-level anchor embeddings in the contrastive embedding space. Specifically, as shown in Fig.~\ref{fig:adaptive_fusion_module}, each anchor is first refined independently using self-attention $\mathcal{F}(\cdot)$ to model inter-channel dependencies, emphasizing discriminative feature channels critical for semantic segmentation. Finally, the refined anchors from layers $\layerIdx$ and $\layerTotal$ are concatenated and fused through an MLP:
\begin{align}
    \hat{\anchor}^\classIdx_\layerIdx =
    \begin{cases}
        \text{MLP}(\mathcal{F}(\anchor_\layerTotal^\classIdx) \oplus \mathcal{F}(\anchor_\layerIdx^\classIdx)) & \text{if}\; \classIdx \in \mathcal{\classTotal}_\layerIdx \cap \mathcal{\classTotal}_\layerTotal \\
        \anchor^\classIdx_\layerIdx &  \text{Otherwise}
    \end{cases},
\label{eq:adaptive_fusion}
\end{align}
where $\oplus$ represents concatenation. Crucially, concatenation preserves all scale-specific representations, avoiding the information loss inherent in summation. Subsequently, the MLP learns non-linear interactions between these refined anchors, producing a more expressive representation. Notably, these improvements are achieved with negligible computational overhead owing to the lightweight structure of the module.

\subsubsection{PA loss}
\label{sec:pa}
Next, we describe how the PA loss is formulated with the adaptive fusion module.
Before that, we briefly introduce the baseline of the PA loss, InfoNCE loss~\cite{gutmann2010noise,oord2018representation}, and highlight the key differences.
Conventional contrastive learning-based methods utilize InfoNCE loss~\cite{gutmann2010noise,oord2018representation}, which is defined as follows:
\begin{align}
    L_{\mathrm{NCE}}\!=\!-\frac{1}{|\embeddingset_{+}|}\!\sum\limits_{\embeddingvec_{+}\!\in \embeddingset_{+}}\!\log\!\frac{\exp(\embeddingvec\!\cdot\! \embeddingvec_+ / \tau)}{\exp(\embeddingvec\!\cdot\! \embeddingvec_+/ \tau)+\!\sum\limits_{\embeddingvec_{-}\!\in\embeddingset_-}\!\exp(\embeddingvec\!\cdot\! \embeddingvec_- / \tau)},
\label{eq:info_nce}
\end{align}
where $\embeddingset_{+/-}$ represents positive and negative samples, respectively. The positive samples $\embeddingset_{+}$ belong to the same semantic class as the anchor $\embeddingvec$, while the negative samples~$\embeddingset_{-}$ belong to a different class. Here, the logit similarities are computed using the dot product between samples and scaled by the temperature hyperparameter $\tau$. The objective is to reduce the distance between $\embeddingvec_{+} \in \embeddingset_{+}$ and anchor $\embeddingvec$, while maximizing the separation between $\embeddingvec_{-} \in \embeddingset_{-}$ and $\embeddingvec$. In~\eqref{eq:info_nce}, since each sample is used once as an anchor $\embeddingvec$, the objective in the contrastive learning changes during the loss computation.

To avoid this switching problem, we proposed the pixel-to-anchor~(PA) loss~\cite{sung2024contextrast}, $L_\mathrm{PA}$, by reforming~\eqref{eq:info_nce} as follows:
\begin{align}
    L_{\mathrm{PA}} = \sum\limits_{\layerIdx=1}^{\layerTotal}\lambda_\layerIdx \biggl[\frac{1}{\classTotal_\layerIdx}\sum_{\hat{\anchor}^\classIdx_\layerIdx\in\hat{\anchors}_\layerIdx} L_\layerIdx\biggr],
    \label{eq:pixel-anchor}
\end{align}
\begin{align}
    L_\layerIdx\!=\!-\frac{1}{|\embeddingset_+|}\!\sum_{\embeddingvec_+\in\embeddingset_+}\!\log\!\frac{\exp(\hat{\anchor}^\classIdx_\layerIdx\!\cdot\!\embeddingvec_+/\tau)}{\exp(\hat{\anchor}^\classIdx_\layerIdx\!\cdot\!\embeddingvec_+/\tau)\!+\! \!\sum\limits_{\embeddingvec_-\in\embeddingset_-}\!\exp(\hat{\anchor}^\classIdx_\layerIdx\!\cdot\!\embeddingvec_-/\tau)},
    \label{eq:pa-for-each-i}
\end{align}
where $\lambda_\layerIdx$ denotes the weighting hyperparameters for the PA loss at the $\layerIdx$-th layer. The logit similarities are computed as the dot product between the anchor $\hat{\anchor}$ and the samples $\embeddingvec_{+/-}$, scaled by the temperature hyperparameter $\tau$.

By doing so, the PA loss optimizes embeddings by pulling positive samples toward the fused anchor $\hat{\anchor}_\layerIdx^\classIdx$ and pushing negative samples away from it.
This also eliminates representation switching during training and improves stability and class-level consistency.

\begin{figure}[t]
    \centering
    \includegraphics[scale=0.44]{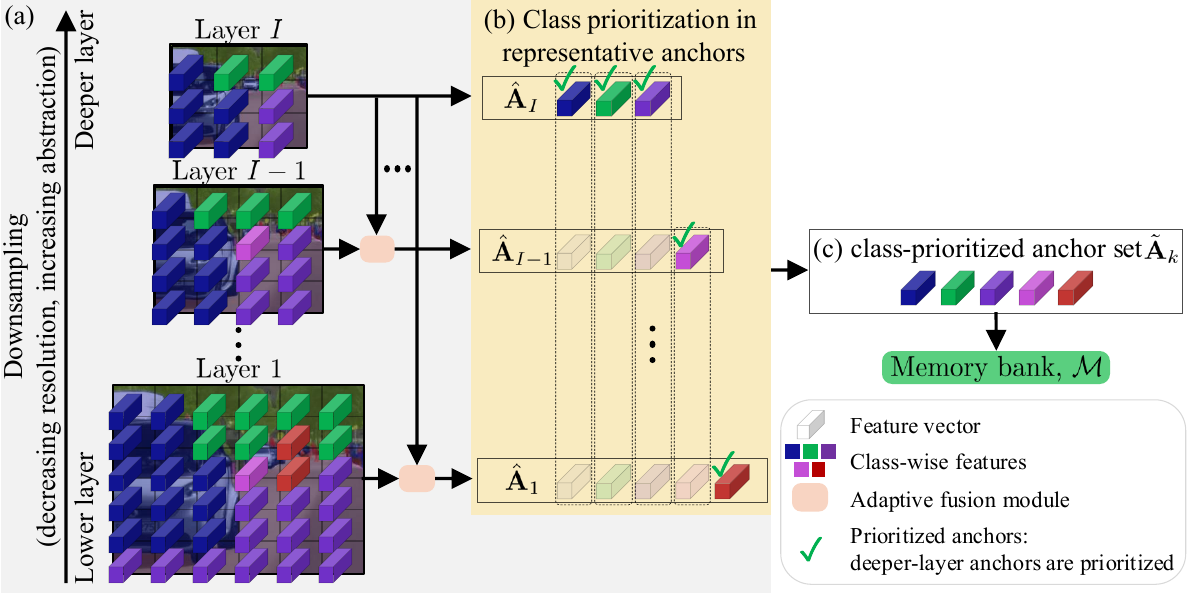}
    \caption{Illustration of the class-prioritized anchor selection and storage process in the memory bank. (a) Features in each scale may contain different sets of classes due to downsampling the ground truth to match the reduced spatial resolution of the features. Thus, some small objects (\eg red and pink features here) tend to vanish in deeper layers, whereas large objects such as cars and roads persist across all layers. (b) Class-prioritized anchors are selected from the deepest available layer where each class appears (green check marks).
    (c) The selected anchors are aggregated to form a class-prioritized anchor set, which is stored in the memory bank.}
    \label{fig:class-prioritized-anchor}
\end{figure}

\subsubsection{AA loss with a memory bank}
\label{sec:aa}
While the fused representative anchors $\hat{\anchors}_i$ used in the PA loss operate within the scope of a mini-batch, a long-tailed distribution issue~\cite{li2022targeted} can arise if certain class features are absent from the mini-batch. These limitations motivate the use of a memory bank that provides more stable and balanced supervision.

However, na\"ively adopting memory bank designs from classification approaches~\cite{he2020momentum, khosla2020supervised} is structurally incompatible with multi-scale semantic segmentation. Specifically, the standard practice of storing only the deepest layer anchors omits classes that disappear at lower resolutions, resulting in incomplete class coverage. Conversely, storing anchors from all layers introduces a scale-induced bias, where the memory bank becomes overwhelmingly dominated by fine-grained features.

First, as shown in Fig.~\ref{fig:class-prioritized-anchor}(a), deeper layers lose more spatial information owing to downsampling. As a result, smaller objects fail to produce an embedding vector via 
$g(\embeddingvec)$, causing the network to focus primarily on larger objects in the image.
Formally, this downsampling naturally leads $|\mathcal{N}_{\layerIdx - 1}| \geq |\mathcal{N}_{\layerIdx}|$ for $\layerIdx \in \{2, \cdots, \layerTotal\}$, 
which causes the deeper layer anchor $\anchor^\classIdx_\layerIdx$ to become nonexistent as $\mathbbm{1}[g(\embeddingvec) = \classIdx]$ in \eqref{eq:anchor} becomes zero. Second, due to the substantially higher spatial resolution of earlier layers, the memory bank is disproportionately filled with fine-grained features, marginalizing the semantic features necessary for global context.
To address these issues, we introduce the AA loss that utilizes \textit{class-prioritized anchors} stored in a memory bank $\mathcal{M}$.

\newcommand{\classid}{c}
The class-prioritized anchors are defined as a set of class-specific representative anchors selected from the deepest available layer where each class appears.
As explained above, the first layer contains the richest spatial information and therefore includes the largest number of existing classes.
Thus, we first define the set of all existing class IDs as $\mathcal{\classTotal}^* \leftarrow \mathcal{\classTotal}_1$.
Next, let us define the deepest layer index that contains the class $\classid \in \mathcal{\classTotal}^*$ as follows:
\begin{equation}
  \layerIdx^*(\classid) = \mathrm{max}\Big\{i \in \big\{1, \cdots, \layerTotal\big\} \; | \; \classid \in \mathcal{\classTotal}_\layerIdx\Big\}.
\label{eq:highest_index}
\end{equation}
Using the definition in \eqref{eq:highest_index}, we define the selection function $\mathcal{S}(\classid)$ that retrieves the representative anchor from the deepest available layer as:
\begin{equation}
    \mathcal{S}(\classid) = \anchor^\classid_{\layerIdx^*(\classid)},
\label{eq:class-prioritized}
\end{equation}
which corresponds to the green check-marked features in Fig.~\ref{fig:class-prioritized-anchor}(b).
Subsequently, as presented in Fig.~\ref{fig:class-prioritized-anchor}(c), the class-prioritized anchor set $\tilde{\anchors}_k$ at the $k$-th training iteration is defined as follows:
\begin{equation}
  \tilde{\anchors}_k = \left\{\mathcal{S}(\classid) \;\middle|\; \classid \in \mathcal{\classTotal}^* \right\}.
\end{equation}

For example, in Fig.~\ref{fig:class-prioritized-anchor}, when the same class representative anchors exist in multiple layers, the one from the deeper layer is selected according to~\eqref{eq:class-prioritized}. 
%Consequently, in $\tilde{\anchors}_k$, the representative anchors for the car, vegetation, and road classes are taken from $\hat{\anchors}_\layerTotal$, while the representative anchor for the person class is taken from $\hat{\anchors}_1$.
Consequently, in $\tilde{\anchors}_k$, representative anchors for classes occupying large spatial regions, \eg car, vegetation, and road classes, are taken from deeper layers, whereas those for classes with small spatial extent, \eg person class, are taken from shallower layers. %~$\hat{\anchors}_1$.

Then, the class-prioritized anchors from each training iteration are stored in a memory bank $\mathcal{M}$, which consists of $\classTotal$ class-wise queues. 
Each queue stores up to~$K_\mathcal{M}$ class-prioritized anchors,~\ie $\classTotal\!\times\! K_\mathcal{M}$ anchors in total. When the number of stored anchors exceeds $K_\mathcal{M}$, the oldest class-prioritized anchor is removed, and the newest one is added. The hyperparameter $K_\mathcal{M}$ controls the memory bank size, balancing storage capacity and memory efficiency.

Then, we sample an equal number of class-prioritized anchors for each class to mitigate the long-tailed distribution issue.
These sampled anchors are then utilized in the AA loss, $L_\mathrm{AA}$, where logit similarities are computed as dot products scaled by temperature $\tau$:
\begin{align}
  L_\mathrm{AA} = \frac{1}{|\mathcal{\classTotal}^*|} \sum_{\classid \in \mathcal{\classTotal}^*}L_a^\classid,
    \label{eq:anchor-anchor}
\end{align}

\begin{align}
    L_a^\classid\!=\!-\frac{1}{K_\mathcal{M}}\!\sum_{\tilde{\anchor}_+ \in \mathcal{M}}\!\log\frac{\exp(\tilde{\anchor}^\classid\!\cdot\tilde{\anchor}_+/\tau)}{\exp(\tilde{\anchor}^\classid\!\cdot\tilde{\anchor}_+/\tau)\!+\!\sum\limits_{\tilde{\anchor}_-\in \mathcal{M}}\!\exp(\tilde{\anchor}^\classid\!\cdot\tilde{\anchor}_-/\tau)},
    \label{eq:loss_a}
\end{align}
where $\tilde{\anchor}^{\classid}$, $\tilde{\anchor}_+$, $\tilde{\anchor}_-$, and $\tau$ represent the class-prioritized anchor for class $\classid \in \mathcal{N}^*$, positive class-prioritized anchor, negative class-prioritized anchor, and the temperature hyperparameter, respectively. By using an equal number of sampled anchors for each class, the AA loss effectively addresses the long-tailed distribution problem. In addition, leveraging class-prioritized anchors in the memory bank provides memory efficiency by eliminating the need to store a large number of individual features for each class. As representative anchors inherently encode feature information from multiple samples, this approach contributes to a compact yet expressive feature storage, facilitating stable and class-balanced training.

\subsubsection{Loss function}
\label{sec:loss}
Finally, the losses $L_\mathrm{PA}$ in~\eqref{eq:pixel-anchor} and $L_\mathrm{AA}$ in~\eqref{eq:anchor-anchor} are jointly optimized alongside the standard pixel-wise cross-entropy loss~$L_{\text{CE}}$~\cite{sun2019high}, offering a complementary supervision that enhances segmentation performance. 
Through this design, while $L_\mathrm{CE}$ guides the model to predict the correct class label for each sample, $L_\mathrm{PA}$ efficiently learns relationships between global-local contexts by applying adaptively fused representative anchors, and $L_\mathrm{AA}$ helps to solve the long-tailed distribution issue by sampling an equal number of class-prioritized anchors from the memory bank.

Formally, the framework is designed to optimize the following loss function:
\begin{equation}
    L = L_{\mathrm{CE}} +  \alpha (L_{\mathrm{PA}} + L_{\mathrm{AA}}),
    \label{eq:objective}
\end{equation}
where $\alpha$ represents the hyperparameter weight for our contrastive learning loss.

\subsection{Boundary-aware negative~(BANE) sampling}
\label{sec:boundary-aware}
In addition to CCL, we enhance the loss by incorporating BANE sampling, an effective hard negative sampling strategy previously proposed in Contextrast~\cite{sung2024contextrast}.
Following previous works~\cite{kalantidis2020hard,wang2021exploring,zhou2024cross}, which emphasize the importance of leveraging informative negatives in contrastive learning,
BANE sampling selects negative samples near prediction boundaries because they are harder to distinguish. 
% Consequently, sampled harder negatives, into~\eqref{eq:pa-for-each-i} improves the model's capability to represent features from different classes more discriminatively.
% This improves the model’s ability to learn more discriminative features.
% BANE sampling to better identify informative hard negative samples $\embeddingset_-$. Embedding vectors located near the boundaries are considered informative negative samples because they are more likely to be challenging to distinguish from negative samples.
Consequently, using harder negative samples as $\embeddingset_-$ in~\eqref{eq:pa-for-each-i} improves the model's capability to learn more class-discriminative features.

As depicted in Fig.~\ref{fig:ext_sampling}, BANE sampling consists of three sequential steps:~(i)~decomposing the prediction output into class-wise binary error maps, (ii)~applying the distance transform~\cite{kimmel1996sub} to the class-wise binary error maps, and (iii)~identifying negative samples according to the computed distance maps.

First, we begin by defining the class-specific binary error map~$\binaryImg^\classIdx_\layerIdx$ at each pixel location $(u,v)$ as follows: 
\begin{equation}
\binaryImg^\classIdx_\layerIdx(u,v)=\mathbbm{1}[(\hat{y}_\layerIdx \neq \classIdx) \land (\gtfunc(\hat{{y}}_\layerIdx) = \classIdx)], \text{where}\; \classIdx \in \mathcal{\classTotal}_\layerIdx,
\end{equation}
where $\hat{y}_\layerIdx$ represents the predicted label at the $\layerIdx$-th layer, obtained by downsampling the final-layer prediction. $\gtfunc(\cdot)$ denotes a labeling function used in~\eqref{eq:anchor}. Each pixel in~$\binaryImg^\classIdx_\layerIdx$ has a value of one for the incorrectly predicted pixel~(\ie~a negative sample), and zero otherwise~(in Fig.~\ref{fig:ext_sampling}(a)).

Subsequently, the binary error map $\binaryImg^\classIdx_\layerIdx$ is converted into a class-wise distance map $\distImg^\classIdx_\layerIdx$ using the Euclidean distance transform~\cite{kimmel1996sub} for each class $\classIdx$ and each layer $\layerIdx$. The pixel value in $\distImg^\classIdx_\layerIdx$ corresponds to the minimal distance between the pixel $(u,v)$ and the edge pixels~$\edgeSet^\classIdx_\layerIdx$ from $\binaryImg^\classIdx_\layerIdx$, with the constraint that only misclassified pixels~(\ie $\binaryImg_\layerIdx^\classIdx(u,v)=1$) are utilized. 

Formally, each pixel value of distance map for each $n$-th class, $\distImg^\classIdx_\layerIdx(u,v)$, is defined as follows:
\begin{equation}
    \distImg^\classIdx_\layerIdx(u,v) \!=\! \mathop{\mathrm{min}}_{(x,y)\in \mathbf{E}^\classIdx_\layerIdx}\!\sqrt{(u\!-\!x)^2\!+\!(v\!-\!y)^2}, \text{s.t.}~\binaryImg^\classIdx_\layerIdx(u,v)\!=\!1.
\end{equation}
This indicates that in the misclassified regions, where the pixel value of $\binaryImg^\classIdx_\layerIdx$ is one,~\ie white regions in Fig.~\ref{fig:ext_sampling}(a), a smaller value in $\distImg^\classIdx_\layerIdx$ indicates that the corresponding pixel is more likely to be located near the object's boundary,~\ie pixels corresponding to the boundary highlighted in blue inside the rightmost zoomed box in Fig.~\ref{fig:ext_sampling}(a).

Finally, from the regions where $\binaryImg^\classIdx_\layerIdx\!=\!1$, we select embedding vectors corresponding to the lowest $K_s$ percentile of distances in $\distImg^\classIdx_\layerIdx$, where $K_s$ is a hyperparameter that uniformly controls the selection ratio across all encoder layers. These selected embeddings serve as harder negative samples for each $n$-th representative anchor. By integrating these more challenging negative samples into~\eqref{eq:pa-for-each-i}, Contextrast++ enhances contrastive learning by encouraging feature representations to move closer to the anchor of the true class while increasing their distance from the anchors of incorrect classes. Consequently, these boundary-aware negative samples contribute to a more accurate modeling of inter-class spatial dependencies, thereby enhancing the network's discriminative power against negative samples.

\begin{figure}[t]
    \centering
    \includegraphics[scale=0.3]{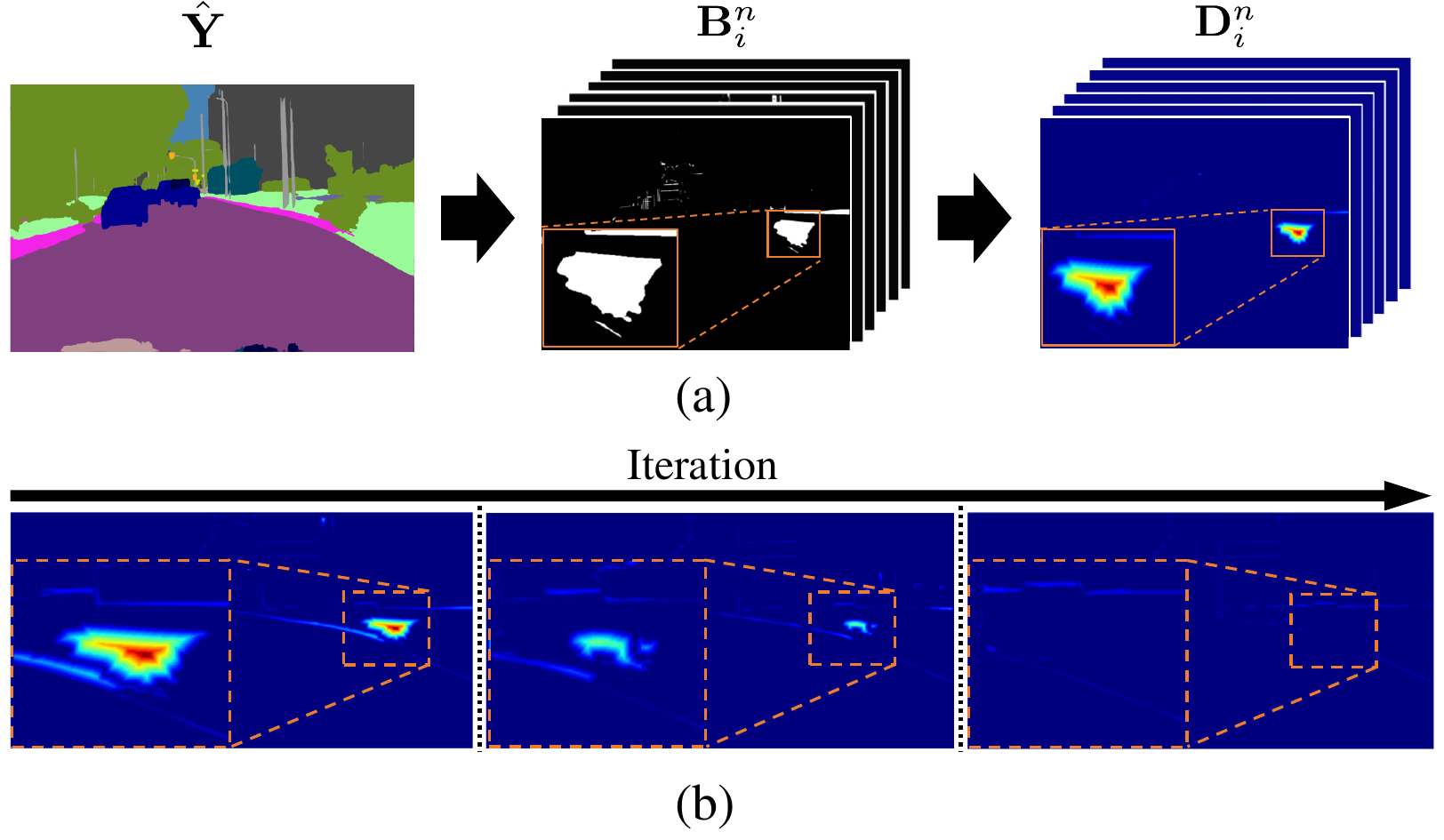}
    \caption{Overview of the boundary-aware negative sampling strategy, illustrating its effectiveness in mitigating under- and over-segmentation during training. (a) The prediction map $\segPred$ is decomposed into class-wise binary masks $\mathbf{B}^\classIdx_\layerIdx$. Then, the distance transform~\cite{kimmel1996sub} is applied to generate class-wise distance maps $\mathbf{D}^\classIdx_\layerIdx$. (b)~Visualization of the distance map progression across training iterations. Misclassified regions gradually diminish as training advances~(best viewed in color).}
    \label{fig:ext_sampling}
\end{figure}

    \begin{table*}[t!]
    \caption{Details of the experimental setup and the baseline semantic segmentation models for contrastive learning across datasets.}
      \centering
      \resizebox{0.99\textwidth}{!}{%
      \begin{tabular}{c|cc|cccccccc}
          \hline
          \rowcolor[HTML]{DAE8FC}
                                      & \multicolumn{2}{c|}{Method} & \multicolumn{8}{c}{Training Settings}                                                                              \\ [-0.4pt] \hhline{~----------}
                                      \rowcolor[HTML]{DAE8FC}
                                      \multirow{-2}{*}{Dataset} & Model        & Backbone     & Crop size  & Learning rate & Momentum & Weight decay & Optimizer & Lr scheduler & Batch size & Training steps \\ \hline
          \multirow{7}{*}{Cityscapes} & DeepLabV3~\cite{chen2017rethinking}    & D-ResNet-101 & 512$\times$1024 & $10^{-2}$               & 0.9      & 5$\times$$10^{-4}$     & SGD       & Poly         & 8          & 40K            \\
                                      & HRNet~\cite{sun2019high}        & HRNetV2-W48  & 512$\times$1024 & $10^{-2}$               & 0.9      & 5$\times$$10^{-4}$     & SGD       & Poly         & 8          & 40K            \\
                                      & OCRNet~\cite{yuan2020object}       & HRNetV2-W48  & 512$\times$1024 & $10^{-2}$               & 0.9      & 5$\times$$10^{-4}$     & SGD       & Poly         & 8          & 40K            \\
                                      & UPerNet~\cite{xiao2018unified}      & Swin-Tiny       & 512$\times$1024 & 6$\times$$10^{-5}$           & 0.9      & $10^{-2}$        & ADAMW     & Linear       & 6          & 40K            \\
                                      & MobileNetV2~\cite{sandler2018mobilenetv2}  & D-ResNet-101 & 512$\times$1024 & $10^{-2}$               & 0.9      & $10^{-2}$         & SGD       & Poly         & 8          & 80K            \\ 
                                      & Mask2Former~\cite{cheng2022masked} & Swin-Tiny
                                           & 512$\times$1024 & $10^{-4}$ & 0.9 & $5\times10^{-2}$ & ADAMW     &  Poly & 8 & 90K \\ 
                                      & SegFormer~\cite{xie2021segformer} & MiT-B0
                                          & 1024$\times$1024 & 6$\times10^{-5}$      & 1.0  
                                          & $10^{-2}$      & ADAMW      & Linear  & 8 & 160K \\ \hline
          \multirow{3}{*}{ADE20K}     & DeepLabV3~\cite{chen2017rethinking}    & D-ResNet-101 & 512$\times$512  & $10^{-2}$              & 0.9      & 5$\times$$10^{-4}$     & SGD       & Poly         & 12         & 80K            \\
                                      & HRNet~\cite{sun2019high}        & HRNetV2-W48  & 512$\times$512  & $10^{-2}$              & 0.9      & 5$\times$$10^{-4}$     & SGD       & Poly         & 12         & 80K            \\
                                      & OCRNet~\cite{yuan2020object}       & HRNetV2-W48  & 512$\times$512  & $10^{-2}$              & 0.9      & 5$\times$$10^{-4}$     & SGD       & Poly         & 12         & 80K            \\ \hline
          \multirow{3}{*}{PASCAL-C}   & DeepLabV3~\cite{chen2017rethinking}    & D-ResNet-101 & 512$\times$512  & $10^{-2}$               & 0.9      & 5$\times$$10^{-4}$      & SGD       & Poly         & 16         & 60K            \\
                                      & HRNet~\cite{sun2019high}        & HRNetV2-W48  & 512$\times$512  & $10^{-2}$               & 0.9      & 5$\times$$10^{-4}$      & SGD       & Poly         & 16         & 60K            \\
                                      & OCRNet~\cite{yuan2020object}       & HRNetV2-W48  & 512$\times$512  & $10^{-2}$               & 0.9      & 5$\times$$10^{-4}$      & SGD       & Poly         & 16         & 60K            \\ \hline
          \multirow{3}{*}{COCO-Stuff} & DeepLabV3~\cite{chen2017rethinking}    & D-ResNet-101 & 512$\times$512  & $10^{-2}$               & 0.9      & 5$\times$$10^{-4}$      & SGD       & Poly         & 16         & 60K            \\
                                      & HRNet~\cite{sun2019high}        & HRNetV2-W48  & 512$\times$512  & $10^{-2}$               & 0.9      & 5$\times$$10^{-4}$      & SGD       & Poly         & 16         & 60K            \\
                                      & OCRNet~\cite{yuan2020object}       & HRNetV2-W48  & 512$\times$512  & $10^{-2}$               & 0.9      & 5$\times$$10^{-4}$      & SGD       & Poly         & 16         & 60K            \\ \hline
          \multirow{3}{*}{CamVid}     & DeepLabV3~\cite{chen2017rethinking}    & D-ResNet-101 & 360$\times$480  & 2$\times$$10^{-2}$           & 0.9      & 5$\times$$10^{-4}$     & SGD       & Poly         & 16         & 6K             \\
                                      & HRNet~\cite{sun2019high}        & HRNetV2-W48  & 360$\times$480  & 2$\times$$10^{-2}$           & 0.9      & 5$\times$$10^{-4}$     & SGD       & Poly         & 16         & 6K             \\
                                      & OCRNet~\cite{yuan2020object}       & HRNetV2-W48  & 360$\times$480  & 2$\times$$10^{-2}$             & 0.9      & 5$\times$$10^{-4}$     & SGD       & Poly         & 16         & 6K             \\ \hline
          \end{tabular}%
      }
      \label{tab:implementation}
\end{table*}
\section{Experiment} \label{sec:exp}
\subsection{Experimental setup}
\label{sec:exp-setup}
\noindent\textbf{Datasets.} We evaluate on five public datasets. Cityscapes~\cite{cordts2016cityscapes} contains 5,000 urban scene images (2,975/500/1,525 train/val/test) annotated with 19 semantic classes. ADE20K~\cite{zhou2017scene} contains 20,210 train and 2,000 validation images across 150 semantic categories. PASCAL-C~\cite{mottaghi2014role} contains 4,998 train and 5,105 test images across 59 classes. COCO-Stuff~\cite{caesar2018coco} contains 9,000 train and 1,000 test images, annotated with 171 classes comprising 80 object and 91 stuff categories. CamVid~\cite{brostow2009semantic} contains 367 train, 101 validation, and 233 test images across 11 classes. All datasets exhibit long-tailed class distributions, where ADE20K, PASCAL-C, and COCO-Stuff present more severe imbalance due to their larger number of semantic categories and greater class diversity.

\vspace{2mm}
% \noindent{\bf Training settings.} To evaluate the effectiveness of our proposed method, we adopted several baseline architectures as listed in Table~\ref{tab:implementation}. All experiments were carried out using MMSegmentation~\cite{mmseg2020}, which is a popular open-source semantic segmentation toolbox.
% We used the same hyperparameter settings across all models and initialized the networks with ImageNet~\cite{deng2009imagenet} pre-trained weights, while the remaining layers were randomly initialized. Data augmentation techniques, including color jittering, horizontal flipping, and random scaling, were applied during training. Further details of training settings are described in Table~\ref{tab:implementation}.
\noindent\textbf{Training settings.} All experiments were carried out using MMSegmentation~\cite{mmseg2020} with ImageNet~\cite{deng2009imagenet} pre-trained weights, and data augmentation including color jittering, horizontal flipping, and random scaling was applied during training. Detailed settings per dataset and backbone are provided in Table~\ref{tab:implementation}. For Contextrast++, we use $\lambda_{1}=0.1$, $\lambda_{2}=0.3$, $\lambda_{3}=0.7$, $\lambda_{4}=1.0$ in~\eqref{eq:pixel-anchor}, $\alpha = 0.1$ in~\eqref{eq:objective}, $K_\mathcal{M}=100$ for the memory bank, and $K_s=50\%$ for BANE sampling. Test-time evaluation follows common practice~\cite{chen2017rethinking,sun2019high,yuan2020object} with multi-scale inference and horizontal flipping, using scaling factors from $0.75$ to $2.0$ at intervals of $0.25$.

% \noindent{\bf Hyperparameter settings for Contextrast++.} Through the ablation studies, $\lambda_{1}$, $\lambda_{2}$, $\lambda_{3}$, and $\lambda_{4}$ in~\eqref{eq:pixel-anchor} are set to 0.1, 0.3, 0.7, and 1.0, respectively;
%  contrastive loss weight $\alpha$ in~\eqref{eq:objective} is set to 0.1;
%  the number of stored features per class in the memory bank $K_\mathcal{M}$ is set to 100;
%  the sampling ratio $K_s$ for BANE sampling is set to 50\% (see Section~\ref{sec:exp}.\textit{C}). 

% \vspace{2mm}
% \noindent \textbf{Test settings.} The evaluation is performed following the common practice~\cite{sun2019high,yuan2020object,chen2017rethinking}, where predictions are averaged over various image scales and augmented with horizontal flipping. We apply a scaling factor ranging from 0.75 to 2.0, with intervals of 0.25.

\vspace{2mm}

\noindent\textbf{Evaluation metric.} We adopt mean Intersection over Union (mIoU)~\cite{wang2021exploring, hu2021region, pissas2022multi, zhou2024cross, sung2024contextrast} as the primary evaluation metric. In addition, we employ boundary mIoU (B-mIoU) to specifically assess segmentation accuracy near object boundaries, defined as:
\begin{equation}
  \text{B-mIoU} = \frac{1}{\classTotal}\sum_{n=1}^{\classTotal}\frac{\text{TP}^\classIdx_{\tau_B}}{\text{TP}^\classIdx_{\tau_B} + \text{FP}^\classIdx_{\tau_B} + \text{FN}^\classIdx_{\tau_B}},
  \label{eq:b-miou}
\end{equation}
where $\tau_B$ is the pixel threshold to adjust the distance from the boundary.

\begin{table*}[t!]
    \renewcommand{\arraystretch}{1.2}
    \centering
    \caption{Comparison of segmentation performance between our approach and recent contrastive learning-based methods on public datasets.
    Experiments were conducted using CNN-based backbones, including DeepLabV3~\cite{chen2017rethinking}, HRNet~\cite{sun2019high}, and OCRNet~\cite{yuan2020object}. \linebreak
    (·) indicates absolute improvement in percentage points (\%p) over the corresponding baseline.}
    \resizebox{0.95\textwidth}{!}{%
    \begin{tabular}{l|lc|ccccc}
        \hline
        \rowcolor[HTML]{DAE8FC}
        \cellcolor[HTML]{DAE8FC}  & \multicolumn{2}{c|}{Description} & \multicolumn{5}{c}{Dataset {[}mIoU (\%){]}}                                   \\ [-0.4pt] \hhline{~-------}
        \rowcolor[HTML]{DAE8FC}
        \multicolumn{1}{c|}{\multirow{-2}{*}{\cellcolor[HTML]{DAE8FC}Method}}                   & \multicolumn{1}{c}{Loss}   & Sampling                & Cityscapes-\texttt{val}    & ADE20K        & PASCAL-C    & COCO-Stuff        & CamVid      \\ \hline
        DeepLabV3~\cite{chen2017rethinking}                                    &    $L_\mathrm{CE}$    & None                    & 77.12         & 42.85         & 52.83         & 36.04         & 78.80         \\
        DeepLabV3 + Multi~\cite{pissas2022multi}                            &   $L_\mathrm{CE} + L_\text{cms} + L_\text{ccs}$     & Random                  & 78.94 \color[HTML]{2D8C00}(+1.82) & 43.86 \color[HTML]{2D8C00}(+1.01) & 52.27 \color[HTML]{8A0101}(-0.56) & 36.34 \color[HTML]{2D8C00}(+0.30) & 79.67 \color[HTML]{2D8C00}(+0.87) \\
        DeepLabV3 + Contextrast~\cite{sung2024contextrast}                      &  $L_\mathrm{CE} + L^{-}_\mathrm{PA}{^*}$      & Boundary-aware   & 79.35 \color[HTML]{2D8C00}(+2.23) & 44.12 \color[HTML]{2D8C00}(+1.27) & 53.81 \color[HTML]{2D8C00}(+0.98) & 36.55 \color[HTML]{2D8C00}(+0.51) & 79.98 \color[HTML]{2D8C00}(+1.18) \\
        \rowcolor[HTML]{EFEFEF}
        DeepLabV3 + Contextrast++                    &   $L_\mathrm{CE} + L_\mathrm{PA}{^*} + L_\mathrm{AA}$     & Boundary-aware    & \textbf{81.02 \color[HTML]{2D8C00}(+3.90)}  & \textbf{45.93 \color[HTML]{2D8C00}(+3.08)}              & \textbf{54.23 \color[HTML]{2D8C00}(+1.40)}              & \textbf{36.97 \color[HTML]{2D8C00}(+0.93)}              & \textbf{80.19 \color[HTML]{2D8C00}(+1.39)}              \\ \hline
        HRNet~\cite{sun2019high}                                        &   $L_\mathrm{CE}$     & None                    & 78.48         & 41.52         & 51.96         & 36.04         & 82.17         \\
        HRNet + Pico~\cite{wang2021exploring}                                 &  $L_\mathrm{CE} + L_\mathrm{NCE}$      & Semi-hard               & 81.00 \color[HTML]{2D8C00}(+2.52) & N/A           & N/A           & N/A           & N/A           \\
        HRNet + Pico+~\cite{zhou2024cross}                                 &  $L_\mathrm{CE} + L_\mathrm{NCE}$      & Semi-hard               & 81.60 \color[HTML]{2D8C00}(+3.12) & N/A           & N/A           & N/A           & N/A           \\
        HRNet + Region~\cite{hu2021region}                               &  $L_\mathrm{CE} + L_\mathrm{NCE} + L_{\mathrm{Aux}}$      & Random                  & 81.90 \color[HTML]{2D8C00}(+3.42) & N/A           & N/A           & N/A           & N/A           \\
        HRNet + Multi~\cite{pissas2022multi}                                &   $L_\mathrm{CE} + L_\text{cms} + L_\text{ccs}$     & Random                  & 81.50 \color[HTML]{2D8C00}(+3.02) & 42.55 \color[HTML]{2D8C00}(+1.03) & 52.17 \color[HTML]{2D8C00}(+0.21) & 36.35 \color[HTML]{2D8C00}(+0.31) & 83.14 \color[HTML]{2D8C00}(+0.97) \\
        HRNet + Contextrast~\cite{sung2024contextrast}                          &   $L_\mathrm{CE} + L^{-}_\mathrm{PA}{^*}$     & Boundary-aware    & 82.20 \color[HTML]{2D8C00}(+3.72) & 42.68 \color[HTML]{2D8C00}(+1.16) & 52.91 \color[HTML]{2D8C00}(+0.95) & 36.34 \color[HTML]{2D8C00}(+0.30) & 84.33 \color[HTML]{2D8C00}(+2.16) \\
        \rowcolor[HTML]{EFEFEF}
        HRNet + Contextrast++                        &    $L_\mathrm{CE} + L_\mathrm{PA}{^*} + L_\mathrm{AA}$    & Boundary-aware    & \textbf{82.87 \color[HTML]{2D8C00}(+4.39)} &  \textbf{43.27 \color[HTML]{2D8C00}(+1.75)}             & \textbf{54.31 \color[HTML]{2D8C00}(+2.35)}              & \textbf{36.73 \color[HTML]{2D8C00}(+0.69)}              & \textbf{84.54 \color[HTML]{2D8C00}(+2.37)}              \\ \hline
        OCRNet~\cite{yuan2020object}                                       &   $L_\mathrm{CE}$     & None                    & 79.95         & 41.29         & 53.25         & 38.08         & 82.69         \\
        OCRNet + Multi~\cite{pissas2022multi}                               &   $L_\mathrm{CE} + L_\text{cms} + L_\text{ccs}$     & Random                  & 81.51 \color[HTML]{2D8C00}(+1.56)  & 42.87 \color[HTML]{2D8C00}(+1.58) & 52.78 \color[HTML]{8A0101}(-0.47) & 38.17 \color[HTML]{2D8C00}(+0.09) & 83.82 \color[HTML]{2D8C00}(+1.13) \\
        OCRNet + Contextrast~\cite{sung2024contextrast}                         &   $L_\mathrm{CE} + L^{-}_\mathrm{PA}{^*}$     & Boundary-aware    & 81.94 \color[HTML]{2D8C00}(+1.99) & 42.87 \color[HTML]{2D8C00}(+1.58) & 53.82 \color[HTML]{2D8C00}(+0.57) & 38.34 \color[HTML]{2D8C00}(+0.26) & 84.10 \color[HTML]{2D8C00}(+1.41) \\
        \rowcolor[HTML]{EFEFEF}
        OCRNet + Contextrast++                       &   $L_\mathrm{CE} + L_\mathrm{PA}{^*} + L_\mathrm{AA}$     & Boundary-aware    &  \textbf{82.92 \color[HTML]{2D8C00}(+2.97)}             & \textbf{43.03 \color[HTML]{2D8C00}(+1.74)}              & \textbf{53.96 \color[HTML]{2D8C00}(+0.71)}              & \textbf{38.71 \color[HTML]{2D8C00}(+0.63)}              & \textbf{84.32 \color[HTML]{2D8C00}(+1.63)}              \\ \hline
        \end{tabular}%
    }
    \begin{flushleft}
      \quad \quad ${^*} L^{-}_\mathrm{PA} / L_\mathrm{PA}$: Pixel-to-anchor loss without/with our adaptive fusion module, respectively.
	  \end{flushleft}
    \label{tab:comparison_cnn}
\end{table*}

\vspace{2mm}

\noindent \textbf{Compared state-of-the-art approaches.} Since single-scale contrastive learning methods~\cite{wang2021exploring, hu2021region, zhou2024cross} (referred to as \textit{Pico}~\cite{wang2021exploring}, \textit{Pico+}~\cite{zhou2024cross}, and Region~\cite{hu2021region}) are not fully open-sourced, we report results from the original papers. Pico employs the InfoNCE loss~\cite{gutmann2010noise, oord2018representation}, Pico+ additionally generates synthetic negative samples, and Region incorporates an auxiliary loss $L_\mathrm{Aux}$. For multi-scale contrastive learning methods, we compare against Pissas~\etal~\cite{pissas2022multi} (Multi) and Contextrast~\cite{sung2024contextrast} as the primary baselines. Multi employs multi-scale and cross-scale contrastive losses $L_\text{cms}$ and $L_\text{ccs}$, while Contextrast uses the pixel-to-anchor loss without adaptive fusion $L_\mathrm{PA}^-$. Our Contextrast++ further incorporates the anchor-to-anchor loss $L_\mathrm{AA}$ with the adaptive fused pixel-to-anchor loss $L_\mathrm{PA}$.

\begin{figure*}[t]
    \centering
    \includegraphics[scale=0.60]{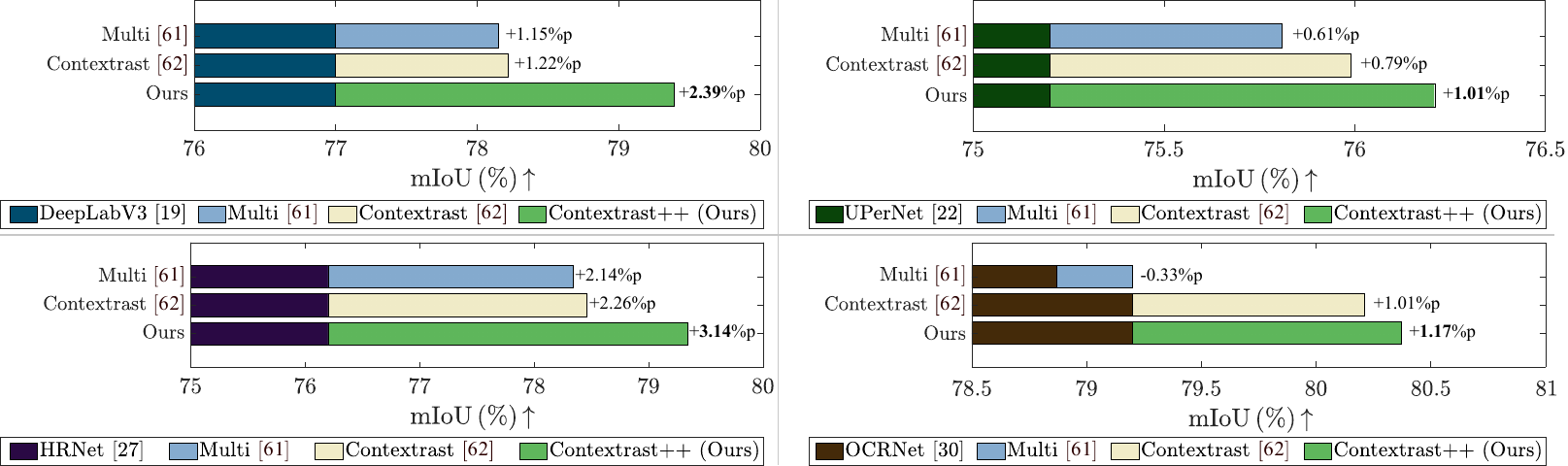}
    \caption{Semantic segmentation performance improvement of baseline architectures~\cite{chen2017rethinking, sun2019high, yuan2020object, xiao2018unified} achieved via state-of-the-art contrastive learning methods evaluated on Cityscapes-\texttt{test}. Note that Multi~\cite{pissas2022multi} slightly degraded performance ($-0.33\%\text{p}$) when applied to OCRNet~\cite{yuan2020object}. As listed in Table~\ref{tab:implementation}, we utilized the UPerNet framework with a Swin-Tiny backbone, replacing the original CNN-based architecture with a Transformer-based architecture.}
\label{fig:miou}
\end{figure*}

\subsection{Evaluation of semantic segmentation performance}
\vspace{2mm}
\noindent \textbf{Comparison with state-of-the-art approaches.} As presented in Table~\ref{tab:comparison_cnn}, Pico+~\cite{zhou2024cross} enhanced the discriminative capability of contrastive learning by introducing synthetic negative samples, thereby outperforming Pico~\cite{wang2021exploring}.
Additionally, incorporating auxiliary loss~$L_\mathrm{{Aux}}$ in Region~\cite{hu2021region} improved mIoU compared with Pico, which only uses cross-entropy loss $L_{\mathrm{CE}}$ with InfoNCE loss $L_{\mathrm{NCE}}$.  
Although Multi~\cite{pissas2022multi} showed performance degradation on PASCAL-C, it generally achieved substantial performance improvements by leveraging multi-scale feature representations in contrastive learning.
In contrast, Contextrast~\cite{sung2024contextrast} demonstrated substantial performance gains across all datasets by leveraging multi-scale feature representations in contrastive learning. 
In particular, Contextrast++ achieved the highest performance improvement, as shown in Table~\ref{tab:comparison_cnn} and Fig.~\ref{fig:miou}.

\begin{figure}[t!]
    \centering
    \includegraphics[scale=0.062]{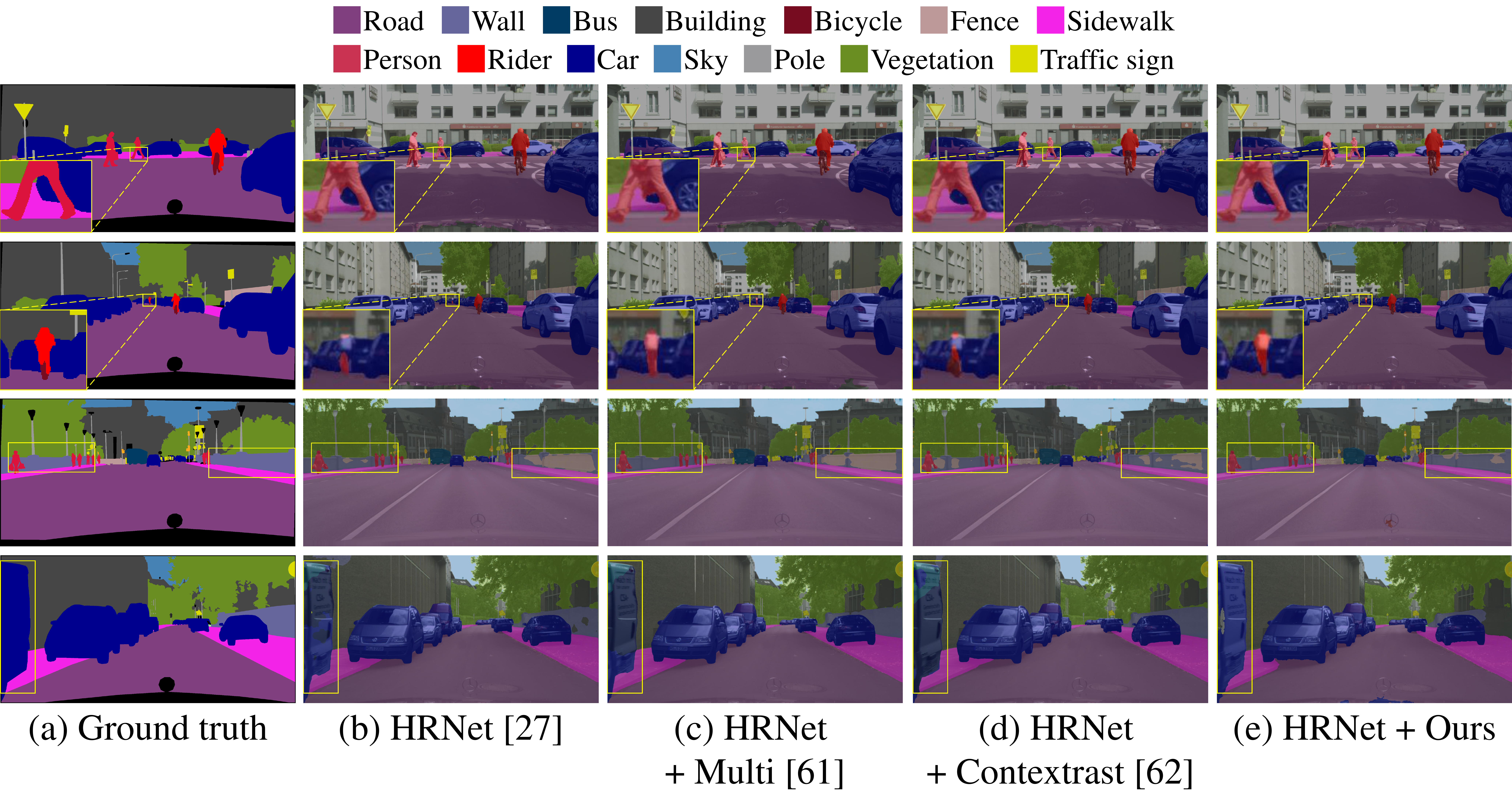}
    \caption{Visualization of (a) ground truth, (b) HRNet~\cite{sun2019high}, (c) HRNet + Multi~\cite{pissas2022multi}, (d) HRNet + Contextrast~\cite{sung2024contextrast}, and (e) HRNet + Ours on Cityscapes-\texttt{val}~(best viewed in color).
    Note that Contextrast++ effectively distinguishes small objects~(first and second rows),
    accurately classifies the wall, which contains excessive local context that makes prediction challenging~(third row),
    and improves predictions on partially occluded or cropped objects near image boundaries~(fourth row).}
    \label{fig:qual_city}
\end{figure}

\begin{figure}[t!]
    \centering
    \includegraphics[scale=0.25]{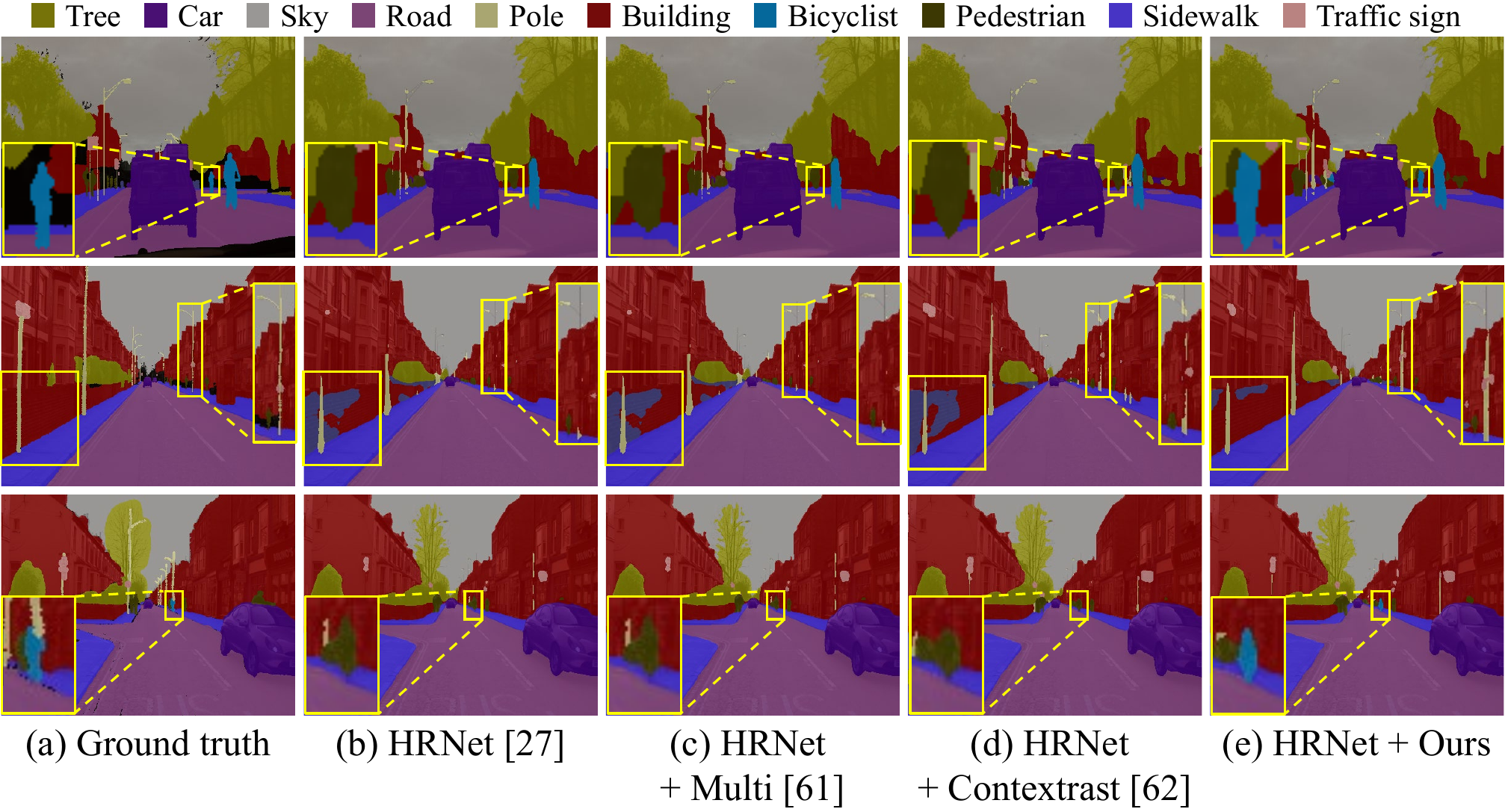}
    \caption{Visualization of (a) ground truth, (b) HRNet~\cite{sun2019high}, (c) HRNet + Multi~\cite{pissas2022multi}, (d) HRNet + Contextrast~\cite{sung2024contextrast}, and (e) HRNet + Ours on the CamVid~dataset~(best viewed in color).
    While all methods perform well on the CamVid dataset, our Contextrast++ effectively distinguishes small objects on the image plane, such as bicyclists and poles.}
    \label{fig:qual_cam}
\end{figure}

As presented in Figs.~\ref{fig:qual_city}, \ref{fig:qual_cam}, \ref{fig:qual_ade}, and \ref{fig:qual_coco}, qualitative results also demonstrate that our Contextrast++ more effectively addresses both over-segmentation and under-segmentation issues than other contrastive learning approaches.
Notably, certain segments on the image plane exhibit substantial improvements, even when compared with Contextrast, highlighting the effectiveness of the additional modules introduced in Contextrast++. 
These results further validate the effectiveness of our approach in improving semantic segmentation accuracy, particularly in challenging regions where existing methods struggle.
For instance, Contextrast++ enables the network to more accurately segment small objects that occupy only a few pixels on the image plane~(Fig.~\ref{fig:qual_cam}), 
and also mitigates under-segmentation of objects, as well as over-segmentation in large background classes such as sea and ground~(Fig.~\ref{fig:qual_coco}).

Therefore, these quantitative and qualitative results demonstrate that our proposed adaptive fusion module and additional AA loss introduced in~\eqref{eq:objective} effectively enhance semantic segmentation performance.

\begin{figure}[t!]
    \centering
    \includegraphics[scale=0.17]{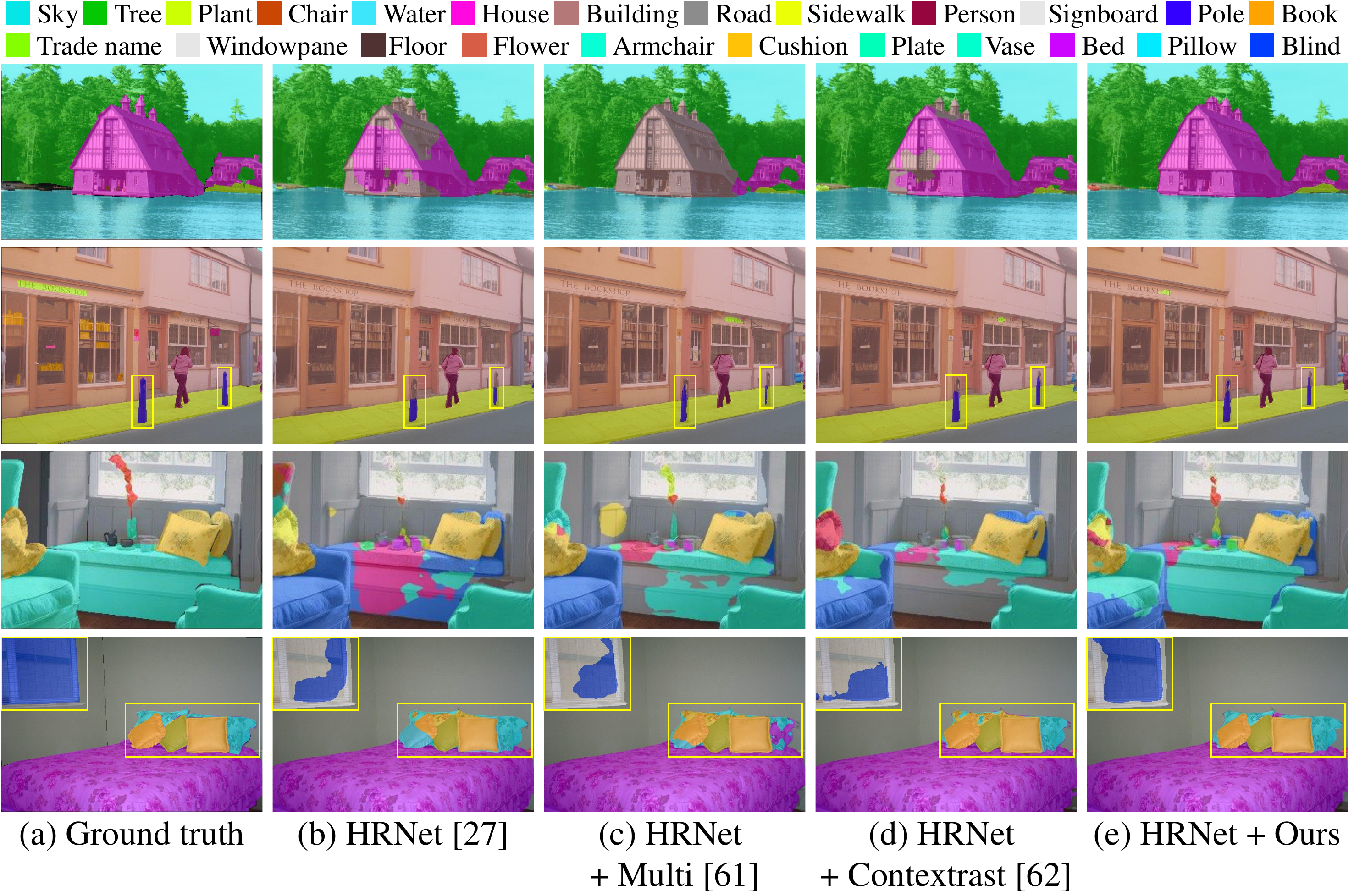}
    \caption{Visualization of (a) ground truth, (b) HRNet~\cite{sun2019high}, (c) HRNet + Multi~\cite{pissas2022multi}, (d) HRNet + Contextrast~\cite{sung2024contextrast}, and (e) HRNet + Ours on the ADE20K dataset~(best viewed in color).
    Note that Contextrast++ yields consistent semantic segmentation performance (first row) and provides more precise segmentation results for small objects than existing state-of-the-art methods (second to fourth rows).}
    \label{fig:qual_ade}
\end{figure}

\begin{figure}[t!]
    \centering
    \includegraphics[scale=0.16]{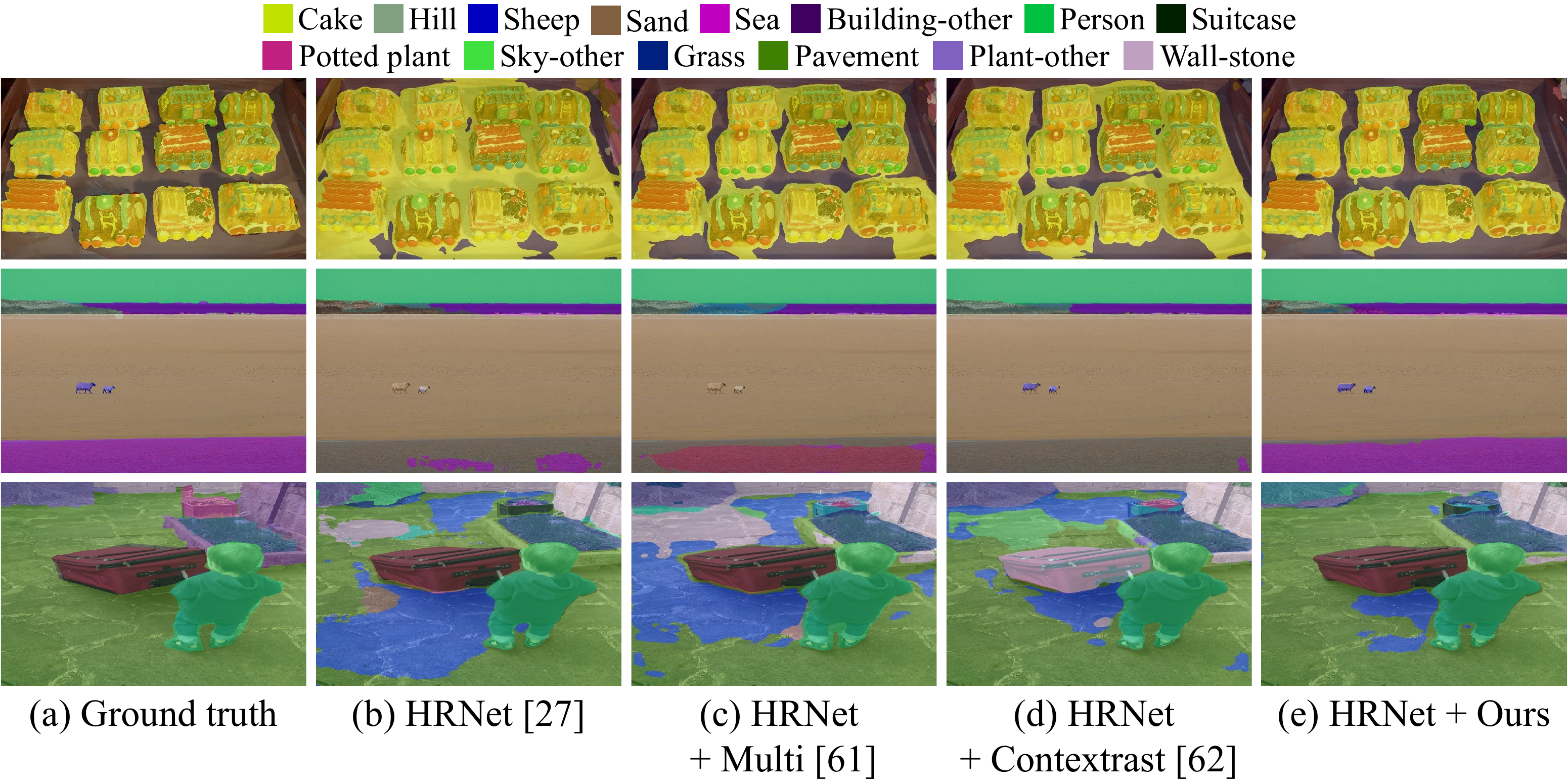}
    \caption{Visualization of (a) ground truth, (b) HRNet~\cite{sun2019high}, (c) HRNet + Multi~\cite{pissas2022multi}, (d) HRNet + Contextrast~\cite{sung2024contextrast}, and (e) HRNet + Ours on the COCO-Stuff dataset~(best viewed in color).
    Our Contextrast++ mitigates under-segmentation of objects as well as over-segmentation in large background classes, such as sea and ground, by refining boundary details.}
    \label{fig:qual_coco}
\end{figure}

\begin{table*}[t]
\centering
\caption{Comparison of performance across head, mid, and tail classes on Cityscapes-\texttt{val}, PASCAL-C, ADE20K, and COCO-Stuff.}
\begin{tabular}{l|ccc|ccc|ccc|ccc}
\hline
\rowcolor[HTML]{DAE8FC} 
\multicolumn{1}{c|}{\cellcolor[HTML]{DAE8FC}}                         & \multicolumn{3}{c|}{\cellcolor[HTML]{DAE8FC}Cityscapes}                                   & \multicolumn{3}{c|}{\cellcolor[HTML]{DAE8FC}PASCAL-C} & \multicolumn{3}{c|}{\cellcolor[HTML]{DAE8FC}ADE20K} & \multicolumn{3}{c}{\cellcolor[HTML]{DAE8FC}COCO-Stuff} \\ \cline{2-13} 
\rowcolor[HTML]{DAE8FC} 
\multicolumn{1}{c|}{\multirow{-2}{*}{\cellcolor[HTML]{DAE8FC}Method}} & \cellcolor[HTML]{DAE8FC}Head & \cellcolor[HTML]{DAE8FC}Mid & \cellcolor[HTML]{DAE8FC}Tail & Head             & Mid              & Tail            & Head            & Mid             & Tail            & Head              & Mid             & Tail             \\ \hline
HRNet~\cite{sun2019high}                                                                 & 90.67                        & 74.05                       & 72.54                        & 69.24            & 57.35            & 27.54           & 55.69           & 34.62           & 32.44           &  62.2                &   34.08              & 12.27                 \\
HRNet + Multi~\cite{pissas2022multi}                                                         & 90.64                        & 73.61                       & 77.32                        & 69.44            & 57.52            & 27.78           & 56.88           & 36.37           & 32.80           &  62.47                 &  34.21               &  14.52                \\
HRNet + Contextrast~\cite{sung2024contextrast}                                                   & 90.91                        & 74.90                       & 79.46                        & 69.84            & 58.61            & 28.35           & 56.40           & 36.70           & 32.86           &  62.41                 &  34.66               &  14.15                \\
\rowcolor[HTML]{EFEFEF} 
HRNet + Contextrast++                                                 & \textbf{90.91}                        & \textbf{75.46}                       & \textbf{80.97}                        & \textbf{70.31}            & \textbf{59.11}            & \textbf{32.36}           & \textbf{57.52}           & \textbf{36.86}           & \textbf{33.54}           &  \textbf{62.76}                 &  \textbf{35.45}               &  \textbf{17.17}                \\ \hline
\end{tabular}
\label{tab:long-tail}
\end{table*}

\vspace{2mm}
\noindent \textbf{Robustness against long-tailed distribution problem.} To more thoroughly assess robustness under long-tailed distributions, we partitioned the classes into three subsets (head, mid, and tail), each containing an approximately equal number of classes based on their pixel frequency. As shown in Table~\ref{tab:long-tail}, Contextrast++ achieves the highest accuracy across all groups, with notably larger gains for tail classes. Specifically, compared with Contextrast, Contextrast++ improves tail-class IoU by +1.51\%p on Cityscapes, +4.01\%p on PASCAL-C, +0.68\%p on ADE20K, and +3.02\%p on COCO-Stuff, demonstrating consistent advantages even on datasets with more severe imbalance.

These improvements primarily arise from the proposed $L_{\mathrm{AA}}$ and the class-prioritized anchors in the memory bank, of which the latter maintains a balanced number of representative anchors for each class. Unlike the $L_{\mathrm{PA}}$ using the mini-batch, this design provides a more stable and balanced supervisory signal, enhancing the representation quality of rare classes without degrading the performance of frequent classes. These quantitative results demonstrate that Contextrast++ effectively mitigates the long-tailed distribution problem and substantially enhances the representation of rare classes.

\vspace{2mm}
\noindent \textbf{Robustness on boundary regions and small objects.} 
As shown in Figs.~\ref{fig:qual_ade} and~\ref{fig:qual_coco}, Contextrast++ produces more accurate segmentation along object boundaries on the ADE20K and COCO-Stuff datasets, which contain over 100 classes and thus exhibit more complex class distributions than Cityscapes and CamVid. Quantitatively, Fig.~\ref{fig:b-miou} shows that Contextrast~\cite{sung2024contextrast} and Contextrast++ improve B-mIoU over Multi~\cite{pissas2022multi} by integrating BANE sampling into a multi-scale contrastive learning framework. Notably, Contextrast++ further outperforms Contextrast by leveraging the adaptive fusion module to dynamically aggregate local and global contexts.

We further conducted an object scale-based mIoU evaluation, categorizing objects into four groups by pixel size: tiny~(up to $16\times16$), small~(up to $32\times32$), medium~(up to $96\times96$), and large~(larger than $96\times96$). As shown in Table~\ref{tab:object_size}, Contextrast++ consistently outperforms the baselines across all groups, with notable gains at finer scales. These results indicate that the adaptive fusion module and BANE sampling jointly enhance fine-grained feature learning: the former by balancing local and global contexts and the latter by focusing on boundary regions, yielding robust segmentation in small object and boundary regions.

\begin{table}[t]
\centering
\caption{Performance comparison of mIoU across different object scales on the PASCAL-C and ADE20K datasets. The evaluation is categorized into four groups based on object pixel size.}
\begin{tabular}{l|cccc}
\hline
\rowcolor[HTML]{DAE8FC} 
\multicolumn{1}{c|}{\cellcolor[HTML]{DAE8FC}}                         & \multicolumn{4}{c}{\cellcolor[HTML]{DAE8FC}PASCAL-C} \\ \cline{2-5} 
\rowcolor[HTML]{DAE8FC} 
\multicolumn{1}{c|}{\multirow{-2}{*}{\cellcolor[HTML]{DAE8FC}Method}} & Large         & Medium        & Small        & Tiny        \\ \hline
HRNet + Multi~\cite{pissas2022multi}                                                                 & 52.90         & 17.12         & 9.35         & 5.1         \\
HRNet + Contextrast~\cite{sung2024contextrast}                                                           & 53.61         & 16.54         & 9.4          & 5.24        \\
\rowcolor[HTML]{EFEFEF} 
HRNet + Contextrast++                                                  & \textbf{54.86}         & \textbf{19.86}         & \textbf{11.51}        & \textbf{6.51}        \\ \hline
\rowcolor[HTML]{DAE8FC} 
\multicolumn{1}{c|}{\cellcolor[HTML]{DAE8FC}}                         & \multicolumn{4}{c}{\cellcolor[HTML]{DAE8FC}ADE20K}         \\ \cline{2-5} 
\rowcolor[HTML]{DAE8FC} 
\multicolumn{1}{c|}{\multirow{-2}{*}{\cellcolor[HTML]{DAE8FC}Method}} & Large         & Medium        & Small        & Tiny        \\ \hline
HRNet + Multi~\cite{pissas2022multi}                                                                 & 43.55         & 11.93         & 6.44         & 5.2         \\
HRNet + Contextrast~\cite{sung2024contextrast}                                                           & 43.63         & 11.84         & 6.26         & 4.82        \\
\rowcolor[HTML]{EFEFEF} 
HRNet + Contextrast++                                                  & \textbf{44.27}         & \textbf{12.07}         & \textbf{6.47}         & \textbf{5.25}        \\ \hline
\end{tabular}
\label{tab:object_size}
\end{table}

\begin{figure*}[t!]
    \centering
    \includegraphics[scale=0.5]{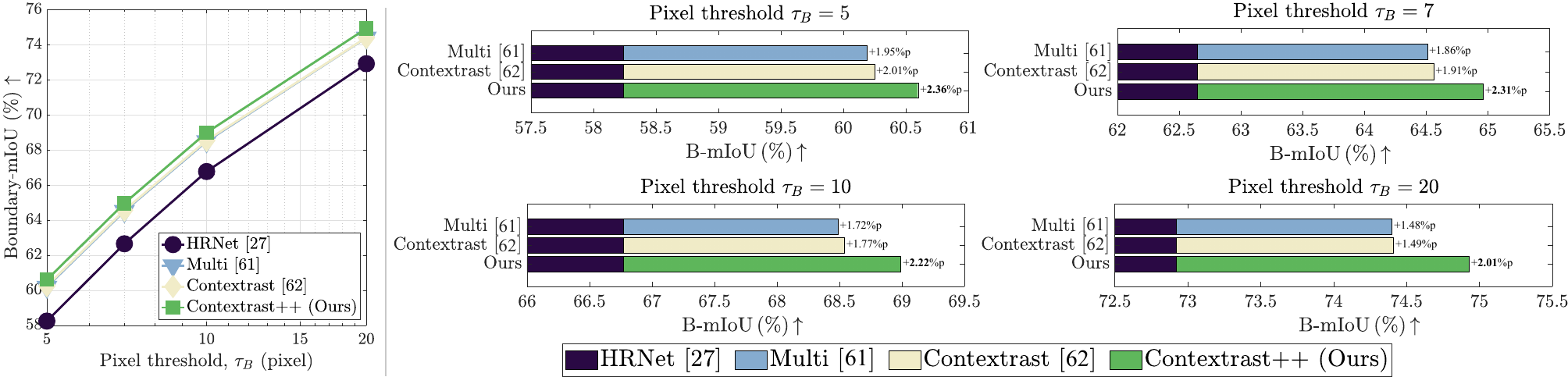}
    \caption{Comparison of boundary mIoU (B-mIoU) performance improvements across state-of-the-art methods on Cityscapes-\texttt{val} using HRNet~\cite{sun2019high} as the baseline model.
  We report B-mIoU and the absolute gain in percentage points at each pixel threshold $\tau_B$.}
    \label{fig:b-miou}
\end{figure*}

\begin{table}[t]
    \caption{Quantitative comparison on Cityscapes-\texttt{val} using lightweight and Transformer-based (UPerNet with a Swin) backbones.}
      \centering
      \renewcommand{\arraystretch}{1.2}
      \begin{tabular}{l|c}
          \hline
          \rowcolor[HTML]{DAE8FC}
          \multicolumn{1}{c|}{Lightweight backbone}       & mIoU (\%) \\ \hline
          MobileNetV2~\cite{sandler2018mobilenetv2}                                     & 71.30  \\
          MobileNetV2 + Pico~\cite{wang2021exploring}                              & 72.30 \color[HTML]{2D8C00}(+1.00)  \\
          MobileNetV2 + Pico+~\cite{zhou2024cross}                             & 72.60 \color[HTML]{2D8C00}(+1.30)  \\
          MobileNetV2 + Multi~\cite{pissas2022multi}                             & 77.58 \color[HTML]{2D8C00}(+6.28)  \\
          MobileNetV2 + Contextrast~\cite{sung2024contextrast}                       & 78.21 \color[HTML]{2D8C00}(+6.91)  \\
          \rowcolor[HTML]{EFEFEF}
          MobileNetV2 + Contextrast++                     & \textbf{78.34 \color[HTML]{2D8C00}(+7.04)} \\ \hline
          \rowcolor[HTML]{DAE8FC}
          \multicolumn{1}{c|}{Transformer-based backbone} & mIoU (\%) \\ \hline
          UPerNet~\cite{xiao2018unified}                                        & 77.75 \\
          UPerNet + Multi~\cite{pissas2022multi}                               & 78.72 \color[HTML]{2D8C00}(+0.97)  \\
          UPerNet + Contextrast~\cite{sung2024contextrast}                         & 79.26 \color[HTML]{2D8C00}(+1.51) \\
          \rowcolor[HTML]{EFEFEF}
          UPerNet + Contextrast++                       & \textbf{79.63 \color[HTML]{2D8C00}(+1.88)} \\ \hline
          Mask2Former~\cite{cheng2022masked} (Swin-t) & 81.87 \\
          Mask2Former + Multi~\cite{pissas2022multi} & 82.59 \color[HTML]{2D8C00}(+0.72) \\
          Mask2Former + Contextrast~\cite{sung2024contextrast} & 81.58 \color[HTML]{8A0101}(-0.29) \\
          \rowcolor[HTML]{EFEFEF}
          Mask2Former + Contextrast++ & \textbf{82.92} \color[HTML]{2D8C00}(+1.05) \\ \hline
          SegFormer~\cite{xie2021segformer} (MiT-B0) & 78.22 \\
          SegFormer + Multi~\cite{pissas2022multi} & 76.19 \color[HTML]{8A0101}(-2.03) \\
          SegFormer + Contextrast~\cite{sung2024contextrast} & 78.41 \color[HTML]{2D8C00}(+0.19)  \\
          \rowcolor[HTML]{EFEFEF}
          SegFormer + Contextrast++ & \textbf{78.55} \color[HTML]{2D8C00}(+0.33) \\ \hline
      \end{tabular}
      \label{tab:lightweight}
  \end{table}

\begin{table}[t]
\centering
\renewcommand{\arraystretch}{1.2}
\caption{Quantitative comparison on ADE20K using Transformer-based backbones.}
\begin{tabular}{l|c}
\hline
\rowcolor[HTML]{DAE8FC} 
\multicolumn{1}{c|}{\cellcolor[HTML]{DAE8FC}}                         & \cellcolor[HTML]{DAE8FC}                            \\
\rowcolor[HTML]{DAE8FC} 
\multicolumn{1}{c|}{\multirow{-2}{*}{\cellcolor[HTML]{DAE8FC}Method}} & \multirow{-2}{*}{\cellcolor[HTML]{DAE8FC}mIoU (\%)} \\ \hline
Mask2Former~\cite{cheng2022masked} (Swin-t)                                                  & 47.12                                               \\
Mask2Former + Multi~\cite{pissas2022multi}                                                   &   47.32 \color[HTML]{2D8C00}{(+0.20)}                                                  \\
Mask2Former + Contextrast~\cite{sung2024contextrast}                                             & 47.73 \color[HTML]{2D8C00}{(+0.61)}                                       \\
\rowcolor[HTML]{EFEFEF} 
Mask2Former + Contextrast++                                           & 48.07 \color[HTML]{2D8C00}{(+0.95)}                                       \\ \hline
SegFormer~\cite{xie2021segformer} (MiT-B0)                                                    & 37.78                                               \\
SegFormer + Multi~\cite{pissas2022multi}                                                     &  37.90 \color[HTML]{2D8C00}{(+0.12)}                                                   \\
SegFormer + Contextrast~\cite{sung2024contextrast}                                               & 37.84 \color[HTML]{2D8C00}{(+0.06)}                                       \\
\rowcolor[HTML]{EFEFEF} 
SegFormer + Contextrast++                                             & 38.00 \color[HTML]{2D8C00}{(+0.22)}                                       \\ \hline
SegFormer~\cite{xie2021segformer} (MiT-B1)                                                    & 41.21                                               \\
SegFormer + Multi~\cite{pissas2022multi}                                                     & 41.91 \color[HTML]{2D8C00}{(+0.70)}                                                    \\
SegFormer + Contextrast~\cite{sung2024contextrast}                                               & 42.10 \color[HTML]{2D8C00}{(+0.89)}                                        \\
\rowcolor[HTML]{EFEFEF} 
SegFormer + Contextrast++                                             & 42.36 \color[HTML]{2D8C00}{(+1.15)}                                       \\ \hline
\end{tabular}
\label{tab:transformer}
\end{table}

\vspace{2mm}
\noindent \textbf{Generalizability of Contextrast++.} To further assess the generalizability of Contextrast++, we evaluated its performance across diverse backbone architectures and datasets.
Table~\ref{tab:lightweight} presents results on Cityscapes-\texttt{val}, where Contextrast++ proves effective not only with the large CNN-based architectures used in Table~\ref{tab:comparison_cnn}, but also with a lightweight CNN-based architecture~\cite{sandler2018mobilenetv2} and Transformer-based architectures~\cite{xiao2018unified, xie2021segformer, cheng2022masked}. Table~\ref{tab:transformer} further evaluates Contextrast++ on ADE20K with Transformer-based backbones, including Mask2Former~\cite{cheng2022masked} and SegFormer~\cite{xie2021segformer}, where Contextrast++ consistently yields improvements over both Multi~\cite{pissas2022multi} and Contextrast~\cite{sung2024contextrast} across all backbones. These results confirm that Contextrast++ is an architecture-agnostic and generalizable contrastive learning framework.

\subsection{Ablation studies}\label{sec:ablation}

\noindent\textbf{Effectiveness of individual components.}
To evaluate the effect of each individual module in Contextrast++, we performed a comprehensive ablation study.
As shown in Table~\ref{tab:abl_module}, each additional module introduced in Contextrast++, \ie~the adaptive fusion module, $L_\mathrm{AA}$, and BANE sampling, substantially improved semantic segmentation performance.

Specifically, replacing the static fusion module,~\ie~\eqref{eq:weighted_sum}, with our adaptive fusion module,~\ie~\eqref{eq:adaptive_fusion}, enhanced the performance by effectively balancing global and local contexts, resulting in better anchor representation.
Employing $L_\mathrm{AA}$ also improved performance by balancing class-wise samples in contrastive learning, thereby addressing the long-tailed distribution issue.
Additionally, instead of semi-hard sampling~\cite{wang2021exploring}, which selects negative samples based on cosine similarity, adopting BANE sampling led to better performance by selecting more informative negative samples.
When combining all the proposed methods, the model achieved the highest mIoU, 82.87\%, demonstrating the importance of each proposed component in this paper.

\vspace{2mm}
\noindent\textbf{Effect of multi-scale feature aggregation.}
Table~\ref{tab:abl_layers} provides a comprehensive ablation study on the contribution of each encoder layer. When only a single scale is used, corresponding to the first and the last rows, respectively, performance drops noticeably. This indicates that no single layer captures a sufficiently comprehensive set of semantic and fine-detailed features.

In particular, removing global semantic features leads to the largest degradation, as evidenced by the first three rows in Table~\ref{tab:abl_layers}, reflecting their essential role in capturing global semantics and object-level structure. Conversely, removing fine-grained features also negatively impacts performance, as evidenced by the last three rows in Table~\ref{tab:abl_layers}, as these layers contain fine-grained spatial details crucial for boundary delineation and small object discrimination.

Performance consistently improves as more scales are combined, demonstrating that each scale provides complementary cues. The best result is achieved when all four scales are aggregated, confirming that Contextrast++ benefits from fine details in early layers and semantically rich information in deeper layers.

This experiment highlights the importance of multi-scale integration and further distinguishes our method from Multi~\cite{pissas2022multi}, which aggregates only a subset of layers, thereby limiting the exploitation of complementary multi-level contextual information.

\begin{table}[]
    \caption{Ablation study of each module on Cityscapes-\texttt{val}.}
    \resizebox{0.98\columnwidth}{!}{%
    \begin{tabular}{c|c|cc|c}
    \hline
    \rowcolor[HTML]{DAE8FC} 
    \cellcolor[HTML]{DAE8FC} Fusion & {\cellcolor[HTML]{DAE8FC} Balancing}  & \multicolumn{2}{c|}{\cellcolor[HTML]{DAE8FC}Sampling}  & \cellcolor[HTML]{DAE8FC} \\ [-0.4pt] 
    \hhline{----~}
    \rowcolor[HTML]{DAE8FC} 
    \cellcolor[HTML]{DAE8FC} Adaptive fusion & $L_\mathrm{AA}$  & Semi-hard  & BANE  & \multirow{-2}{*}{\cellcolor[HTML]{DAE8FC}mIoU (\%)} \\ \hline
     &  &  &  & 81.88 \\ 
    \checkmark &  &  &  & 82.18 \\ 
    \checkmark & \checkmark &  &  & 82.35  \\
    \checkmark &  & \checkmark  &  & 82.23  \\ 
    \checkmark &  &  & \checkmark  & 82.44 \\
    \rowcolor[HTML]{EFEFEF} 
    \checkmark & \checkmark  &  & \checkmark & \textbf{82.87} \\ \hline
    \end{tabular}}
    \label{tab:abl_module}
\end{table}

\begin{table}[t]
\centering
\caption{Ablation study of the effect of multi-scale aggregation in Contextrast++ on Cityscapes-\texttt{val}.}
\begin{tabular}{cccc|c}
\hline
\rowcolor[HTML]{DAE8FC}
1st Layer & 2nd Layer & 3rd Layer & 4th Layer & mIoU (\%) \\
\hline
  \checkmark      &         &         &         & 81.02 \color[HTML]{8A0101}(-1.85)         \\
  \checkmark      &  \checkmark       &         &         &  81.61 \color[HTML]{8A0101}(-1.26)        \\
  \checkmark      &  \checkmark      &  \checkmark       &         &   81.79 \color[HTML]{8A0101}(-1.08)        \\
  \rowcolor[HTML]{EFEFEF}
  \checkmark      &  \checkmark       &  \checkmark       &  \checkmark       & \textbf{82.87}      \\
        &  \checkmark      & \checkmark & \checkmark & 82.36 \color[HTML]{8A0101}(-0.51) \\
        & & \checkmark & \checkmark & 82.19 \color[HTML]{8A0101}(-0.68) \\
        & & & \checkmark & 81.89 \color[HTML]{8A0101}(-0.89) \\
\hline
\end{tabular}
\label{tab:abl_layers}
\end{table}

\begin{table}[t]
    \centering
    \caption{Comparison of different weight settings with the Cityscapes-\texttt{val}.}
    \begin{tabular}[t]{cccc|c}
    \hline
    \rowcolor[HTML]{DAE8FC}
    $\lambda_1$ & $\lambda_2$ & $\lambda_3$ & $\lambda_4$ & \multicolumn{1}{c}{mIoU (\%)} \\ \hline
    1.0        & 1.0        & 1.0       & 1.0       & 82.27                                                 \\ \hline
    1.0        & 0.8        & 0.6       & 0.4       & 82.56                                                 \\ \hline
    1.0        & 0.75       & 0.5       & 0.25      & 79.81                                                 \\ \hline
    1.0        & 0.7        & 0.4       & 0.1       & 81.46                                                 \\ \hline
    \rowcolor[HTML]{EFEFEF} 0.1        & 0.3        & 0.7       & 1.0       & \textbf{82.87}                                                 \\ \hline
    0.1        & 0.4        & 0.7       & 1.0       & 82.55                                                 \\ \hline
    0.25       & 0.5        & 0.75      & 1.0       & 82.23                                                 \\ \hline
    0.4        & 0.6        & 0.8       & 1.0       & 80.65                                                 \\ \hline
    \end{tabular}
    \label{tab:lambda}
\end{table}

\begin{figure}[t]
    \centering
    \includegraphics[scale=0.265]{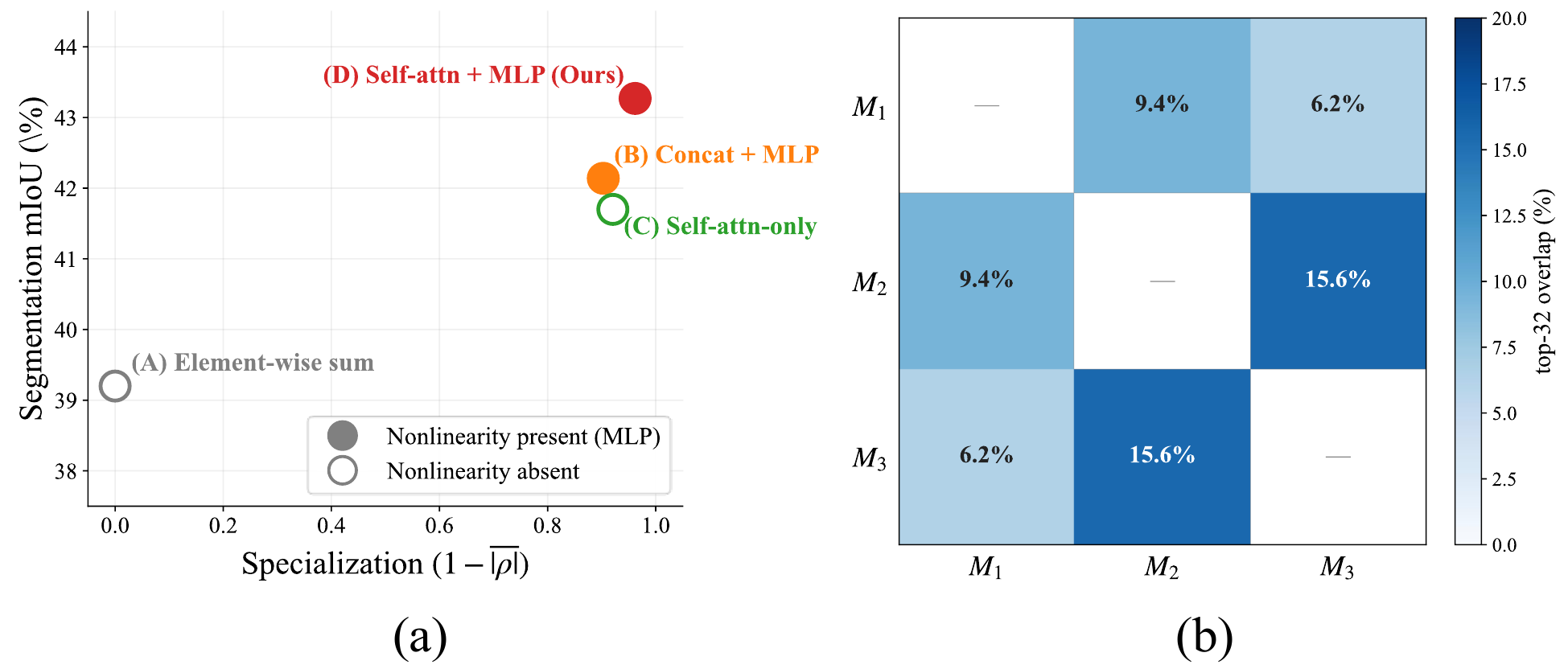}
    \caption{(a)~Design-space view of fusion variants on ADE20K, positioned by channel specialization $1-\overline{|\rho|}$ ($x$-axis) and mIoU ($y$-axis); filled/open circles encode the presence/absence of MLP. Our adaptive fusion module occupies the upper-right corner, whereas Concat~+~MLP and the Self-Attention-only variant fail symmetrically, showing that neither property alone suffices. (b)~Pairwise top-32 channel overlap between modules $M_i$ and $M_j$ on ADE20K. The low mean overlap confirms that the three modules attend to largely disjoint channel subsets.}
    \label{fig:specialization}
\end{figure}

\begin{table}[t]
    \centering
    \caption{Comparison of the adaptive fusion module on Cityscapes-\texttt{val}.}
    \begin{tabular}{l|c} \hline
    \rowcolor[HTML]{DAE8FC}
    \multicolumn{1}{c|}{Method} & mIoU (\%)  \\ \hline
    Concat + MLP                & 82.30     \\
    Multi-head self-attention~\cite{vaswani2017attention}   & 82.37 \\
    Cross-attention~\cite{vaswani2017attention}             & 82.14 \\
    Self-attention + weighted sum & 82.39 \\
    Self-attention + element-wise sum + MLP & 81.62 \\
    Gated fusion (scalar) & 82.36 \\
    Gated fusion (channel-wise) & 80.46 \\
    \rowcolor[HTML]{EFEFEF}
    Adaptive fusion module (Ours)              & \textbf{82.87} \\ \hline
    \end{tabular}
    \label{tab:adaptive_abl}
\end{table}

\begin{table}[t]
\centering
\caption{Performance comparison under different anchor selection strategies for class-prioritized anchors.}
\begin{tabular}{c|c}
\hline
\rowcolor[HTML]{DAE8FC} 
Anchor selection strategy            & mIoU (\%)    \\ \hline
\rowcolor[HTML]{FFFFFF} 
\cellcolor[HTML]{FFFFFF}Earliest layer & 81.73  \\
\rowcolor[HTML]{FFFFFF} 
Average across layer                 & 82.36  \\
\rowcolor[HTML]{EFEFEF} 
Deepest layer (Ours)                 & \textbf{82.87}  \\ \hline
\end{tabular}
\label{tab:anchor-selection}
\end{table}

\begin{table}[t]
\centering
\caption{Performance comparison of different memory bank types with the same memory budget.}
\begin{tabular}{c|c}
\hline
\rowcolor[HTML]{DAE8FC} 
Memory bank type                     & mIoU (\%)     \\ \hline
\rowcolor[HTML]{FFFFFF} 
\cellcolor[HTML]{FFFFFF}Pixel memory~\cite{wang2021exploring, zhou2024cross} & 81.94  \\
\rowcolor[HTML]{EFEFEF} 
Anchor memory (Ours)                 & \textbf{82.87}  \\ \hline
\end{tabular}
\label{tab:mem_type}
\end{table}

\begin{table}[t]
\centering
\caption{Ablation study on memory bank designs varying with selection criteria.}
\begin{tabular}{c|c}
\hline
\rowcolor[HTML]{DAE8FC} 
Memory bank design       & mIoU (\%)                          \\ \hline
\rowcolor[HTML]{FFFFFF} 
Earliest layer             & 81.63  \\
\rowcolor[HTML]{FFFFFF} 
Deepest layer            & 81.79                         \\
\rowcolor[HTML]{FFFFFF} 
All layers                & 80.00                             \\
\rowcolor[HTML]{EFEFEF} 
Class-prioritized anchor (Ours) & \textbf{82.87}                        \\ \hline
\end{tabular}
\label{tab:mem_design}
\end{table}

\vspace{2mm}
\noindent\textbf{Impact of weight settings for contrastive learning.}
We conducted experiments with various hyperparameter combinations to determine the optimal weight settings for multi-scale contrastive loss.
First, we tuned the hyperparameters $\lambda_i$ in~\eqref{eq:pixel-anchor} to balance the contributions of different scale levels.
As shown in Table~\ref{tab:lambda}, a setting of $\lambda_1 = 0.1$, $\lambda_2 = 0.3$, $\lambda_3 = 0.7$, and $\lambda_4 = 1.0$ achieved the highest mIoU of 82.87\%. 

One interesting aspect is that when $\lambda_3$ and $\lambda_4$ are set to smaller values, performance drops noticeably; see the third and fourth rows of Table~\ref{tab:lambda}.
These results suggest that balancing contributions from different layers plays a key role in overall performance.
While deeper-layer features capture more global context and earlier-layer features focus on local details, 
adjusting their relative contributions does not lead to a straightforward increase or decrease in performance.
Instead, we empirically observed that an appropriate balance between global and local features leads to better segmentation performance.

After optimizing $\lambda_i$, we further tuned the weight $\alpha$ to balance the contributions of cross-entropy loss and contrastive loss.
As presented in Fig.~\ref{fig:plots}(a), $\alpha = 0.1$ yielded the best performance, suggesting that an excessively strong influence of the contrastive loss may hinder effective training.

\vspace{2mm}
\noindent\textbf{Design space analysis of the adaptive fusion module.}
To establish that the Self-Attention~+~MLP design is principled rather than incidental, we decompose the space of fusion architectures along two axes that together govern the function each module computes: (i) \emph{channel specialization}, the degree to which the three fusion modules $\{M_1, M_2, M_3\}$ attend to mutually distinct channel subsets, and (ii) \emph{nonlinear channel refinement}, the existence of a non-affine transformation that maps fused activations to class-discriminative representations.

To quantify channel specialization, for each fusion module $M_i$ with input $z \in \mathbb{R}^{d \times H_i \times W_i}$ and output $y = M_i(z)$, we define a per-channel sensitivity vector $\mathbf{s}^{(i)} = (s_1^{(i)}, \dots, s_d^{(i)}) \in \mathbb{R}^d$, where
\begin{equation}
s_c^{(i)} \;=\; \sqrt{\mathbb{E}_x\,\big\|\partial y/\partial z_c\big\|_F^2}
\label{eq:sensitivity}
\end{equation}
measures how strongly the output responds, in expectation over validation inputs $x$, to perturbations of input channel $c$; $\|\cdot\|_F$ denotes the Frobenius norm. The squared Jacobian-column norm is computed with a Hutchinson estimator~\cite{hutchinson1989stochastic, bekas2007estimator}.
The specialization score is then $1-\overline{|\rho|}$, where $\overline{|\rho|} = \tfrac{1}{3}\sum_{i<j}|\rho(\mathbf{s}^{(i)}, \mathbf{s}^{(j)})|$ is the mean absolute Spearman rank correlation across the three module pairs.

Fig.~\ref{fig:specialization}(a) places four fusion variants on the plane spanned by $1-\overline{|\rho|}$ and mIoU on ADE20K.
Element-wise summation~(A) lacks per-module parameters, producing functionally indistinguishable modules and the lowest accuracy.
Concat~+~MLP~(B) and Self-Attention-only~(C) sit at opposite corners of the design space yet degrade by comparable margins: the input-independent routing of~(B) collapses channel preferences across modules, whereas the orthogonal channel rankings of~(C) cannot be transformed into class-discriminative features without a point-wise nonlinearity.
Ours~(D) occupies the upper-right corner with the highest specialization and mIoU, confirming that neither property alone suffices and that their conjunction is what couples channel-level orthogonality to segmentation accuracy.

Fig.~\ref{fig:specialization}(b) confirms this finding at the top-$K$ level by reporting pairwise overlap among the $32$ channels of highest sensitivity in each $\mathbf{s}^{(i)}$: the mean overlap is only $10.4\%$, and the largest pair ($M_2$--$M_3$, $15.6\%$) is an expected consequence of $M_2$ and $M_3$ operating on adjacent scales.
The agreement between the global rank-level statistic and the local top-$K$ set-overlap statistic, computed independently on the same attribution maps but probing different scales of the channel distribution, indicates that the observed orthogonality is a property of the learned representation.

\vspace{2mm}
\noindent\textbf{Performance comparison in the adaptive fusion module.}
We conducted an extensive ablation study that includes fusion variants inspired by existing multi-scale fusion methods~\cite{gu2022multi, shi2023transformer, zhou2024boundary}, in order to evaluate alternative strategies within the adaptive fusion module. These variants were designed to emulate the core fusion behaviors of existing approaches: concatenation with an MLP layer (referred to as concat + MLP) and channel-attention~\cite{zhou2024boundary}, weighted-sum gating~\cite{shi2023transformer}, and element-wise summation after resolution alignment~\cite{gu2022multi}. This setup allows a direct empirical comparison within our framework.

As shown in Table~\ref{tab:adaptive_abl}, the Concat + MLP underperforms because it treats all channels uniformly and cannot capture the relative importance of multi-scale anchors. Multi-head self-attention also results in inferior performance. 
Note that multi-head self-attention is applied to a single anchor vector; the multi-head mechanism unnecessarily fragments the 256-dimensional global semantics into smaller subspaces, hindering the capture of holistic channel correlations. 
Cross-attention performs worse due to semantic mismatch between anchors from different layers, which makes cross-scale alignment difficult and injects noise. 
Self-attention variants that use weighted or element-wise summation also degrade performance, either by retaining the limitations of static weighting or by discarding complementary information before projection. 
Gated fusion provides modest benefits but is still less effective overall, especially in the channel-wise case where overly aggressive suppression leads to unstable representations. 

In contrast, our adaptive fusion module exhibits the best performance by enabling channel-wise refinement while preserving complementary information across scales through subsequent concatenation and projection, producing more discriminative multi-scale anchor representations.

\vspace{2mm}
\noindent\textbf{Impact of class-prioritized anchor selection.}
To examine the effectiveness of the class-prioritized anchor design, we compared three anchor selection strategies: selecting anchors from the earliest layer, averaging across all layers, and selecting from the deepest layer. As shown in Table~\ref{tab:anchor-selection}, earliest-layer anchors yield the lowest performance due to their limited semantic abstraction and higher sensitivity to local noise. While averaging across layers improves performance by incorporating multi-scale cues, it dilutes the representation through uneven semantic abstraction. Selecting anchors from the deepest layer achieves the best performance, validating our choice to use these deepest-layer anchors as the foundation for our class-prioritized anchors in $L_\mathrm{AA}$.

% To examine the effectiveness of the class-prioritized anchor design, we compared three anchor selection strategies: selecting anchors from the earliest layer, averaging anchors across all layers, and selecting anchors from the deepest layer. As shown in Table~\ref{tab:anchor-selection}, selecting the earliest-layer anchors yields the lowest performance, likely due to their limited semantic abstraction and higher sensitivity to local noise. Averaging anchors across layers improves performance by incorporating multi-scale cues, yet the resulting representation can be diluted because different layers provide features with varying degrees of semantic abstraction.

% Selecting anchors from the deepest layer achieves the best performance, indicating that deeper layer representations provide more stable and semantically rich representations. These results validate our choice of using deepest-layer anchors for class-prioritized anchor selection, as they offer the most reliable and discriminative representations for $L_{\mathrm{AA}}$.

\vspace{2mm}
\noindent\textbf{Impact of memory bank size and design.} 
We analyzed the impact of memory bank size, defined as the product of the number of classes $\classTotal$ and the number of stored features per class $K_\mathcal{M}$. As illustrated in Fig.~\ref{fig:plots}(b), performance peaked at $K_\mathcal{M}=100$ and declined thereafter, with a more noticeable drop at $K_\mathcal{M}=400$. This decline arises because an excessively large memory bank causes $L_\mathrm{AA}$ to dominate the overall objective in~\eqref{eq:objective}, suppressing other loss terms and reducing feature diversity. We further compared memory bank designs. Table~\ref{tab:mem_type} shows that our anchor-level memory outperforms pixel-level memory~\cite{wang2021exploring, zhou2024cross} by providing a semantically richer representation. Within the anchor memory, Table~\ref{tab:mem_design} compares anchor selection strategies: using anchors from a single layer or all layers yields lower performance, whereas our class-prioritized design achieves the best performance of 82.87\% mIoU, confirming its suitability for multi-scale structures and class imbalance mitigation.

\vspace{2mm}
\noindent\textbf{BANE sampling analysis.}
To validate the motivation for BANE sampling, we examined how the difficulty of negative samples varies with their distance from the boundaries. For each encoder layer, we computed the cosine similarity between misclassified pixels and the class anchors. As shown in Fig.~\ref{fig:cos_dist}, negative samples located closer to boundaries consistently exhibit lower cosine similarity, demonstrating that boundary-adjacent negative samples constitute inherently harder negative samples. 
% This analysis provides strong empirical justification for the BANE sampling strategy, which intentionally focuses on negative samples near boundaries to improve discriminative learning in these ambiguous regions.

\begin{figure}[t]
    \centering
    \subfloat{\includegraphics[scale=0.07]{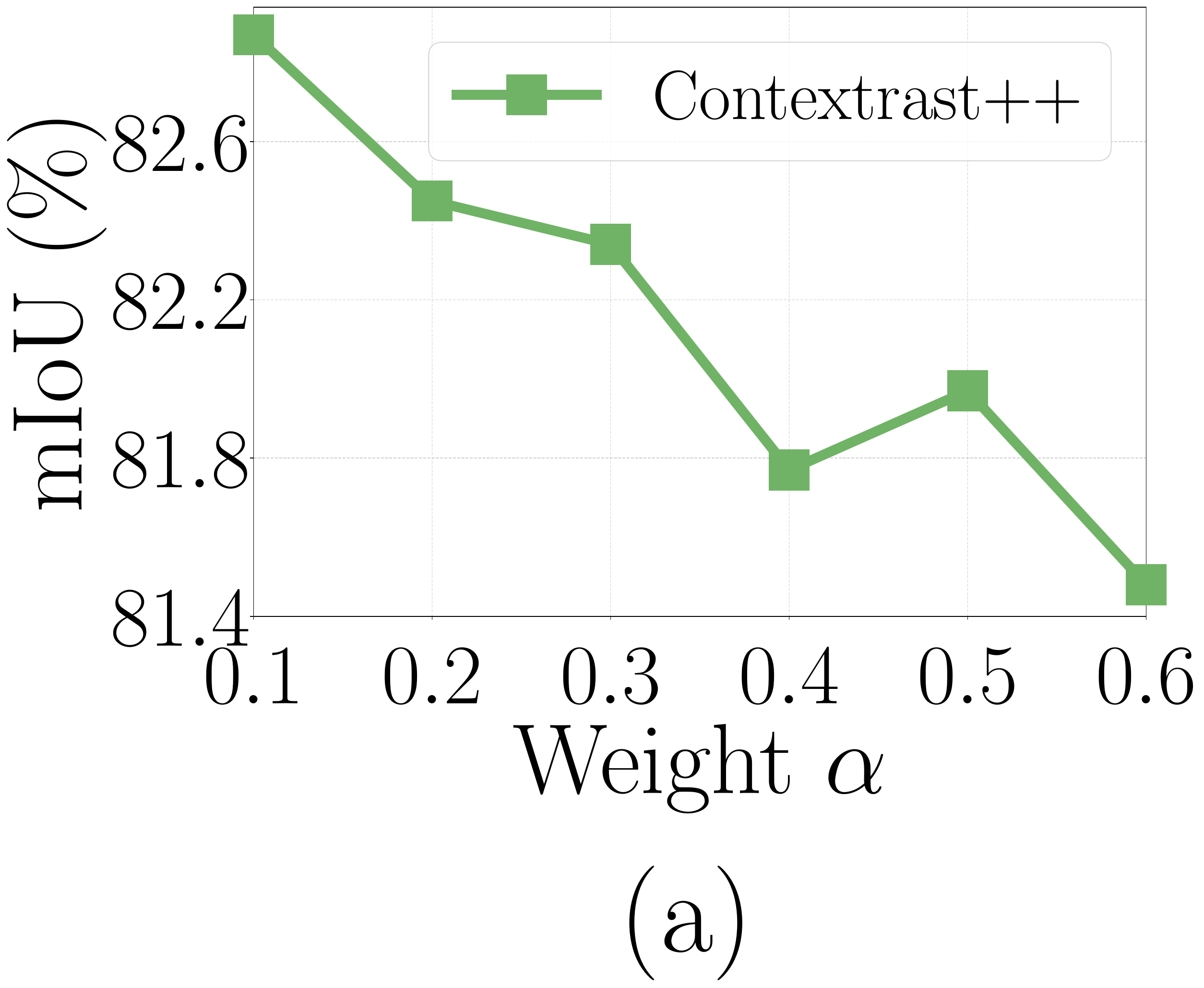}}
    \subfloat{\includegraphics[scale=0.07]{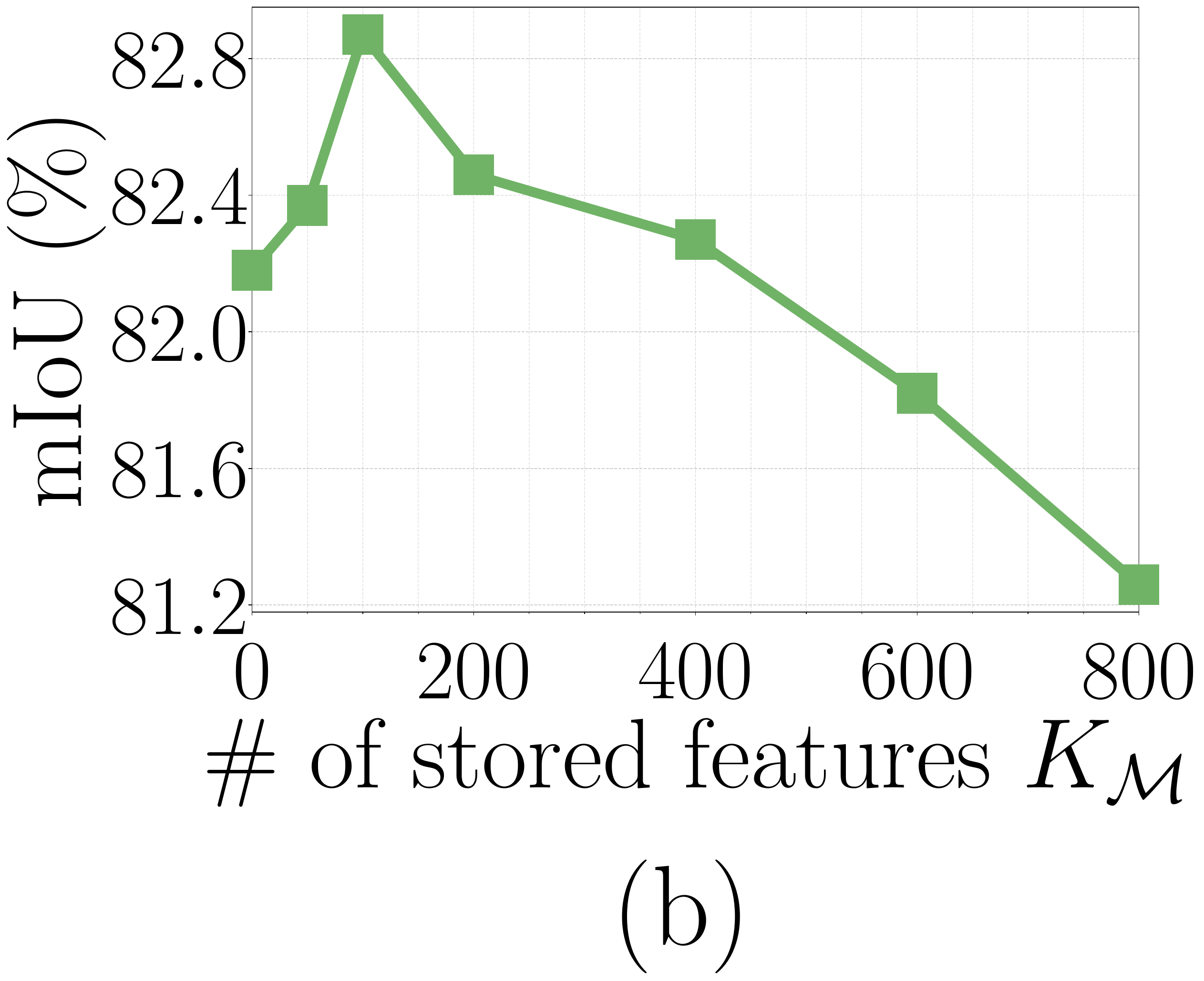}}
    \subfloat{\includegraphics[scale=0.07]{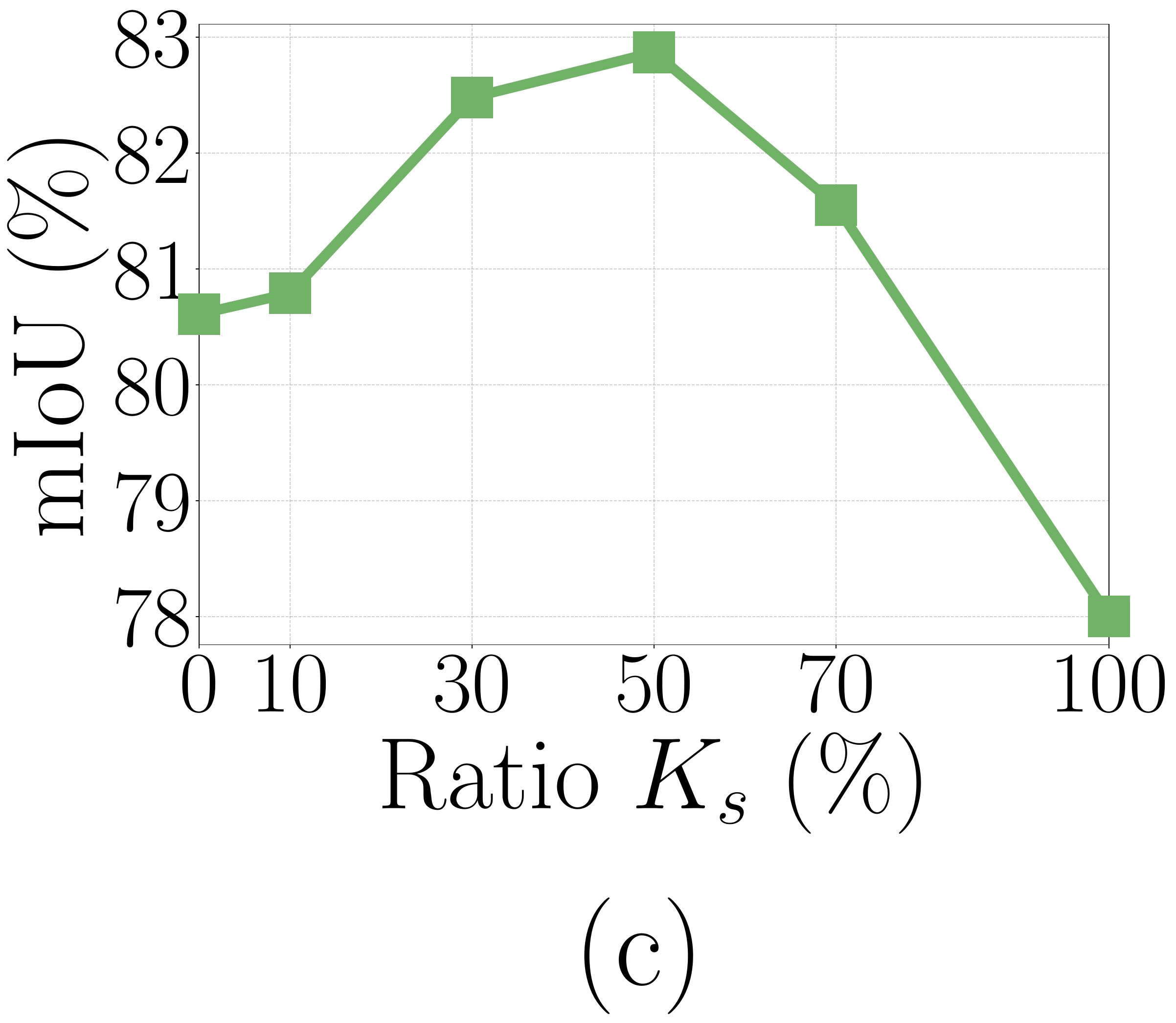}}
    \caption{Performance analyses on Cityscapes-\texttt{val} with HRNet~\cite{sun2019high}: Performance changes with (a)~the hyperparameter weight $\alpha$ in (\eqref{eq:objective}), (b)~number of stored features per class $K_\mathcal{M}$, and (c)~sampling ratio $K_s$ for the boundary-aware negative sampling.}
    \label{fig:plots}
\end{figure}

\begin{figure}[t]
    \centering
    \includegraphics[scale=0.185]{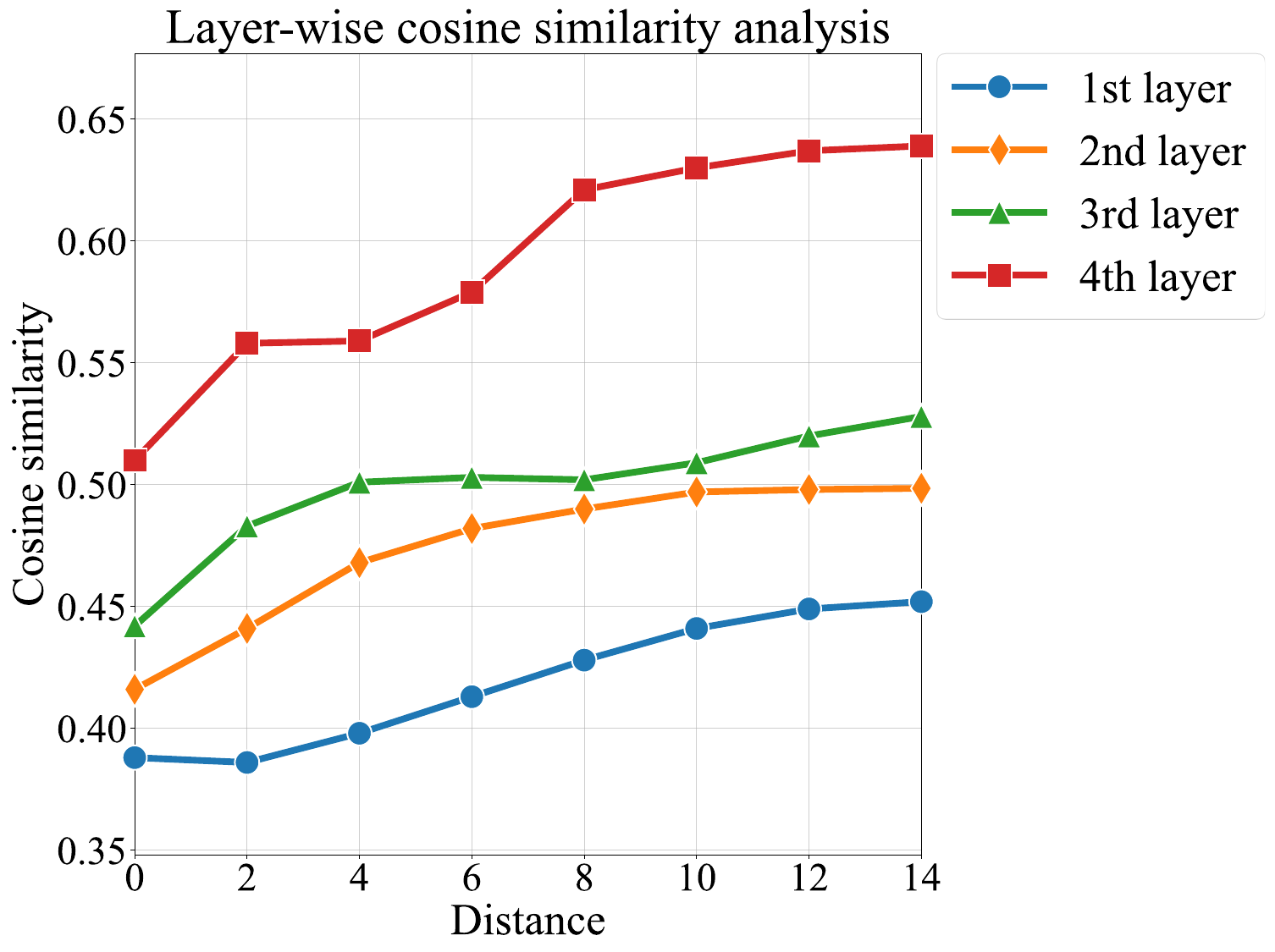}
    \caption{Average cosine similarity between misclassified pixels and representative anchors at each layer of HRNet~\cite{sun2019high} on Cityscapes-\texttt{val}, measured as a function of distance from incorrect prediction boundaries. Boundary-adjacent samples consistently exhibit lower cosine similarity, indicating that they constitute harder negative samples than inner-region samples.}
    \label{fig:cos_dist}
\end{figure}

\vspace{2mm}
\noindent\textbf{Impact of sampling ratio in BANE sampling.}
We evaluate the impact of BANE sampling by testing multiple values of the sampling ratio $K_s$. As shown in Fig.~\ref{fig:plots}(c), performance peaks at a 50\% sampling ratio and deteriorates as the proportion of negative samples grows excessively large, since optimizing deep metric learning with the hardest negative examples can lead to local minima in the early stage of training~\cite{schroff2015facenet, xie2022delving, cai2020all, xuan2020hard}.

% We evaluate the impact of BANE sampling by testing multiple values of the sampling ratio $K_s$, which is presented in Section~\ref{sec:method}.\textit{C}.
% As shown in Fig.~\ref{fig:plots}, our analysis indicates that the sampling strategy yields the greatest performance improvement when a 50\% sampling ratio is used. However, the performance deteriorates when the proportion of negative samples becomes excessively large.
% This phenomenon can be explained by the fact that optimizing deep metric learning with the hardest negative examples can lead to local minima in the early stages of training~\cite{schroff2015facenet, xie2022delving,cai2020all, xuan2020hard}.
% As a result, selecting an appropriate ratio for negative sampling is crucial to maximizing performance.

\begin{table}[t]
    \centering
    \caption{Comparison of FLOPs and the total number of parameters for each contrastive learning method on Cityscapes-\texttt{val}.}
    \begin{tabular}{l|c|c|c}
    \hline
    \rowcolor[HTML]{DAE8FC} 
    \multicolumn{1}{c|}{\cellcolor[HTML]{DAE8FC}Method} & FLOPs   & Params & mIoU (\%) \\ \hline
    HRNet~\cite{sun2019high}                                               & 187.87G & 65.86M & 78.48  \\
    HRNet + Multi~\cite{pissas2022multi}                                         & 187.87G & 66.24M & 81.50      \\
    HRNet + Contextrast~\cite{sung2024contextrast}                                 & 187.87G & 66.24M & 82.20  \\
    \rowcolor[HTML]{EFEFEF} 
    HRNet + Contextrast++                               & 187.87G & 67.82M & \textbf{82.87}   \\ \hline
    \end{tabular}
    \label{tab:flops}
\end{table}

\begin{table}[t]
\centering
\caption{Comparison of memory usage and average training time on Cityscapes-\texttt{val}.}
\begin{tabular}{l|c|c}
\rowcolor[HTML]{DAE8FC} 
\hline
\multicolumn{1}{c|}{\cellcolor[HTML]{DAE8FC}Method} & Memory (MiB) & \begin{tabular}[c]{@{}c@{}}Avg. training time\\ (hours:minutes)\end{tabular} \\ \hline
HRNet~\cite{sun2019high}                                               & 26,692 & 8:58                                                                         \\
HRNet + Multi~\cite{pissas2022multi}                                       & 27,688 & 9:52                                                                         \\
HRNet + Contextrast~\cite{sung2024contextrast}                                 & 27,788 & 10:01                                                                        \\
+ Adaptive fusion module                     & 27,862 & 10:17                                                                        \\
+ AA loss + Memory bank                      & 27,844 & 10:10                                                                        \\
\rowcolor[HTML]{EFEFEF} 
HRNet + Contextrast++                               & 27,924 & 10:22                                                                        \\ \hline
\end{tabular}
\label{tab:memory_training}
\end{table}

\vspace{2mm}
\noindent\textbf{Comparison of memory efficiency.}
To evaluate the computational efficiency of Contextrast++, we first compared FLOPs and parameter counts, as shown in Table~\ref{tab:flops}. Note that because contrastive learning components are used only during training, all contrastive methods introduce no additional FLOPs during inference. Thus, the main difference lies in training-time parameters. Multi~\cite{pissas2022multi} and Contextrast~\cite{sung2024contextrast} use the same number of projection heads and therefore share identical parameter counts, whereas Contextrast++ introduces a lightweight self-attention module that results in only a small parameter increase.

To more concretely assess the overhead, we measured average GPU memory usage and training time on Cityscapes-\texttt{val}, as shown in Table~\ref{tab:memory_training}. Relative to Contextrast, Contextrast++ adds only 136 MiB of memory and 21 minutes of training time during the entire schedule, while leaving inference unchanged. This minimal overhead demonstrates that Contextrast++ improves multi-scale contextual modeling and long-tailed robustness without compromising practical training efficiency.

\subsection{Qualitative analyses of feature representations}
This subsection presents additional feature-level analyses using qualitative visualizations.
Specifically, we examine the final-layer feature maps, gradient-weighted class activation mapping~(Grad-CAM)~\cite{selvaraju2017grad}. 
% and t-distributed stochastic neighbor embedding~(t-SNE)~\cite{van2008visualizing} plots.

\begin{figure}[t!]
    \centering
    \includegraphics[scale=0.19]{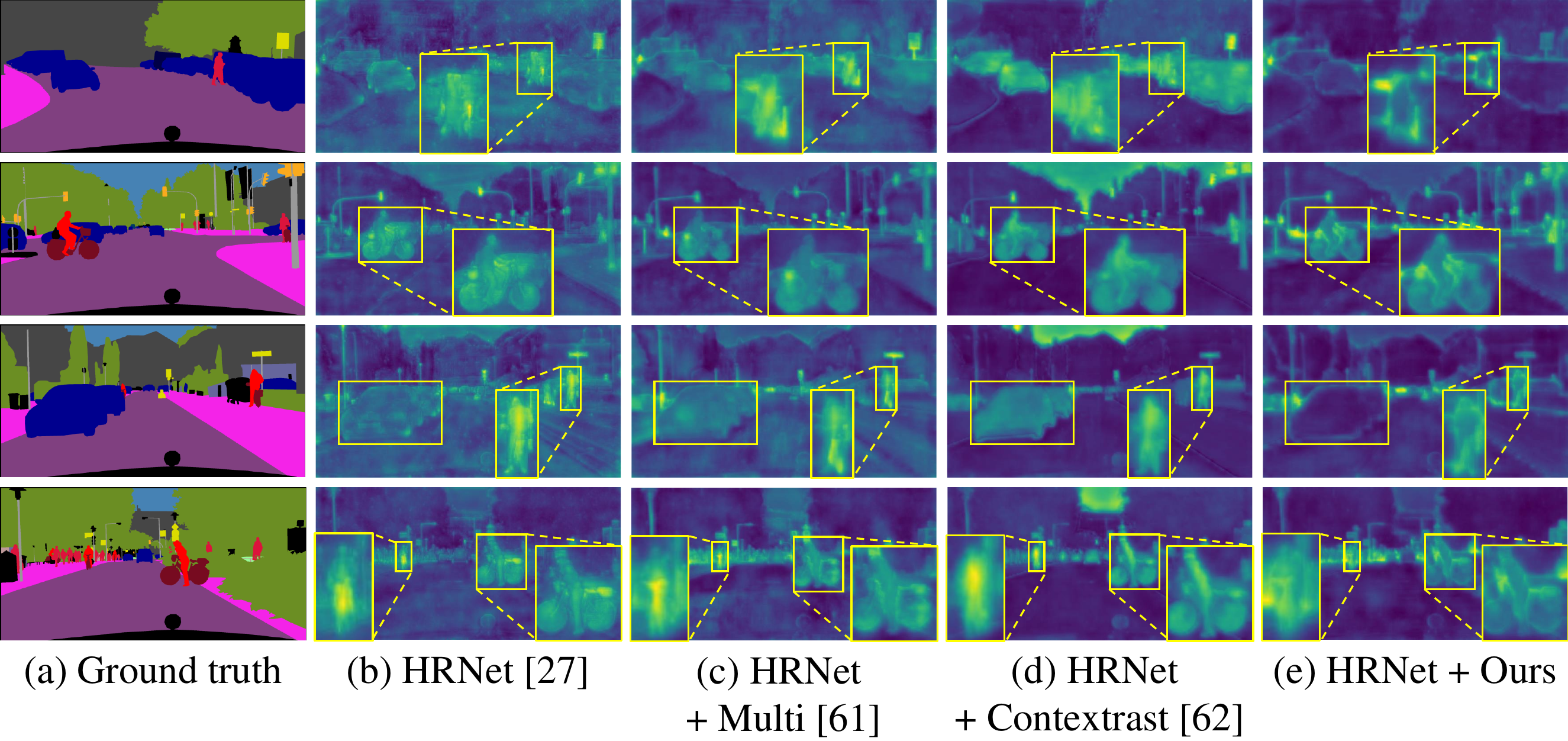}
    \caption{Feature map visualization on Cityscapes-\texttt{val}. (a) Ground truth, (b) HRNet~\cite{sun2019high}, (c) HRNet + Multi~\cite{pissas2022multi}, (d) HRNet + Contextrast~\cite{sung2024contextrast}, and (e) HRNet + Ours.
    The visualizations in (b)\textendash(e) show feature activation maps, where brighter (yellow) regions indicate stronger activations and greater model focus.
    Zoomed-in boxes highlight areas where our method attends more accurately to small objects and fine-grained boundaries~(best viewed in color).}
    \label{fig:city_featmap}
\end{figure}

\vspace{2mm}
\noindent\textbf{Visualization with feature map highlighting.}
As presented in Fig.~\ref{fig:city_featmap}, the feature activation maps also demonstrate that Contextrast++ achieves improvements in focus, precision, and adaptability compared with existing methods.
While HRNet with Multi or with Contextrast showed refined global contexts at the object level, these approaches occasionally lost fine-grained local details, particularly along object boundaries;
see the zoomed-in boxes that highlight activated features for the person and bicyclist classes in Fig.~\ref{fig:city_featmap}.
In contrast, Contextrast++ produced more emphasized and consistent object boundaries than Multi and Contextrast, owing to its ability to achieve a better balance between local and global contexts.
For instance, the bicyclist and motorcycle classes were more clearly distinguishable in the feature maps; see the second and fourth rows in Fig.~\ref{fig:city_featmap}.

% \begin{figure}[t]
%     \centering
%     \includegraphics[scale=0.16]{figures/grad_cam_paper9_pdf.pdf}
%     \caption{Gradient-weighted class activation mapping (Grad-CAM)~\cite{selvaraju2017grad} visualization on the Cityscapes-\texttt{val}. (a) Ground truth, (b) HRNet~\cite{sun2019high}, (c) HRNet + Multi~\cite{pissas2022multi}, (d) HRNet + Contextrast~\cite{sung2024contextrast}, and (e) HRNet + Ours.
%     The Grad-CAM heatmaps in (b)\textendash(e) highlight class-specific regions where the model attends during prediction. Red regions indicate stronger attention, while cooler colors represent lower attention levels.
%     Zoomed-in boxes emphasize fine-grained differences in attention, particularly for small objects (\eg person, truck) and class boundaries.
%     The leftmost labels ``Car,'' ``Person,'' and ``Truck'' indicate the target class for each visualization.
%     Notably, our method shows more precise activation focused on the target class, while suppressing irrelevant regions belonging to other classes~(best viewed in color).
%   }
%     \label{fig:grad_cam}
% \end{figure}
\begin{figure}[t]
    \centering
    \includegraphics[scale=0.16]{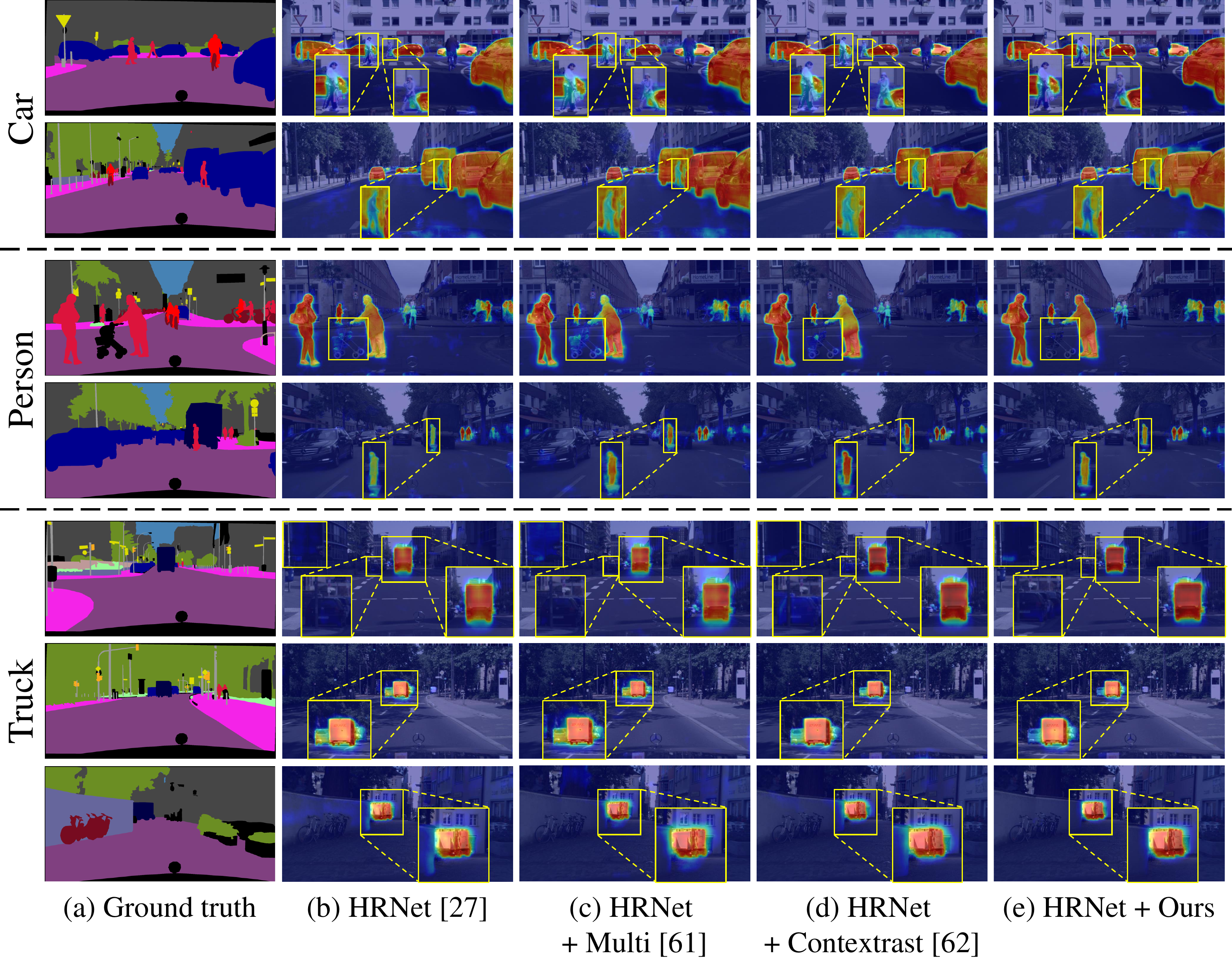}
    \caption{Grad-CAM~\cite{selvaraju2017grad} visualizations on Cityscapes-\texttt{val}. (a) Ground truth, (b) HRNet~\cite{sun2019high}, (c) HRNet + Multi~\cite{pissas2022multi}, (d) HRNet + Contextrast~\cite{sung2024contextrast}, and (e) HRNet + Ours. The leftmost labels `Car', `Person', and `Truck' indicate the target class for each row. Zoomed-in boxes highlight fine-grained differences in attention for small objects and class boundaries (best viewed in color).
  }
    \label{fig:grad_cam}
\end{figure}
\vspace{2mm}
\noindent\textbf{Visualization with Grad-CAM.}
Grad-CAM highlights class-specific key regions on the image plane where the model attends during prediction.
As illustrated in Fig.~\ref{fig:grad_cam}, Contextrast++ focused more precisely on relevant class regions while effectively disregarding irrelevant areas. 
For instance, in the first example of ``Car'' row in Fig.~\ref{fig:grad_cam}, Contextrast++ reduces its attention on the nearby person partially occluding the car, whereas other methods incorrectly focus on both the car and the occluding person, leading to confusion between classes.

Similarly, in the first example of the ``Person'' row, Contextrast++ completely ignores the stroller, whereas existing state-of-the-art methods incorrectly highlight it. 
Furthermore, from the second example of the ``Person'' row to all examples of the ``Truck'' row,
our Contextrast++ successfully focused sharply within object boundaries, avoiding over-expansion beyond the objects themselves.

% Therefore, these qualitative results support the effectiveness of Contextrast++ in refining feature representations, particularly for small objects and boundaries, and improving attention to relevant regions while ignoring neighboring objects.

% \begin{figure}[t!]
%     \includegraphics[width=0.48\textwidth]{figures/tsne_base_++.pdf}
%     \caption{t-distributed stochastic neighbor embedding~(t-SNE)~\cite{van2008visualizing} visualizations of feature representations obtained from (a) HRNet~\cite{sun2019high} as a baseline model and (b) Contextrast++ results on Cityscapes-\texttt{val}. Different colors represent distinct class labels. Visualizations are ordered from left to right, corresponding to the 1st to 4th layers.
%     The results show that with Contextrast++, the feature distributions for each class become more compact and better separated~(best viewed in color).}	
%     \label{fig:tsne_city}
% \end{figure}
\begin{figure}[t!]
    \centering
    \includegraphics[width=0.45\textwidth]{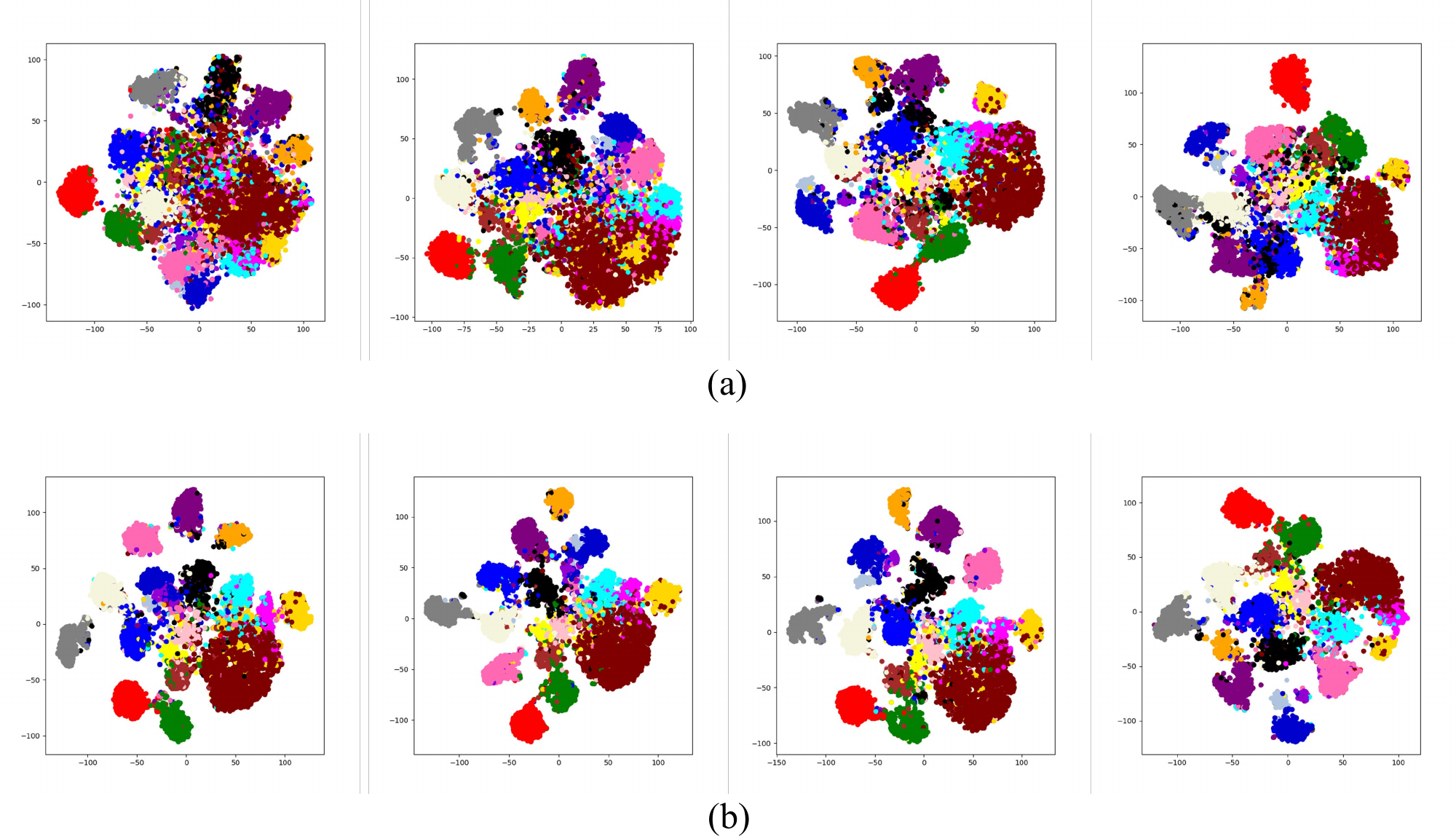}
    \caption{t-distributed stochastic neighbor embedding~(t-SNE)~\cite{van2008visualizing} visualizations of feature representations obtained from (a) HRNet~\cite{sun2019high} and (b) Contextrast++ on Cityscapes-\texttt{val}, ordered from 1st to 4th layers (left to right). Different colors represent distinct semantic classes. Contextrast++ yields more compact intra-class clusters and clearer inter-class separation across all layers (best viewed in color).}	
    \label{fig:tsne_city}
\end{figure}

\vspace{2mm}

\noindent\textbf{Visualization with t-SNE.} Finally, we qualitatively analyze the distribution of learned features across layers using t-distributed stochastic neighbor embedding~(t-SNE), as shown in Fig.~\ref{fig:tsne_city}. Compared with the baseline, Contextrast++ exhibits improved intra-class compactness and inter-class separation across all layers, reflecting the model's improved ability to incorporate global contextual information through the adaptive fusion module. These results demonstrate that our method effectively enhances semantic discriminability across scales, resulting in more robust and class-consistent feature representations.

    \section{Conclusion}

In this paper, we proposed Contextrast++, an enhanced contextual contrastive learning framework that effectively and adaptively integrates multi-scale information and addresses key challenges in semantic segmentation.
Building upon our previous work, Contextrast~\cite{sung2024contextrast}, which introduced CCL and boundary-aware negative (BANE) sampling,
our approach further enhances feature learning and segmentation accuracy through the PA loss with the adaptive fusion module, as well as the AA loss.

The adaptive fusion module replaces the static weighted sum used in Contextrast~\cite{sung2024contextrast} with a self-attention-based mechanism, thereby enabling adaptive integration of representative anchors across scales.
This allows the model to better balance local and global contexts, improving feature representation diversity while maintaining efficiency.
The AA loss ensures class-wise balanced learning, mitigating the long-tailed distribution issue by leveraging a memory bank with representative anchors instead of raw features.
Finally, BANE sampling enhances fine-grained feature learning by selecting hard negative samples along object boundaries, thereby improving segmentation accuracy, particularly in complex and ambiguous regions.

Through extensive experiments on five public datasets and various CNN and transformer backbones, we demonstrated that Contextrast++ outperforms prior contrastive learning methods on the majority of configurations, with the largest gains on long-tailed datasets and at object boundaries.
Ablation studies further validated the effectiveness of each proposed module, highlighting the contributions of our PA loss with the adaptive fusion module, AA loss, and BANE sampling. Despite these encouraging results, there is further space for improvement.
In future work, we plan to extend Contextrast++ to real-world robotic applications such as semantic mapping and navigation in urban environments.
In these settings, robust and fine-grained semantic understanding is essential for tasks such as obstacle avoidance, scene interpretation, and long-term mapping.

}

{
    \small
    \bibliographystyle{IEEEtran}
    \bibliography{main}

@String(AAAI = {AAAI})

@inproceedings{deng2009imagenet,
  title={{ImageNet: A large-scale hierarchical image database}},
  author={Deng, Jia and Dong, Wei and Socher, Richard and Li, Li-Jia and Li, Kai and Fei-Fei, Li},
  booktitle={Proc. IEEE/CVF Conference on Computer Vision and Pattern Recognition},
  pages={248--255},
  year={2009}
}

@inproceedings{cordts2016cityscapes,
  title={The {C}ityscapes dataset for semantic urban scene understanding},
  author={Cordts, Marius and Omran, Mohamed and Ramos, Sebastian and Rehfeld, Timo and Enzweiler, Markus and Benenson, Rodrigo and Franke, Uwe and Roth, Stefan and Schiele, Bernt},
  booktitle={Proc. IEEE/CVF Conference on Computer Vision and Pattern Recognition},
  pages={3213--3223},
  year={2016}
}

@inproceedings{zhou2017scene,
  title={{Scene parsing through ADE20K dataset}},
  author={Zhou, Bolei and Zhao, Hang and Puig, Xavier and Fidler, Sanja and Barriuso, Adela and Torralba, Antonio},
  booktitle={Proc. IEEE/CVF Conference on Computer Vision and Pattern Recognition},
  pages={633--641},
  year={2017}
}

@article{chen2017deeplab,
  title={{DeepLab: Semantic image segmentation with deep convolutional nets, atrous convolution, and fully connected CRFs}},
  author={Chen, Liang Chieh and Papandreou, George and Kokkinos, Iasonas and Murphy, Kevin and Yuille, Alan L},
  journal={IEEE Transactions on Pattern Analysis and Machine Intelligence},
  volume={40},
  number={4},
  pages={834--848},
  year={2017},
  publisher={IEEE}
}

@article{chen2017rethinking,
  title={{Rethinking atrous convolution for semantic image segmentation}},
  author={Chen, Liang Chieh and Papandreou, George and Schroff, Florian and Adam, Hartwig},
  journal={arXiv preprint arXiv:1706.05587},
  year={2017}
}

@inproceedings{chen2018encoder,
  title={{Encoder-decoder with atrous separable convolution for semantic image segmentation}},
  author={Chen, Liang-Chieh and Zhu, Yukun and Papandreou, George and Schroff, Florian and Adam, Hartwig},
  booktitle={Proc. European Conference on Computer Vision},
  pages={801--818},
  year={2018}
}

@inproceedings{zhao2017pyramid,
  title={{Pyramid scene parsing network}},
  author={Zhao, Hengshuang and Shi, Jianping and Qi, Xiaojuan and Wang, Xiaogang and Jia, Jiaya},
  booktitle={Proc. IEEE/CVF Conference on Computer Vision and Pattern Recognition},
  pages={2881--2890},
  year={2017}
}

@inproceedings{xiao2018unified,
  title={{Unified perceptual parsing for scene understanding}},
  author={Xiao, Tete and Liu, Yingcheng and Zhou, Bolei and Jiang, Yuning and Sun, Jian},
  booktitle={Proc. European Conference on Computer Vision},
  pages={418--434},
  year={2018}
}

@inproceedings{yurtkulu2019semantic,
  title={{Semantic segmentation with extended {DeepLabv3} architecture}},
  author={Yurtkulu, Salih Can and {\c{S}}ahin, Yusuf H{\"u}seyin and Unal, Gozde},
  booktitle={Proc. Signal Processing and Communications Applications Conference},
  pages={1--4},
  year={2019},
}

@inproceedings{yuan2020object,
  title={Object-contextual representations for semantic segmentation},
  author={Yuan, Yuhui and Chen, Xilin and Wang, Jingdong},
  booktitle={Proc. European Conference on Computer Vision},
  pages={173--190},
  year={2020},
}

@article{khosla2020supervised,
  title={Supervised contrastive learning},
  author={Khosla, Prannay and Teterwak, Piotr and Wang, Chen and Sarna, Aaron and Tian, Yonglong and Isola, Phillip and Maschinot, Aaron and Liu, Ce and Krishnan, Dilip},
  journal={Advances in Neural Information Processing Systems},
  volume={33},
  pages={18661--18673},
  year={2020}
}

@inproceedings{wang2021exploring,
  title={Exploring cross-image pixel contrast for semantic segmentation},
  author={Wang, Wenguan and Zhou, Tianfei and Yu, Fisher and Dai, Jifeng and Konukoglu, Ender and Van Gool, Luc},
  booktitle={Proc. IEEE/CVF International Conference on Computer Vision},
  pages={7303--7313},
  year={2021}
}

@inproceedings{pissas2022multi,
  title={Multi-scale and cross-scale contrastive learning for semantic segmentation},
  author={Pissas, Theodoros and Ravasio, Claudio S and Cruz, Lyndon Da and Bergeles, Christos},
  booktitle={Proc. European Conference on Computer Vision},
  pages={413--429},
  year={2022},
}

@inproceedings{hu2021region,
  title={Region-aware contrastive learning for semantic segmentation},
  author={Hu, Hanzhe and Cui, Jinshi and Wang, Liwei},
  booktitle={Proc. IEEE/CVF International Conference on Computer Vision},
  pages={16291--16301},
  year={2021}
}

@inproceedings{he2020momentum,
  title={Momentum contrast for unsupervised visual representation learning},
  author={He, Kaiming and Fan, Haoqi and Wu, Yuxin and Xie, Saining and Girshick, Ross},
  booktitle={Proc. IEEE/CVF Conference on Computer Vision and Pattern Recognition},
  pages={9729--9738},
  year={2020}
}

@inproceedings{long2015fully,
  title={Fully convolutional networks for semantic segmentation},
  author={Long, Jonathan and Shelhamer, Evan and Darrell, Trevor},
  booktitle={Proc. IEEE/CVF Conference on Computer Vision and Pattern Recognition},
  pages={3431--3440},
  year={2015}
}

@article{brostow2009semantic,
  title={Semantic object classes in video: A high-definition ground truth database},
  author={Brostow, Gabriel J and Fauqueur, Julien and Cipolla, Roberto},
  journal={Pattern Recognition Letters},
  volume={30},
  number={2},
  pages={88--97},
  year={2009},
  publisher={Elsevier}
}

@inproceedings{caesar2018coco,
  title={{COCO-stuff: Thing and stuff classes in context}},
  author={Caesar, Holger and Uijlings, Jasper and Ferrari, Vittorio},
  booktitle={Proc. IEEE/CVF Conference on Computer Vision and Pattern Recognition},
  pages={1209--1218},
  year={2018}
}

@inproceedings{mottaghi2014role,
  title={The role of context for object detection and semantic segmentation in the wild},
  author={Mottaghi, Roozbeh and Chen, Xianjie and Liu, Xiaobai and Cho, Nam-Gyu and Lee, Seong-Whan and Fidler, Sanja and Urtasun, Raquel and Yuille, Alan},
  booktitle={Proc. IEEE/CVF Conference on Computer Vision and Pattern Recognition},
  pages={891--898},
  year={2014}
}

@inproceedings{zhong2023understanding,
  title={Understanding imbalanced semantic segmentation through neural collapse},
  author={Zhong, Zhisheng and Cui, Jiequan and Yang, Yibo and Wu, Xiaoyang and Qi, Xiaojuan and Zhang, Xiangyu and Jia, Jiaya},
  booktitle={Proc. IEEE/CVF Conference on Computer Vision and Pattern Recognition},
  pages={19550--19560},
  year={2023}
}

@inproceedings{yuan2020segfix,
  title={{SegFix: Model-agnostic boundary refinement for segmentation}},
  author={Yuan, Yuhui and Xie, Jingyi and Chen, Xilin and Wang, Jingdong},
  booktitle={Proc. European Conference on Computer Vision},
  pages={489--506},
  year={2020},
}

@inproceedings{wang2022active,
  title={Active boundary loss for semantic segmentation},
  author={Wang, Chi and Zhang, Yunke and Cui, Miaomiao and Ren, Peiran and Yang, Yin and Xie, Xuansong and Hua, Xian-Sheng and Bao, Hujun and Xu, Weiwei},
  booktitle={Proc. AAAI Conference on Artificial Intelligence},
  volume={36},
  number={2},
  pages={2397--2405},
  year={2022}
}

@article{sun2019high,
  title={High-resolution representations for labeling pixels and regions},
  author={Sun, Ke and Zhao, Yang and Jiang, Borui and Cheng, Tianheng and Xiao, Bin and Liu, Dong and Mu, Yadong and Wang, Xinggang and Liu, Wenyu and Wang, Jingdong},
  journal={arXiv preprint arXiv:1904.04514},
  year={2019}
}

@article{hong2021deep,
  title={Deep dual-resolution networks for real-time and accurate semantic segmentation of road scenes},
  author={Hong, Yuanduo and Pan, Huihui and Sun, Weichao and Jia, Yisong},
  journal={arXiv preprint arXiv:2101.06085},
  year={2021}
}

@inproceedings{gutmann2010noise,
  title={Noise-contrastive estimation: A new estimation principle for unnormalized statistical models},
  author={Gutmann, Michael and Hyv{\"a}rinen, Aapo},
  booktitle={Proc. International Conference on Artificial Intelligence and Statistics},
  pages={297--304},
  year={2010},
}

@article{oord2018representation,
  title={Representation learning with contrastive predictive coding},
  author={Oord, Aaron van den and Li, Yazhe and Vinyals, Oriol},
  journal={arXiv preprint arXiv:1807.03748},
  year={2018}
}

@article{kimmel1996sub,
  title={Sub-pixel distance maps and weighted distance transforms},
  author={Kimmel, Ron and Kiryati, Nahum and Bruckstein, Alfred M},
  journal={Journal of Mathematical Imaging and Vision},
  volume={6},
  pages={223--233},
  year={1996},
  publisher={Springer}
}

@inproceedings{li2022targeted,
  title={Targeted supervised contrastive learning for long-tailed recognition},
  author={Li, Tianhong and Cao, Peng and Yuan, Yuan and Fan, Lijie and Yang, Yuzhe and Feris, Rogerio S and Indyk, Piotr and Katabi, Dina},
  booktitle={Proc. IEEE/CVF Conference on Computer Vision and Pattern Recognition},
  pages={6918--6928},
  year={2022}
}

@article{li2018pyramid,
  title={Pyramid attention network for semantic segmentation},
  author={Li, Hanchao and Xiong, Pengfei and An, Jie and Wang, Lingxue},
  journal={arXiv preprint arXiv:1805.10180},
  year={2018}
}

@inproceedings{ke2018adaptive,
  title={Adaptive affinity fields for semantic segmentation},
  author={Ke, Tsung-Wei and Hwang, Jyh-Jing and Liu, Ziwei and Yu, Stella X},
  booktitle={Proc. European Conference on Computer Vision},
  pages={587--602},
  year={2018}
}

@inproceedings{yu2018learning,
  title={Learning a discriminative feature network for semantic segmentation},
  author={Yu, Changqian and Wang, Jingbo and Peng, Chao and Gao, Changxin and Yu, Gang and Sang, Nong},
  booktitle={Proc. IEEE/CVF Conference on Computer Vision and Pattern Recognition},
  pages={1857--1866},
  year={2018}
}

@inproceedings{yu2020context,
  title={Context prior for scene segmentation},
  author={Yu, Changqian and Wang, Jingbo and Gao, Changxin and Yu, Gang and Shen, Chunhua and Sang, Nong},
  booktitle={Proc. IEEE/CVF Conference on Computer Vision and Pattern Recognition},
  pages={12416--12425},
  year={2020}
}

@inproceedings{fu2019dual,
  title={Dual attention network for scene segmentation},
  author={Fu, Jun and Liu, Jing and Tian, Haijie and Li, Yong and Bao, Yongjun and Fang, Zhiwei and Lu, Hanqing},
  booktitle={Proc. IEEE/CVF Conference on Computer Vision and Pattern Recognition},
  pages={3146--3154},
  year={2019}
}

@inproceedings{choi2020cars,
  title={{Cars can't fly up in the sky: Improving urban-scene segmentation via height-driven attention networks}},
  author={Choi, Sungha and Kim, Joanne T and Choo, Jaegul},
  booktitle={Proc. IEEE/CVF Conference on Computer Vision and Pattern Recognition},
  pages={9373--9383},
  year={2020}
}

@inproceedings{huynh2021progressive,
  title={Progressive semantic segmentation},
  author={Huynh, Chuong and Tran, Anh Tuan and Luu, Khoa and Hoai, Minh},
  booktitle={Proc. {IEEE}/CVF Conference on Computer Vision and Pattern Recognition},
  pages={16755--16764},
  year={2021}
}

@inproceedings{li2022deep,
  title={Deep hierarchical semantic segmentation},
  author={Li, Liulei and Zhou, Tianfei and Wang, Wenguan and Li, Jianwu and Yang, Yi},
  booktitle={Proc. IEEE/CVF Conference on Computer Vision and Pattern Recognition},
  pages={1246--1257},
  year={2022}
}

@inproceedings{strudel2021segmenter,
  title={Segmenter: Transformer for semantic segmentation},
  author={Strudel, Robin and Garcia, Ricardo and Laptev, Ivan and Schmid, Cordelia},
  booktitle={Proc. IEEE/CVF International Conference on Computer Vision},
  pages={7262--7272},
  year={2021}
}

@inproceedings{liu2021swin,
  title={{Swin Transformer: Hierarchical vision transformer using shifted windows}},
  author={Liu, Ze and Lin, Yutong and Cao, Yue and Hu, Han and Wei, Yixuan and Zhang, Zheng and Lin, Stephen and Guo, Baining},
  booktitle={Proc. IEEE/CVF International Conference on Computer Vision},
  pages={10012--10022},
  year={2021}
}

@inproceedings{woo2023convnext,
  title={{ConvNext v2: Co-designing and scaling ConvNets with masked autoencoders}},
  author={Woo, Sanghyun and Debnath, Shoubhik and Hu, Ronghang and Chen, Xinlei and Liu, Zhuang and Kweon, In So and Xie, Saining},
  booktitle={Proc. IEEE/CVF Conference on Computer Vision and Pattern Recognition},
  pages={16133--16142},
  year={2023}
}

@inproceedings{sanderson2022fcn,
  title={{FCN-transformer feature fusion for polyp segmentation}},
  author={Sanderson, Edward and Matuszewski, Bogdan J},
  booktitle={Proc. Annual Conference on Medical Image Understanding and Analysis},
  pages={892--907},
  year={2022},
  
}

@article{hurtado2022semantic,
  title={Semantic scene segmentation for robotics},
  author={Hurtado, Juana Valeria and Valada, Abhinav},
  journal={Deep Learning for Robot Perception and Cognition},
  pages={279--311},
  year={2022}
}

@inproceedings{tzelepi2021semantic,
  title={Semantic scene segmentation for robotics applications},
  author={Tzelepi, Maria and Tefas, Anastasios},
  booktitle={Proc. International Conference on Information, Intelligence, Systems \& Applications},
  pages={1--4},
  year={2021}
}

@article{tan2022semantic,
  title={Semantic diffusion network for semantic segmentation},
  author={Tan, Haoru and Wu, Sitong and Pi, Jimin},
  journal={Advances in Neural Information Processing Systems},
  volume={35},
  pages={8702--8716},
  year={2022}
}

@article{kalantidis2020hard,
  title={Hard negative mixing for contrastive learning},
  author={Kalantidis, Yannis and Sariyildiz, Mert Bulent and Pion, Noe and Weinzaepfel, Philippe and Larlus, Diane},
  journal={Advances in Neural Information Processing Systems},
  volume={33},
  pages={21798--21809},
  year={2020}
}

@inproceedings{schroff2015facenet,
  title={{FaceNet: A unified embedding for face recognition and clustering}},
  author={Schroff, Florian and Kalenichenko, Dmitry and Philbin, James},
  booktitle={Proc. IEEE/CVF Conference on Computer Vision and Pattern Recognition},
  pages={815--823},
  year={2015}
}

@article{xie2022delving,
  title={Delving into inter-image invariance for unsupervised visual representations},
  author={Xie, Jiahao and Zhan, Xiaohang and Liu, Ziwei and Ong, Yew-Soon and Loy, Chen Change},
  journal={International Journal of Computer Vision},
  volume={130},
  number={12},
  pages={2994--3013},
  year={2022},
  publisher={Springer}
}

@article{cai2020all,
  title={Are all negatives created equal in contrastive instance discrimination?},
  author={Cai, Tiffany Tianhui and Frankle, Jonathan and Schwab, David J and Morcos, Ari S},
  journal={arXiv preprint arXiv:2010.06682},
  year={2020}
}

@inproceedings{cheng2022masked,
  title={Masked-attention mask transformer for universal image segmentation},
  author={Cheng, Bowen and Misra, Ishan and Schwing, Alexander G and Kirillov, Alexander and Girdhar, Rohit},
  booktitle={Proc. IEEE/CVF Conference on Computer Vision and Pattern Recognition},
  pages={1290--1299},
  year={2022}
}

@article{zhou2024cross,
  title={Cross-image pixel contrasting for semantic segmentation},
  author={Zhou, Tianfei and Wang, Wenguan},
  journal={IEEE Transactions on Pattern Analysis and Machine Intelligence},
  year={2024},
  volume={46},
  number={8},
  pages={5398--5412},
}

@inproceedings{sung2024contextrast,
  title={Contextrast: Contextual Contrastive Learning for Semantic Segmentation},
  author={Sung, Changki and Kim, Wanhee and An, Jungho and Lee, Wooju and Lim, Hyungtae and Myung, Hyun},
  booktitle={Proc. IEEE/CVF Conference on Computer Vision and Pattern Recognition},
  pages={3732--3742},
  year={2024}
}

@misc{mmseg2020,
    title={{MMSegmentation}: {OpenMMLab} Semantic Segmentation Toolbox and Benchmark},
    author={MMSegmentation Contributors},
    howpublished = {\url{https://github.com/open-mmlab/mmsegmentation}},
    year={2020}
}

@article{yang2024lcfnets,
  title={{LCFNets: Compensation strategy for real-time semantic segmentation of autonomous driving}},
  author={Yang, Lu and Bai, Yiwen and Ren, Fenglei and Bi, Chongke and Zhang, Ronghui},
  journal={IEEE Transactions on Intelligent Vehicles},
  year={2024},
  volume={9},
  number={4},
  pages={4715--4729},
}

@article{ni2023robust,
  title={Robust {3D} semantic segmentation based on multi-phase multi-modal fusion for intelligent vehicles},
  author={Ni, Peizhou and Li, Xu and Xu, Wang and Kong, Dong and Hu, Yue and Wei, Kun},
  journal={IEEE Transactions on Intelligent Vehicles},
  volume={9},
  number={1},
  pages={1602--1614},
  year={2023}
}

@article{feng2024segmentation,
  title={Segmentation of Road Negative Obstacles Based on Dual Semantic-feature Complementary Fusion for Autonomous Driving},
  author={Feng, Zhen and Guo, Yanning and Sun, Yuxiang},
  journal={IEEE Transactions on Intelligent Vehicles},
  year={2024},
  volume={9},
  number={4},
  pages={4687--4697},
  publisher={IEEE}
}

@article{gu2024clft,
  title={{CLFT: camera-LiDAR fusion transformer for semantic segmentation in autonomous driving}},
  author={Gu, Junyi and Bellone, Mauro and Pivo{\v{n}}ka, Tom{\'a}{\v{s}} and Sell, Raivo},
  journal={arXiv preprint arXiv:2404.17793},
  year={2024}
}

@article{wu2024s,
  title={{$S^3$M-Net: joint learning of semantic segmentation and stereo matching for autonomous driving}},
  author={Wu, Zhiyuan and Feng, Yi and Liu, Chuang-Wei and Yu, Fisher and Chen, Qijun and Fan, Rui},
  journal={IEEE Transactions on Intelligent Vehicles},
  year={2024},
  volume={9},
  number={2},
  pages={3940--3951},
  publisher={IEEE}
}

@article{liang2024multi,
  title={Multi-branch Differential Bidirectional Fusion Network for {RGB-T} Semantic Segmentation},
  author={Liang, Wenli and Shan, Caifeng and Yang, Yuanjian and Han, Jungong},
  journal={IEEE Transactions on Intelligent Vehicles},
  year={2025},
  volume={10},
  number={4},
  pages={2362--2372},
  publisher={IEEE}
}

@article{fan2022mlfnet,
  title={{MLFNet: Multi-level fusion network for real-time semantic segmentation of autonomous driving}},
  author={Fan, Jiaqi and Wang, Fei and Chu, Hongqing and Hu, Xiao and Cheng, Yifan and Gao, Bingzhao},
  journal={IEEE Transactions on Intelligent Vehicles},
  volume={8},
  number={1},
  pages={756--767},
  year={2022},
  publisher={IEEE}
}

@article{ziwen2023multi,
  title={Multi-objective Neural Architecture Search for Efficient and Fast Semantic Segmentation on Edge},
  author={ZiWen, Dou and Dong, Ye},
  journal={IEEE Transactions on Intelligent Vehicles},
  year={2023},
  volume={9},
  number={1},
  pages={1346--1357},
  publisher={IEEE}
}

@article{ye2024invpt++,
  title={{InvPT++}: Inverted pyramid multi-task transformer for visual scene understanding},
  author={Ye, Hanrong and Xu, Dan},
  journal={IEEE Transactions on Pattern Analysis and Machine Intelligence},
  volume={46},
  number={12},
  pages={7493--7508},
  year={2024},
  publisher={IEEE}
}

@article{xu2024mctformer+,
  title={{MCTformer+}: Multi-class token transformer for weakly supervised semantic segmentation},
  author={Xu, Lian and Bennamoun, Mohammed and Boussaid, Farid and Laga, Hamid and Ouyang, Wanli and Xu, Dan},
  journal={IEEE Transactions on Pattern Analysis and Machine Intelligence},
  year={2024},
  volume={46},
  number={12},
  pages={8380--8395}
}

@article{zhou2024object,
  title={Object-Centric Representation Learning for Video Scene Understanding},
  author={Zhou, Yi and Zhang, Hui and Park, Seung-In and Yoo, ByungIn and Qi, Xiaojuan},
  journal={IEEE Transactions on Pattern Analysis and Machine Intelligence},
  year={2024},
  volume={46},
  number={12},
  pages={8410--8423}
}

@article{chen2024frequency,
  title={Frequency-aware feature fusion for dense image prediction},
  author={Chen, Linwei and Fu, Ying and Gu, Lin and Yan, Chenggang and Harada, Tatsuya and Huang, Gao},
  journal={IEEE Transactions on Pattern Analysis and Machine Intelligence},
  year={2024},
  volume={46},
  number={12},
  pages={10763--10780}
}

@article{zhou2024prototype,
  title={Prototype-based semantic segmentation},
  author={Zhou, Tianfei and Wang, Wenguan},
  journal={IEEE Transactions on Pattern Analysis and Machine Intelligence},
  year={2024},
  volume={46},
  number={10},
  pages={6858--6872},
}

@inproceedings{sandler2018mobilenetv2,
  title={{MobileNetV2: Inverted residuals and linear bottlenecks}},
  author={Sandler, Mark and Howard, Andrew and Zhu, Menglong and Zhmoginov, Andrey and Chen, Liang-Chieh},
  booktitle={Proc. IEEE/CVF Conference on Computer Vision and Pattern Recognition},
  pages={4510--4520},
  year={2018}
}

@article{dosovitskiy2020image,
  title={An image is worth 16x16 words: Transformers for image recognition at scale},
  author={Dosovitskiy, Alexey and Beyer, Lucas and Kolesnikov, Alexander and Weissenborn, Dirk and Zhai, Xiaohua and Unterthiner, Thomas and Dehghani, Mostafa and Minderer, Matthias and Heigold, Georg and Gelly, Sylvain and others},
  journal={arXiv preprint arXiv:2010.11929},
  year={2020}
}

@inproceedings{selvaraju2017grad,
  title={Grad-{CAM}: Visual explanations from deep networks via gradient-based localization},
  author={Selvaraju, Ramprasaath R and Cogswell, Michael and Das, Abhishek and Vedantam, Ramakrishna and Parikh, Devi and Batra, Dhruv},
  booktitle={Proc. IEEE/CVF International Conference on Computer Vision},
  pages={618--626},
  year={2017}
}

@article{van2008visualizing,
  title={Visualizing data using {t-SNE}.},
  author={Van der Maaten, Laurens and Hinton, Geoffrey},
  journal={Journal of Machine Learning Research},
  volume={9},
  number={11},
  pages={2579--2605},
  year={2008}
}

@article{vaswani2017attention,
  title={Attention is all you need},
  author={Vaswani, Ashish and Shazeer, Noam and Parmar, Niki and Uszkoreit, Jakob and Jones, Llion and Gomez, Aidan N and Kaiser, {\L}ukasz and Polosukhin, Illia},
  journal={Advances in Neural Information Processing Systems},
  volume={30},
  year={2017}
}

@inproceedings{xuan2020hard,
  title={Hard negative examples are hard, but useful},
  author={Xuan, Hong and Stylianou, Abby and Liu, Xiaotong and Pless, Robert},
  booktitle={Proc. European Conference on Computer Vision},
  pages={126--142},
  year={2020}
}

@article{xie2021segformer,
  title={SegFormer: Simple and efficient design for semantic segmentation with transformers},
  author={Xie, Enze and Wang, Wenhai and Yu, Zhiding and Anandkumar, Anima and Alvarez, Jose M and Luo, Ping},
  journal={Advances in Neural Information Processing Systems},
  volume={34},
  pages={12077--12090},
  year={2021}
}

@article{zhou2024boundary,
  title={Boundary-guided lightweight semantic segmentation with multi-scale semantic context},
  author={Zhou, Quan and Wang, Linjie and Gao, Guangwei and Kang, Bin and Ou, Weihua and Lu, Huimin},
  journal={IEEE Transactions on Multimedia},
  volume={26},
  pages={7887--7900},
  year={2024},
  publisher={IEEE}
}

@inproceedings{shi2023transformer,
  title={Transformer scale gate for semantic segmentation},
  author={Shi, Hengcan and Hayat, Munawar and Cai, Jianfei},
  booktitle={Proc. IEEE/CVF Conference on Computer Vision and Pattern Recognition},
  pages={3051--3060},
  year={2023}
}

@inproceedings{gu2022multi,
  title={Multi-scale high-resolution vision transformer for semantic segmentation},
  author={Gu, Jiaqi and Kwon, Hyoukjun and Wang, Dilin and Ye, Wei and Li, Meng and Chen, Yu-Hsin and Lai, Liangzhen and Chandra, Vikas and Pan, David Z},
  booktitle={Proc. IEEE/CVF Conference on Computer Vision and Pattern Recognition},
  pages={12094--12103},
  year={2022}
}

@article{li2021ctnet,
  title={{CTNet: Context-based tandem network for semantic segmentation}},
  author={Li, Zechao and Sun, Yanpeng and Zhang, Liyan and Tang, Jinhui},
  journal={IEEE Transactions on Pattern Analysis and Machine Intelligence},
  volume={44},
  number={12},
  pages={9904--9917},
  year={2021},
  publisher={IEEE}
}

@inproceedings{liu2013weakly,
  title={Weakly-supervised dual clustering for image semantic segmentation},
  author={Liu, Yang and Liu, Jing and Li, Zechao and Tang, Jinhui and Lu, Hanqing},
  booktitle={Proc. IEEE/CVF Conference on Computer Vision and Pattern Recognition},
  pages={2075--2082},
  year={2013}
}

@article{zhu2025merging,
  title={Merging context clustering with visual state space models for medical image segmentation},
  author={Zhu, Yun and Zhang, Dong and Lin, Yi and Feng, Yifei and Tang, Jinhui},
  journal={IEEE Transactions on Medical Imaging},
  volume={44},
  number={5},
  pages={2131--2142},
  year={2025},
  publisher={IEEE}
}

@article{zhang2020causal,
  title={Causal intervention for weakly-supervised semantic segmentation},
  author={Zhang, Dong and Zhang, Hanwang and Tang, Jinhui and Hua, Xian-Sheng and Sun, Qianru},
  journal={Advances in Neural Information Processing Systems},
  volume={33},
  pages={655--666},
  year={2020}
}

@inproceedings{pei2022hierarchical,
  title={Hierarchical feature alignment network for unsupervised video object segmentation},
  author={Pei, Gensheng and Shen, Fumin and Yao, Yazhou and Xie, Guo-Sen and Tang, Zhenmin and Tang, Jinhui},
  booktitle={Proc. European Conference on Computer Vision},
  pages={596--613},
  year={2022},
}

@article{chen2023multi,
  title={Multi-granularity denoising and bidirectional alignment for weakly supervised semantic segmentation},
  author={Chen, Tao and Yao, Yazhou and Tang, Jinhui},
  journal={IEEE Transactions on Image Processing},
  volume={32},
  pages={2960--2971},
  year={2023},
  publisher={IEEE}
}

@inproceedings{zhang2021self,
  title={Self-regulation for semantic segmentation},
  author={Zhang, Dong and Zhang, Hanwang and Tang, Jinhui and Hua, Xian-Sheng and Sun, Qianru},
  booktitle={Proc. IEEE/CVF International Conference on Computer Vision},
  pages={6953--6963},
  year={2021}
}

@article{chen2024spatial,
  title={Spatial structure constraints for weakly supervised semantic segmentation},
  author={Chen, Tao and Yao, Yazhou and Huang, Xingguo and Li, Zechao and Nie, Liqiang and Tang, Jinhui},
  journal={IEEE Transactions on Image Processing},
  volume={33},
  pages={1136--1148},
  year={2024},
  publisher={IEEE}
}

@inproceedings{fu2025segman,
  title={{SegMAN}: Omni-scale context modeling with state space models and local attention for semantic segmentation},
  author={Fu, Yunxiang and Lou, Meng and Yu, Yizhou},
  booktitle={Proc. IEEE/CVF Conference on Computer Vision and Pattern Recognition},
  pages={19077--19087},
  year={2025}
}

@article{shi2025llmformer,
  title={{LLMFormer}: Large language model for open-vocabulary semantic segmentation},
  author={Shi, Hengcan and Dao, Son Duy and Cai, Jianfei},
  journal={International Journal of Computer Vision},
  volume={133},
  number={2},
  pages={742--759},
  year={2025},
  publisher={Springer}
}

@article{yang2025m,
  title={{M-SEE}: A multi-scale encoder enhancement framework for end-to-end Weakly Supervised Semantic Segmentation},
  author={Yang, Ziqian and Zhao, Xinqiao and Yao, Chao and Zhang, Quan and Xiao, Jimin},
  journal={Pattern Recognition},
  volume={162},
  pages={111348},
  year={2025},
  publisher={Elsevier}
}

@article{hutchinson1989stochastic,
  title={A stochastic estimator of the trace of the influence matrix for Laplacian smoothing splines},
  author={Hutchinson, Michael F},
  journal={Communications in Statistics-Simulation and Computation},
  volume={18},
  number={3},
  pages={1059--1076},
  year={1989},
  publisher={Taylor \& Francis}
}

@article{bekas2007estimator,
  title={An estimator for the diagonal of a matrix},
  author={Bekas, Costas and Kokiopoulou, Effrosyni and Saad, Yousef},
  journal={Applied numerical mathematics},
  volume={57},
  number={11-12},
  pages={1214--1229},
  year={2007},
  publisher={Elsevier}
}
}
\begin{IEEEbiography}[{\includegraphics[width=1in,height=1.25in,clip,keepaspectratio]{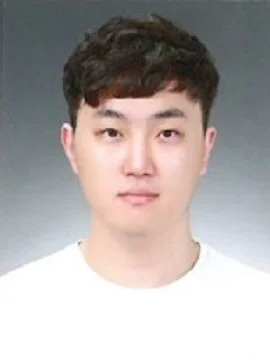}}]{Changki Sung}
  received the B.S. degree in electrical and computer engineering from the University of New Hampshire and M.S. in civil engineering and Ph.D. degrees in robotics program from the Korea Advanced Institute of Science and Technology (KAIST), Daejeon, Republic of Korea, in 2018, 2021, and 2025, respectively. He is currently a postdoctoral researcher with Information \& Electronics Research Institute, KAIST. His research interests include vision-language navigation, vision-language action, semantic segmentation, and spatial AI.
\end{IEEEbiography}
\vspace*{-30pt}
\begin{IEEEbiography}[{\includegraphics[width=1in,height=1.25in,clip,keepaspectratio]{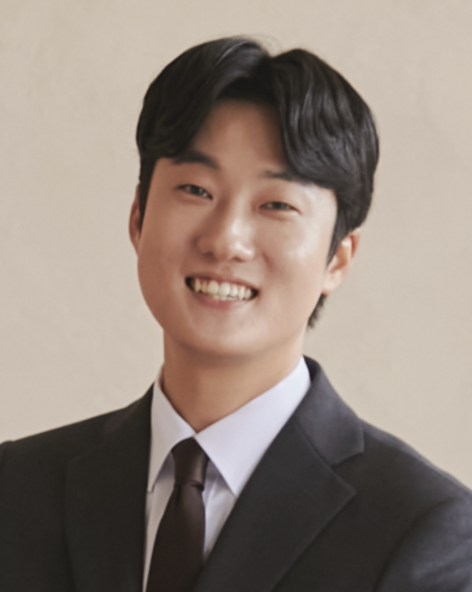}}]{Hyungtae Lim}
	received the B.S. degree in mechanical engineering, and M.S. and Ph.D. degrees in electrical engineering from the Korea Advanced Institute of Science and Technology (KAIST), Daejeon, Republic of Korea, in 2018, 2020, and 2023, respectively.
    He is currently a postdoctoral associate in the Laboratory for Information \& Decision Systems~(LIDS), Massachusetts Institute of Technology~(MIT), Massachusetts, USA.
    His research interests include SLAM (simultaneous localization and mapping), 3D registration, 3D perception, long-term map management, spatial AI, and deep learning.
\end{IEEEbiography}
\vspace*{-30pt}
\begin{IEEEbiography}[{\includegraphics[width=1in,height=1.25in,clip,keepaspectratio]{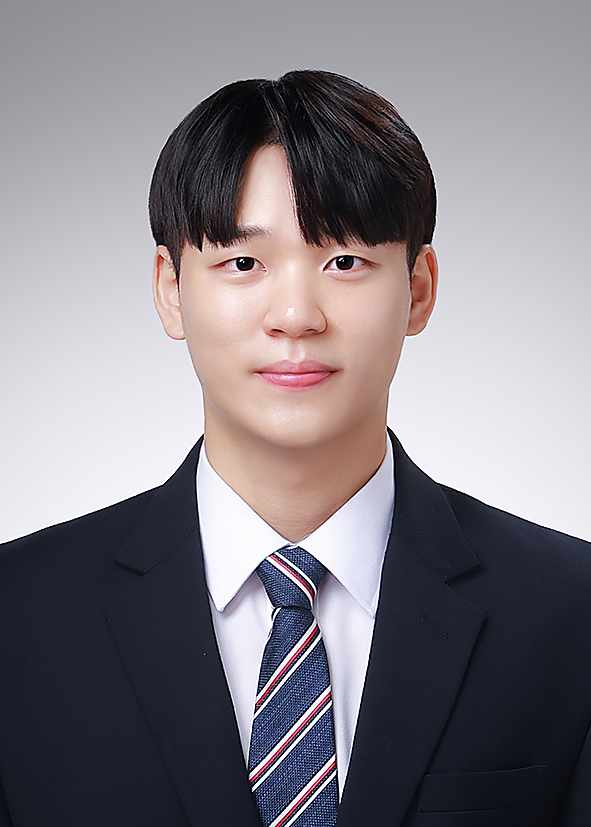}}]{Wanhee Kim}
  received the B.S. degree in Automobile and IT Convergence from Kookmin University, Seoul, Republic of Korea, in 2024.
  He is currently pursuing the M.S. degree in the Robotics Program at the Korea Advanced Institute of Science and Technology (KAIST).
  His research interests include robotics and semantic segmentation, VLM-based scene understanding, and spatial AI.
\end{IEEEbiography}
\vspace*{-30pt}
\begin{IEEEbiography}[{\includegraphics[width=1in,height=1.25in,clip,keepaspectratio]{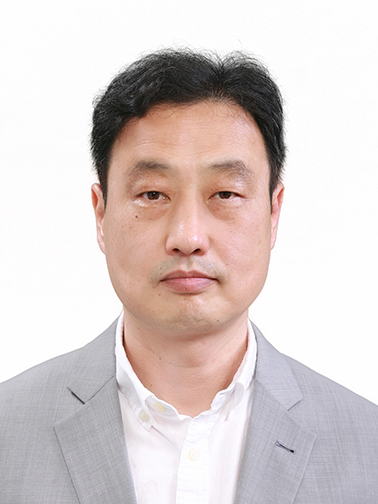}}]{Youngwoo Seo}
  Dr. Youngwoo Seo is a field-roboticist with experience of building mobile robots such as self-driving cars, drones, a high-speed transport – hyperloop, unmanned ground vehicles for more than two decades, and a seasoned executive with track records of managing diverse teams to deliver what matters. He currently serves as an Executive Vice President at Hanwha Aerospace. For this role, his primary goal is to spearhead the endeavor of stepping up the company’s game to become a global top-tier, defense solution provider. To that end, among other responsibilities, he has been leading and overseeing robotics and autonomous systems R\&D, carving out the international unmanned systems market, and playing a role of technology catalyst for AI, ML and robotics. He earned a Ph.D. and a master’s degree in robotics from Carnegie Mellon University, and a master’s degree in computer science from Seoul National University.
\end{IEEEbiography}
\vspace*{-30pt}
\begin{IEEEbiography}[{\includegraphics[width=1in,height=1.25in,clip,keepaspectratio]{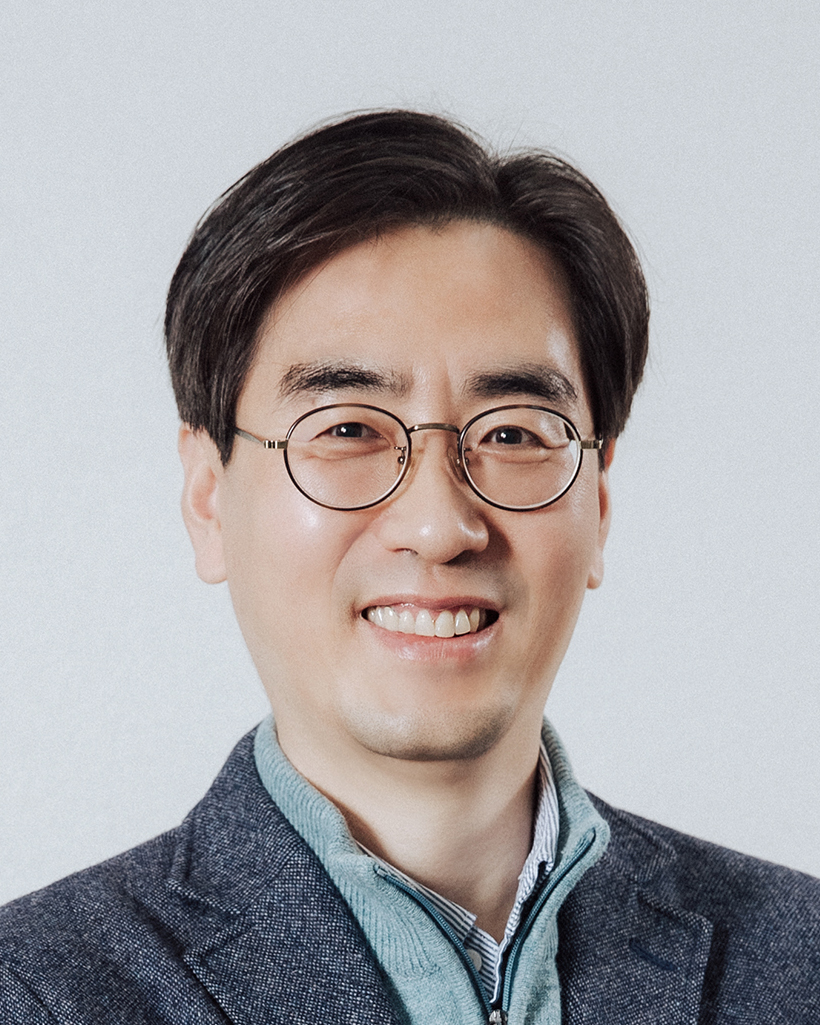}}]{Hyun Myung}
	 received the B.S., M.S., and Ph.D. degrees in electrical engineering from the Korea Advanced Institute of Science and Technology (KAIST), Daejeon, Republic of Korea, in 1992, 1994, and 1998, respectively.
     He was a Senior Researcher with the Electronics and Telecommunications Research Institute, Daejeon, from 1998 to 2002, a CTO and the Director with the Digital Contents Research Laboratory, Emersys Corporation, Daejeon, from 2002 to 2003, and a Principle Researcher with the Samsung Advanced Institute of Technology, Yongin, Korea, from 2003 to 2008.
     Since 2008, he has been a Professor with the Department of Civil and Environmental Engineering, KAIST, and he was the Chief of the KAIST Robotics Program.
     From 2019, he is a Professor with the School of Electrical Engineering.
     His current research interests include autonomous robot navigation, SLAM (simultaneous localization and mapping), SHM (structural health monitoring), spatial AI/machine learning, and swarm robots.
\end{IEEEbiography}

\end{document}